\documentclass{article}

\PassOptionsToPackage{numbers,compress}{natbib}
 \usepackage[preprint]{neurips_2026}

\usepackage[utf8]{inputenc} 
\usepackage[T1]{fontenc}    
\usepackage{hyperref}       
\usepackage{url}            
\usepackage{booktabs}       
\usepackage{amsfonts}       
\usepackage{nicefrac}       
\usepackage{xcolor}         
\usepackage{amsmath}
\usepackage{makecell}
\usepackage{multirow}
\usepackage[table]{xcolor}
\usepackage{subcaption} 
\usepackage{tikz}
\usepackage{wrapfig}
\usepackage{microtype}      
\usepackage{tabularx}
\usetikzlibrary{shapes.geometric, arrows, positioning, fit, backgrounds, calc}

\newcommand{\best}[1]{\cellcolor{orange!25}#1}

\title{CausalMotion: Structured Physical Reasoning as Keyframe and Trajectory Guidance for Training-Free Video Generation}

\author{%
  Sihan Zhuang$^{1,2 \star}$, Xinyuan Chen$^{1\dagger}$, Tianfan Xue$^{3,1\dagger}$, Yaohui Wang$^{1\dagger}$ \\
$^{1}$Shanghai Artificial Intelligence Laboratory, $^{2}$ShanghaiTech University \\
$^{3}$The Chinese University of Hong Kong \\
}

\begin{document}
{\let\thefootnote\relax\footnotetext{$^\star$ Work done during an internship at Shanghai AI Laboratory.}}
{\let\thefootnote\relax\footnotetext{$^\dagger$ Corresponding author.}}

\maketitle

\vspace{-4mm}
\begin{figure}[htbp]
    \centering
    \includegraphics[width=\linewidth]{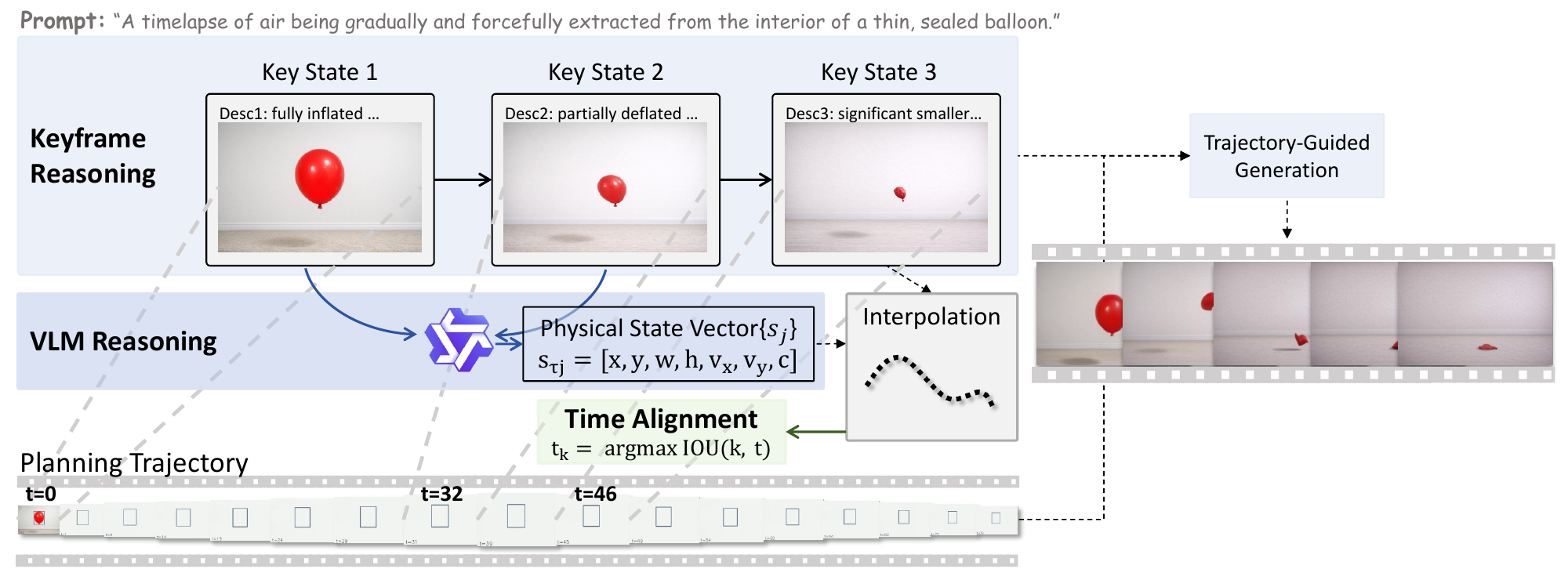}
    \caption{\textbf{Overview of CausalMotion.} Our training‑free framework performs structured physical reasoning to generate keyframes and trajectories, then guides the diffusion model with these intermediate representations.
}
    \label{fig:overview}
\end{figure}

\begin{abstract}
Recent advances in diffusion-based video generation have significantly improved visual quality and short-term temporal coherence. However, existing methods still struggle to produce videos with physically consistent and causally plausible dynamics, especially in scenarios involving long-horizon interactions. This limitation arises from the fact that video diffusion models primarily learn physical consistency implicitly, while vision-language models can directly model physical laws. Based on this idea, in this work, we propose \textbf{CausalMotion}, a training-free framework that injects explicit physical reasoning into video generation through structured intermediate representations. Our key idea is to decouple reasoning from generation by leveraging a vision-language model to decompose a text prompt into a sequence of causally consistent keyframes and object-centric motion trajectories. These representations are then aligned and integrated as soft constraints to guide a pretrained video diffusion model during inference. This design enables explicit modeling of object dynamics and causal transitions without requiring additional training or supervision. Extensive experiments show that our method consistently improves physical plausibility and temporal coherence, particularly in dynamics-intensive scenarios, while maintaining high perceptual video quality. Project Page: \url{https://zhuangsh0713.github.io/CausalMotion/}.

\end{abstract}

\section{Introduction}
\label{sec:intro}
Video generation aims to synthesize temporally coherent and visually realistic dynamic sequences conditioned on inputs such as text prompts, reference images, or motion signals. Recent advances in generative modeling, particularly diffusion-based approaches \cite{wan2025wan,kong2024hunyuanvideo,yang2024cogvideox}, have significantly improved visual fidelity and short-term temporal consistency. However, these models still struggle with scenarios that require physically consistent and causally valid state transitions, especially in compositional settings involving long-horizon dynamics.

A key limitation of existing approaches is their reliance on implicitly learned statistical correlations from large-scale video data, with limited explicit modeling of causal structure or physical constraints~\cite{motamed2025generative}. For instance, Runway Gen 3~\cite{runway2024platform} generates a candle appearing and being lit when a burning match is lowered into water, producing visually realistic but physically impossible frames. Moreover, high-quality datasets with rich physical interactions are scarce and typically lack structured annotations of causality and physics, making it difficult for models to learn generalizable reasoning capabilities~\cite{kang2024far}. Consequently, generated videos often exhibit failure modes such as missing intermediate states (e.g., an ice cream cone melts directly from solid to puddle, skipping all softened stages), temporal discontinuities (objects flash from one location to another), or physically implausible behaviors (objects passing through other objects).

In parallel, multimodal large models (MLLMs)~\cite{hurst2024gpt, deepseekai2025deepseekv32pushingfrontieropen, bai2025qwen3vltechnicalreport, vteam2026glm45vglm41vthinkingversatilemultimodal, guo2025seed15vltechnicalreport}
have demonstrated strong capabilities in cross-modal understanding and structured reasoning. Through chain-of-thought inference~\cite{wei2022chain, zhang2022automatic, zhang2024mathverse, kojima2022large}, these models can explicitly decompose complex events into causal steps and infer physically plausible state transitions. For instance, an MLLM can reason that a water droplet on a sloped surface will flow downward along the steepest path. However, MLLMs lack the ability to generate temporally consistent visual content, limiting their direct applicability to video synthesis.

This gap motivates the integration of explicit causal and physical reasoning into video generation without retraining generative models. In this paper, we propose CausalMotion, a training-free, reasoning-guided video generation framework. Our key idea is to decouple reasoning from generation, and use a vision-language model to explicitly structure the generation process via intermediate representations. As illustrated in Figure~\ref{fig:overview}, our framework operates in three stages. First, a VLM decomposes the input prompt into causally consistent keyframes. Using object locations from the first keyframe and textual descriptions of all keyframes, the VLM predicts a motion trajectory (bounding boxes tracking object changes over time). Keyframes and trajectories are temporally aligned, then guide the diffusion model's latent updates to synthesize the final video.

Our approach introduces explicit causal structure and physical priors without modifying model parameters or requiring supervision. Operating at inference time, it remains lightweight while improving temporal coherence and physical plausibility, achieving a PhyGenBench average score of 0.65 (67\% above baseline) and a VBench quality score of 82.52\% (surpassing LTX-Video at 80.57\% and Wan2.1 at 76.21\%).

We summarize our contributions as follows:
\begin{itemize}
\item We propose CausalMotion, a training-free framework that integrates vision-language model reasoning into video generation, enabling explicit modeling of causal structure and physical dynamics.
\item We introduce a structured intermediate representation consisting of keyframes and motion trajectories, which reduces ambiguity in temporal modeling and improves controllability.
\item Our method operates entirely at inference time, requiring no additional training or annotations, achieving a state-of-the-art average score of 0.65 on PhyGenBench (a 67\% improvement over the LTX-Video baseline of 0.39).
\end{itemize}

\section{Related Work}
\subsection{Video Generation}

Diffusion-based video generation has emerged as the dominant paradigm for high-quality synthesis.
Recent models, such as (\cite{wan2025wan, kong2024hunyuanvideo, ali2025world}), achieve strong visual fidelity and short-term temporal consistency by learning spatiotemporal distributions from large-scale data.
However, these approaches primarily model statistical correlations rather than explicit causal structure or physical dynamics.
As highlighted by VBench~\cite{zheng2025vbench}, current progress largely focuses on perceptual quality and motion smoothness, leaving physical consistency and causal reasoning underexplored.
Our work addresses this limitation by introducing explicit, inference-time physical grounding.

\subsection{Physics-Consistent Video Generation}

\paragraph{Simulator-based methods.}
Methods such as PhysGen~\cite{liu2024physgen}, NewtonGen~\cite{Yuan_2025_NewtonGen}, and follow-ups ~\cite{tan2026physmotion,chen2025physgen3d,li2025wonderplay,realwonder2026} integrate physics simulators into the generation pipeline by estimating scene properties and explicitly modeling object dynamics.
While physically grounded, they rely on simplified assumptions (e.g., rigid-body dynamics) and introduce substantial computational and modeling overhead, limiting scalability to open-world scenarios.

\paragraph{Non-simulator methods.}
An alternative line of work enforces physical consistency without explicit simulation, either through training-based alignment or foundation-model guidance.
Training-based approaches incorporate physical priors via losses, rewards, or reinforcement learning, such as PISA ~\cite{li2025pisa}, Force Prompting~\cite{gillman2025force}, NewtonRewards~\cite{le2025newtonrewards}, and PhysRVG~\cite{zhang2026physrvg}, but require large-scale training and carefully designed supervision, with performance tightly coupled to constraint quality.
Complementarily, recent methods leverage vision-language or video foundation models to provide structured guidance, including VLIPP~\cite{yang2025vlipp}, Think Before You Diffuse~\cite{zhang2025thinkdiffusellmsguidedphysicsaware}, VideoREPA~\cite{zhang2025videorepa}, ProPhy~\cite{wang2025prophyprogressivephysicalalignment}, and PhyRPR~\cite{zhao2026phyrpr}.
These methods inject high-level physical priors via intermediate representations (e.g., motion attributes).


\subsection{Vision-Language Models for Reasoning}

Chain-of-Thought (CoT) prompting~\cite{wei2022chain} enables large models to decompose complex reasoning into structured intermediate steps and has been extended to multimodal settings for physical and causal reasoning.
Prior works exploit VLMs to generate intermediate representations such as layouts or semantic plans~\cite{lian2023llm,wu2024self,pan2024vlp}, while VChain~\cite{huang2025vchain} demonstrates their ability to produce physically plausible intermediate states. Building on these advances, we use VLM reasoning to generate structured guidance (keyframes and physical state representations).


\section{Method}
\label{sec:method}

\begin{figure}[htbp]
    \centering
    \includegraphics[width=\linewidth]{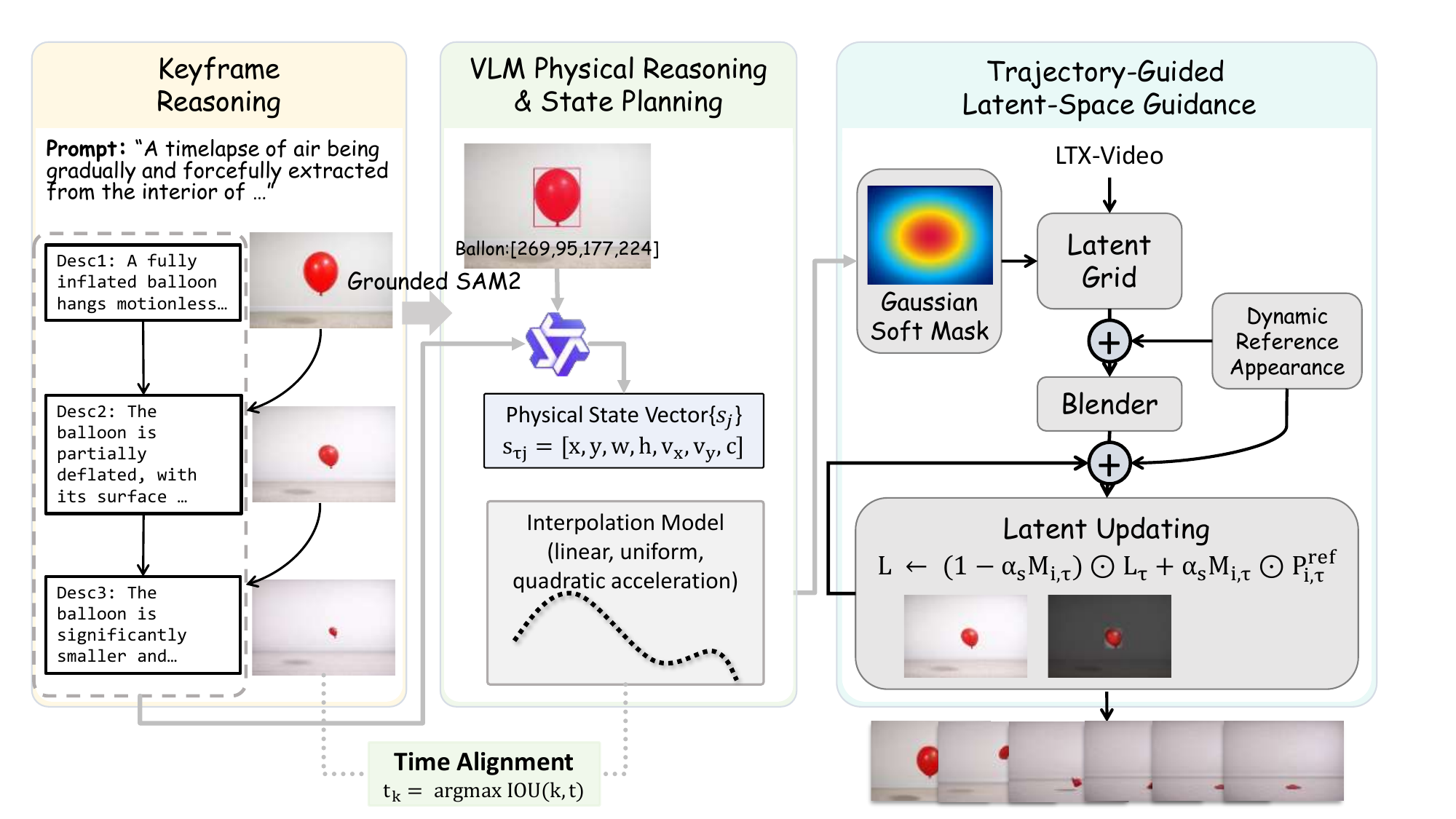}
    \caption{\textbf{The architecture of our method.}
    Given a text prompt, CausalMotion first performs iterative visual reasoning to decompose complex events into causally consistent key states and generates corresponding keyframes that capture critical transitions.
    It then localizes key objects, predicts sparse physical state vectors, and constructs dense motion trajectories through physics-aware interpolation and temporal alignment.
    Finally, the generated trajectories are projected into the diffusion latent space, where appearance anchors from keyframes and localized latent updates jointly guide denoising toward physically plausible motion and temporally consistent video generation.
}
    \label{fig:method}
\end{figure}

We propose CausalMotion, a \textbf{Physical Grounding} framework for video generation that improves temporal coherence and physical plausibility without any additional training or fine-tuning. Our key idea is to decouple video generation into structured state reasoning and trajectory-constrained synthesis. Instead of relying on implicit dynamics learned by diffusion models or depending on existing physics simulator, we explicitly model video evolution as a sequence of causally consistent states and physically grounded trajectories, and use them to guide the generation process.

Formally, we reinterpret video generation as a \emph{structured state evolution problem}, where each frame is constrained by (i) key causal states and (ii) physically consistent object trajectories. As shown in Figure~\ref{fig:method}, our framework instantiates this idea in three stages: (1) keyframe generation via iterative visual reasoning, (2) physical state and trajectory planning, and (3) trajectory-guided latent-space video generation. 

\subsection{Keyframe Generation via Visual Thought Reasoning}
\label{sec:visual_thought_reasoning}

Text prompts often implicitly encode a sequence of state transitions, including object motion, interaction, and transformation. However, such structure is not explicitly modeled in standard video generation pipelines, leading to inconsistencies in long-horizon dynamics.

We address this by leveraging a vision-language model (VLM) to perform \emph{Visual Thought Reasoning}, which decomposes the input prompt into a sequence of causally consistent intermediate states or events.
Given a prompt $p$, the VLM first performs a global scene understanding step and infers the \emph{consequence}—the most likely whole state or primary outcome of the described scenario. This global causal constraint anchors the subsequent keyframe generation.

Starting from the initial state and consequence, the model generates a textual description $\mathit{txt}_0$ of the first key state and produces the corresponding image $\mathit{img}_0$ via a text-to-image model (gpt-image-1 in our experiment).
In each subsequent step $i$, conditioned on the generated history $\mathit{img}_{i-1}$, previous description $\mathit{txt}_{i-1}$ and global constraint \emph{consequence}, the VLM predicts the next textual instruction $\mathit{txt}_i$ describing the key change at step $i$, yielding $\mathit{img}_i$. This autoregressive process continues until all key state transitions implied by \emph{consequence} are covered.

The final output is a \emph{Chain of Visual Thoughts and Captions}:
\begin{align*}
  \bigl\{\mathit{img}_0,\,\mathit{img}_1,\,\ldots,\,\mathit{img}_{N-1}\bigr\}
  \;\text{ and }\;
  \bigl\{\mathit{txt}_0,\,\mathit{txt}_1,\,\ldots,\,\mathit{txt}_{N-1}\bigr\},
\end{align*}
which provide structured, interpretable intermediate representations for object detection, trajectory planning, and physics-constrained video generation.

\subsection{Physical State and Trajectory Planning}
\label{sec:trajectory_reasoning}

Given the generated keyframes, we explicitly construct physically consistent object trajectories by modeling object dynamics as structured physical states rather than implicit motion patterns.

We first obtain the initial spatial locations of key objects identified by the VLM. Specifically, the VLM reasons over the text prompt and the sequence of keyframes to implicitly determine which objects are involved in physical changes. To localize these objects in image space, we apply a segmentation model (\cite{ravi2024sam2segmentimages,liu2023grounding,ren2024grounding,ren2024grounded,kirillov2023segany,jiang2024trex2}) to the initial keyframe image $img_0$, obtaining bounding boxes for all key objects $\{o_k\}$. Formally, the bounding box of the $k$-th object at the initial time step is given by:
\begin{equation}
  b_{0,k} = S_{seg}(img_0, o_k), \quad k = 1, 2, \ldots, K
  \label{eq:initial_bbox}
\end{equation}
where $S_{seg}(\cdot)$ denotes the segmentation model and $b_{0,k}$ corresponds to the initial bounding box of object $o_k$. These bounding boxes serve as the spatial initialization for subsequent trajectory planning.

To enable explicit modeling of object dynamics, we introduce a \textbf{Physical State Vector} for each object $i$ at key time $\tau_j$:
\begin{equation}
  \mathbf{s}_{\tau_j}^i =
  \bigl[
    x_{\tau_j}^i,\;
    y_{\tau_j}^i,\;
    w_{\tau_j}^i,\;
    h_{\tau_j}^i,\;
    v_{x,\tau_j}^i,\;
    v_{y,\tau_j}^i,\;
    \mathbf{c}_{\tau_j}^i
  \bigr],
  \label{eq:state_vector}
\end{equation}
where $(x,y,w,h)$ denote spatial extent, $(v_x, v_y)$ denote velocity, and $\mathbf{c}$ encodes contact relations. This representation unifies geometry, motion, and interaction.

The VLM acts as a high-level physical state planner. Its input consists of: (i) the video description that indicates the dominant physical laws governing the scene (e.g., gravity, momentum conservation, fluid dynamics), (ii) the initial keyframe image $img_0$, and (iii) the initial bounding box locations $\{b_{0,k}\}$. Through chain-of-thought reasoning, the VLM outputs key physical states $\{\mathbf{s}_{\tau_j}^i\}$ at selected timesteps $\{\tau_j\}$, inferring object motion based on learned physical priors such as gravity and collision. Specifically, the reasoning process proceeds as: (1) analyzing the scene description to identify applicable physical laws; (2) inferring potential interactions and motion trends under these physical constraints; (3) predicting bounding box positions and scales over time in image space.

Due to the context length constraints, the VLM predicts at a sparse set of key timesteps (e.g., 13 timesteps in our experiments). To match the temporal resolution required by the subsequent video diffusion model, we interpolate these key predictions to generate a dense trajectory of $T$ frames (121 frames in our experiments). Given these key states, we construct dense trajectories by interpolating between them using motion models selected according to the inferred physical regime. Specifically,
\begin{equation}
  b_t^i =
  \begin{cases}
    \text{linear interpolation} & \text{(uniform motion)} \\
    b_{\tau_j}^i + v_{\tau_j}^i\,\Delta t + \tfrac{1}{2} a^i (\Delta t)^2
    & \text{(accelerated motion)}
  \end{cases}
  \label{eq:adaptive_interpolation}
\end{equation}
where $\Delta t = t - \tau_j$. This interpolation preserves the overall motion trend while introducing smooth temporal variation, providing a continuous object motion prior for subsequent high-frame-rate video generation. This also enables modeling of both linear and non-linear motion patterns, avoiding the limitations of naive interpolation. 

\subsection{Time alignment}
\label{sec:time_align}

To ensure consistency between discrete keyframes and continuous trajectories, we assign each keyframe to a trajectory timestep. Let $t_k$ denote the assigned timestep for keyframe $k$. We perform a greedy assignment that respects temporal ordering: starting from $t_0 = 0$, for each subsequent keyframe $k = 1, \dots, K-1$, we select $t_k$ as the timestep that maximizes the IoU between keyframe $k$ and the trajectory at that timestep, subject to $t_{k-1} < t_k < T$. Formally,

\begin{equation}
t_k = \arg\max_{t \in (t_{k-1}, T-k)} \mathrm{IoU}(k, t), \quad k = 1, 2, \ldots, K,
\label{eq:greedy_alignment}
\end{equation}
with $t_0 = 0$. This greedy strategy enforces monotonicity while locally maximizing spatial alignment at each step. The resulting mapping ${t_k}$ provides explicit temporal anchors that bridge keyframe reasoning and trajectory dynamics. Further analysis of the consistency between keyframe predictions and trajectory representations is provided in Appendix~\ref{sec:discussion_traj_mapping}.

\subsection{Trajectory-Guided Latent-Space Guidance}
\label{sec:latent_guidance}

Beyond keyframe conditioning, we incorporate the physics-aware trajectory from Eq.~\eqref{eq:adaptive_interpolation} as a soft constraint during diffusion sampling, enabling a balance between the model prior and physical consistency of interpolated frames.

Following the temporal downsampling structure of the diffusion model, we first encode the per-frame trajectory into the latent temporal space:
\begin{equation}
  B_{i,\tau} =
  \Pi_{\mathrm{vae}}\!\left(
    \frac{1}{|I_\tau|}\sum_{t \in I_\tau} \bar{b}_i^t
  \right),
\end{equation}
where $\Pi_{\mathrm{vae}}$ denotes the pixel-to-latent scaling map.

To preserve object appearance across time, we extract an appearance anchor from each keyframe. Specifically, given keyframe $k$ with latent representation $Z_k$, we crop a local patch around the object position using a Gaussian soft mask:
\begin{equation}
  A_{i,\tau_k}
  = \mathrm{Crop}\!\bigl(Z_k,\;\mathrm{Shrink}(B_{i,\tau_k}, \rho_s)\bigr).
\end{equation}
These anchors serve as stable visual references during denoising.

However, directly using fixed anchors can lead to temporal inconsistency. To address this, we construct a dynamic reference appearance $P^{\mathrm{ref}}_{i,\tau}$ at each latent step $\tau$ by blending three sources: local propagation from the previous step, and anchors from two neighbouring keyframes:
\begin{equation}
  P^{\mathrm{ref}}_{i,\tau}
  = w_a^{-}\,P^{(-)}_{i,\tau}
  + w_p\,P^{(\mathrm{prev})}_{i,\tau}
  + w_a^{+}\,P^{(+)}_{i,\tau},
  \quad w_a^{-} + w_p + w_a^{+} = 1,
\end{equation}
where the weights are adaptively determined based on the temporal position of $\tau$ relative to adjacent keyframes. This formulation enables smooth appearance transition while maintaining temporal coherence.

Finally, we enforce trajectory-guided consistency via local latent updates. At each trajectory position $B_{i,\tau}$, we define a Gaussian soft mask:
\begin{equation}
  M_{i,\tau}(x,y)
  = \exp\!\left(-\frac{\|x - c_{i,\tau}\|^2}{2\sigma^2}\right),
\end{equation}
where $c_{i,\tau}$ denotes the trajectory center. The latent tensor is then updated during each denoising step as:
\begin{equation}
  L_{\tau}
  \;\leftarrow\;
  (1 - \alpha_s M_{i,\tau})\odot L_{\tau}
  \;+\;
  \alpha_s M_{i,\tau}\odot P^{\mathrm{ref}}_{i,\tau},
  \label{eq:latent_update}
\end{equation}
where $\alpha_s$ controls the guidance strength.
This update is spatially localized, preserving the global generative distribution while enforcing trajectory-aligned motion and appearance consistency.

\subsection{Implementation}
We adopt LTX-Video~\cite{hacohen2024ltx} as the video synthesis backbone, which supports conditioning on multiple keyframes at user‑specified temporal indices. To leverage the physics‑aware trajectory, we use the temporal anchors ${t_k}$ obtained from Eq.~\eqref{eq:greedy_alignment} as the explicit timesteps for keyframe conditioning. For each anchor $t_k$, we provide the corresponding keyframe $img_k$ from ~\ref{sec:visual_thought_reasoning}. 

In addition to keyframe conditioning, LTX-Video expects a scalar guidance strength parameter $\lambda$ for each keyframe call. For each temporal anchor $t_k$ obtained from Eq.~\eqref{eq:greedy_alignment}, we invoke LTX-Video with the corresponding keyframe $k$ and its associated strength $\lambda$ (a fixed hyperparameter). We choose an appropriate hyperparameter in our experiment (e.g. 0.9 for all frames). 

\section{Experiments}

\subsection{Experiments Details}
For keyframe generation, we adopt gpt-image-1.5, a high-fidelity text-to-image model with strong instruction-following capability and consistent visual rendering across iterative generations. 
For trajectory reasoning, we employ Qwen-VL-2.5-72B~\cite{bai2025qwen25vltechnicalreport} as the vision-language model, leveraging its strong multimodal understanding and structured reasoning ability to infer object dynamics and interactions.

All experiments are conducted at a spatial resolution of $720 \times 480$ and a frame rate of 30 FPS. 
For text-to-video (T2V) generation, we follow the default configuration of LTX-Video, generating videos of 121 frames, corresponding to approximately 4 seconds in duration.

We evaluate our method on both physical reasoning and perceptual quality benchmarks to provide a comprehensive assessment.

\textbf{PhyGenBench}~\cite{pmlr-v267-meng25c} is designed to evaluate physical understanding in video generation. 
It covers a diverse set of physical phenomena and explicit physical laws across four categories: \emph{mechanics}, \emph{optics}, \emph{thermal}, and \emph{material}.

\textbf{VBench}~\cite{huang2024vbench} is used to evaluate perceptual video quality independent of physical reasoning. 
It assesses key dimensions of video generation from both temporal and visual quality perspectives. This complementary evaluation allows us to verify that improvements in physical realism do not come at the cost of perceptual quality.

\subsection{Evaluation}

\paragraph{Quantitative comparisons}
We first quantitatively evaluate our method on PhyGenBench using the PhyGenEval protocol. Table~\ref{tab:quantitative} reports the performance across four categories of physical phenomena, including mechanics, optics, thermal, and material properties.

\begin{table}[htbp]
\centering
\small
\setlength{\tabcolsep}{4pt}
\begin{tabular}{lccccc}
\toprule
\textbf{Model Variant} & \textbf{Mechanics $\uparrow$} & \textbf{Optics $\uparrow$} & \textbf{Thermal$\uparrow$} & \textbf{Material$\uparrow$} & \textbf{Average$\uparrow$} \\
\midrule
CogvideoX-T2V-5B~\cite{yang2024cogvideox} & 0.43 & 0.55 & 0.40 & 0.42 & 0.45 \\
LTX-Video-T2V~\cite{hacohen2024ltx} &0.35 &0.45 &0.36 &0.38 &0.39 \\
OpenSora~\cite{opensora} &0.43 &0.50 &0.44 &0.37 &0.44 \\
\midrule
CogvideoX-I2V-5B &0.48 &0.69 &0.43 &0.41& 0.52 \\
SVD-XT~\cite{blattmann2023stable} & 0.46 &0.68& 0.48& 0.41& 0.52 \\
LTX-Video-I2V &0.47& 0.65& 0.46& 0.37& 0.50 \\
SG-I2V~\cite{namekata2024sg} &0.52 &0.69 &0.51 &0.39 &0.54 \\
\midrule
LLM-Grounding Video Diffusion~\cite{lian2023llm} &0.32 &0.41 &0.26 &0.24 &0.31\\
PhyT2V~\cite{xue2025phyt2v} &0.49 &0.61 &0.49 &0.47 &0.52 \\
VideoDPO~\cite{liu2025videodpo} &0.48 &0.60 &0.47 &0.58 &0.54 \\
Diffphy~\cite{zhang2025thinkdiffusellmsguidedphysicsaware} &0.53 &0.59 &0.58 &0.46 &0.54 \\
PhyGDPO~\cite{cai2025phygdpo} &0.55 &0.60 &0.58 &0.47 &0.55 \\
VLIPP~\cite{yang2025vlipp} &0.55 & \best{0.71} &0.60 &0.53 &0.60 \\
CausalMotion  & \best{0.61} & \best{0.71} & \best{0.68} & \best{0.61} &  \best{0.65}\\
\bottomrule

\end{tabular}
\caption{Quantitative comparison on PhyGenBench across four categories of physical phenomena. Our method achieves the best overall performance, with notable improvements in dynamics-intensive scenarios such as mechanics and thermal.}
\label{tab:quantitative}

\end{table}

Our model achieves the best performance across all categories, reaching an average score of 0.65 and outperforming all baselines by a clear margin. Notably, the improvements are most significant in \emph{mechanics} (+0.06 over VLIPP) and \emph{thermal} (+0.08 over VLIPP), which require accurate modeling of object dynamics and causal interactions. Compared to strong baselines such as VLIPP~\cite{yang2025vlipp} and PhyGDPO~\cite{cai2025phygdpo}, our method achieves higher performance without requiring additional training in challenging scenarios. It highlights the advantage of structured reasoning and trajectory constraints at inference time.

\begin{table}[htbp]
  \centering
  \setlength{\tabcolsep}{4pt}
  \begin{tabular}{l c c c c c}
    \toprule
    \multirow{3}{*}{\textbf{Model Variant} } 
    & \multirow{3}{*}{\makecell{VBench \\ Quality Score $\uparrow$}}
    & \multicolumn{4}{c}{VLM-as-judge $\uparrow$}  \\
    
    \cmidrule(lr){3-6}
    & 
    & \makecell{Physical\\plausibility}
    & \makecell{Temporal\\consistency}
    & \makecell{Semantic\\alignment} 
    & Overall \\
    
    \midrule
    Wan2.1-T2V-1.3B~\cite{wan2.1}
      & 76.21\% 
      & 1.58
      & 3.58
      & 1.16
      & 2.10 \\
    VChain 
      & 78.49\%
      & -
      & -
      & -
      & - \\
    LTX-Video 
      & 80.57\% 
      & 2.53
      & 3.89
      & 1.68
      & 2.70 \\

    -w/o keyframe reasoning
    &80.91\%&2.60&4.65&2.35&3.20 \\

    -w/o trajectory physical state
    &82.44\%&3.05&3.65&3.40&3.37 \\

    -w/o latent trajectory guidance
    &82.46\%&3.20&3.90&3.45&3.55 \\

    CausalMotion 
      & \best{82.52}\%
      & \best{3.10}
      & \best{3.95}
      & \best{3.70}
      & \best{3.58} \\
    \bottomrule
  \end{tabular}
  \caption{Video quality and high-level consistency evaluation using VBench and VLM-as-judge. Our method improves both perceptual quality and higher-level properties, including physical plausibility, temporal consistency, and semantic alignment.}
  \label{tab1::t2v_results}
\end{table}

To further assess perceptual quality, we evaluate our method on VBench using the same 20 prompts as VChain~\cite{huang2025vchain} for fair comparison.

As shown in Table~\ref{tab1::t2v_results}, our model achieves the highest VBench score (82.52\%), indicating that introducing physical constraints does not degrade visual quality. Instead, it slightly improves overall generation quality compared to the base LTX-Video model.

Beyond standard metrics, we employ a vision-language model (Gemini 2.5 Flash~\cite{comanici2025gemini}) as a judge to evaluate higher-level properties, including \emph{physical plausibility}, \emph{temporal consistency}, and \emph{semantic alignment}. Our method significantly outperforms all models across all dimensions, improving the overall score from 2.70 to 3.58. In particular, the gain in temporal consistency (3.89 $\rightarrow$ 4.58) indicates that trajectory guidance effectively stabilizes motion across frames, while the large improvement in semantic alignment (1.68 $\rightarrow$ 4.00) suggests that keyframe reasoning helps better ground the generated content to the input prompt.

\paragraph{Qualitative comparisons}

\begin{figure*}[htbp]
\centering
\setlength{\tabcolsep}{0pt}
\renewcommand{\arraystretch}{0}

\vspace{-4mm}
\begin{minipage}[t]{0.49\textwidth}
\centering
Milk is poured into a cup of black coffee. \\[2mm]

\begin{tabular}{c c c c c}
\raisebox{2\height}{(a)} &
  \includegraphics[width=0.24\textwidth]{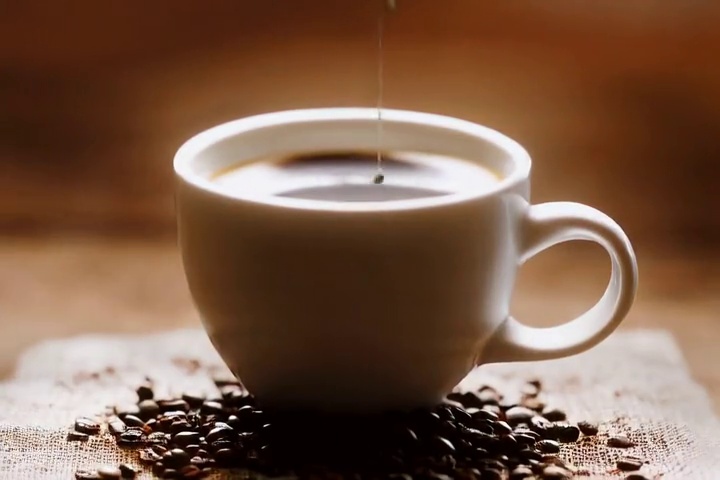} &
  \includegraphics[width=0.24\textwidth]{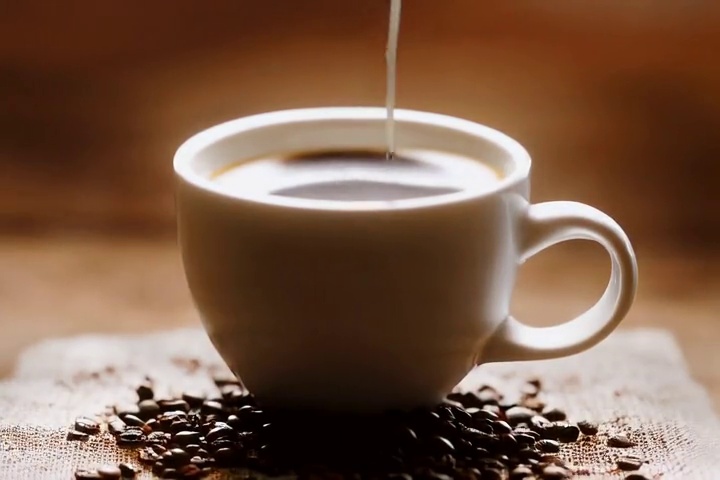} &
  \includegraphics[width=0.24\textwidth]{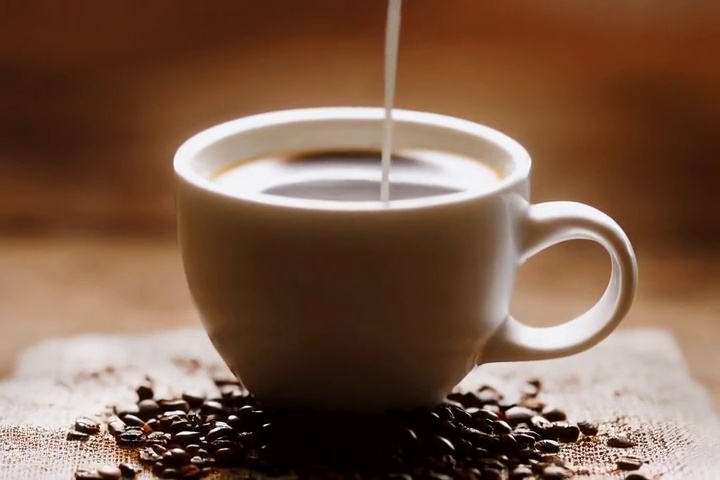} &
  \includegraphics[width=0.24\textwidth]{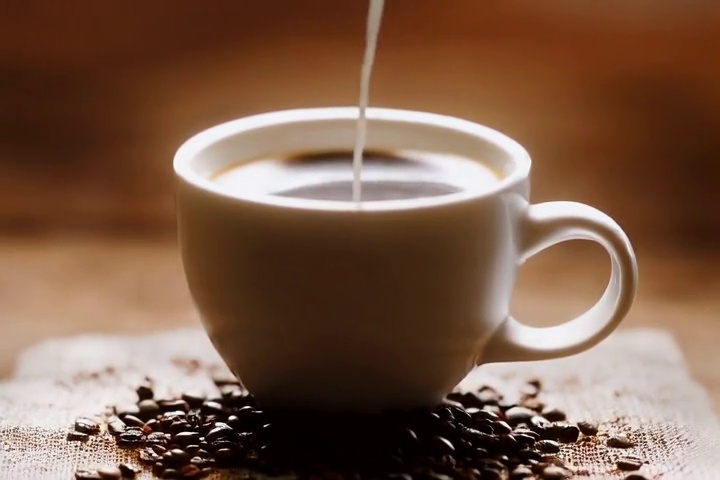} \\
\raisebox{2\height}{(b)} &
  \includegraphics[width=0.24\textwidth]{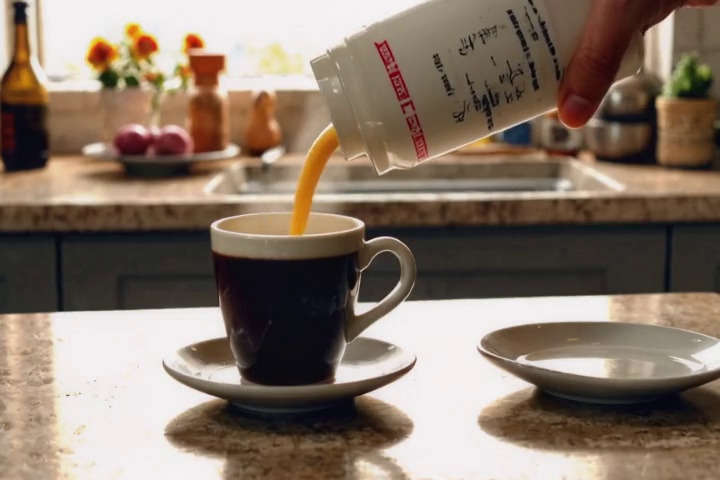} &
  \includegraphics[width=0.24\textwidth]{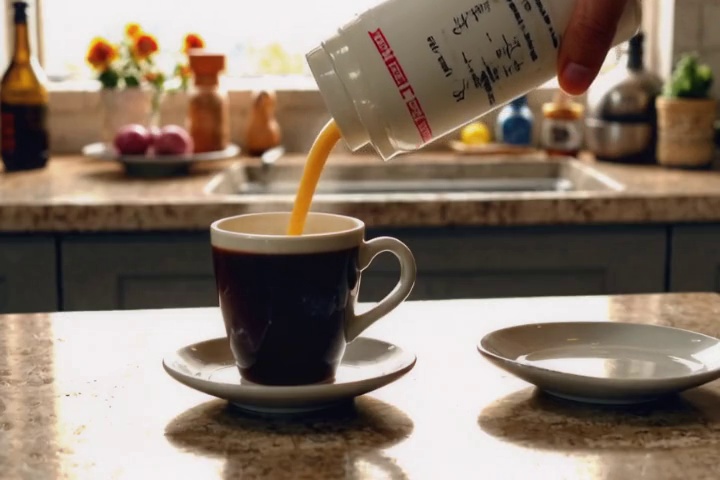} &
  \includegraphics[width=0.24\textwidth]{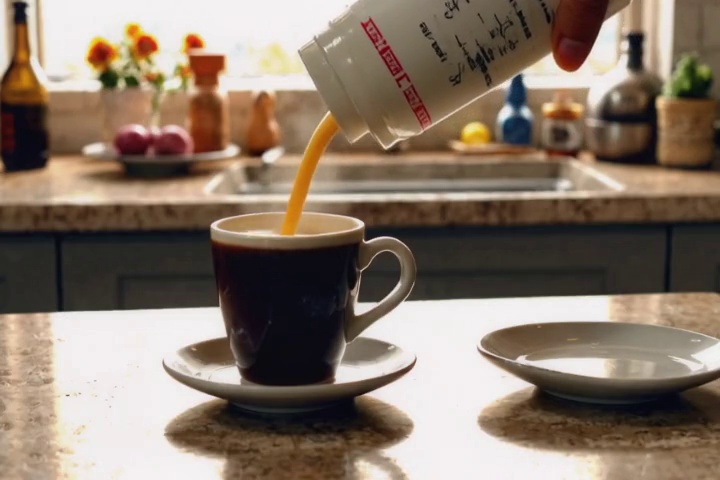} &
  \includegraphics[width=0.24\textwidth]{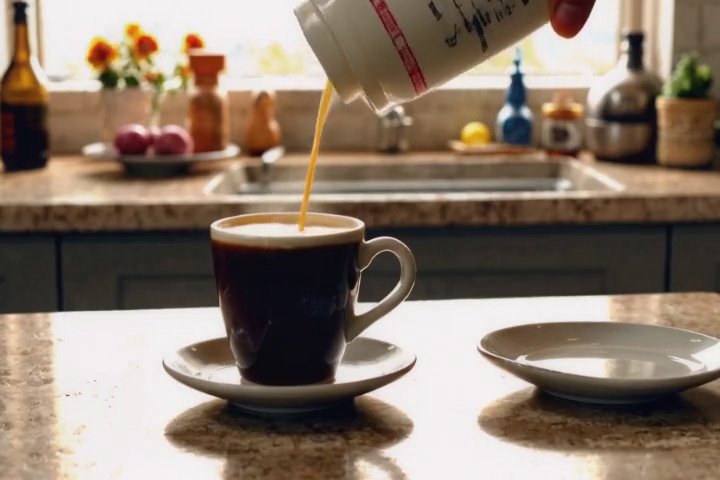} \\
\raisebox{2\height}{(c)} &
  \includegraphics[width=0.24\textwidth]{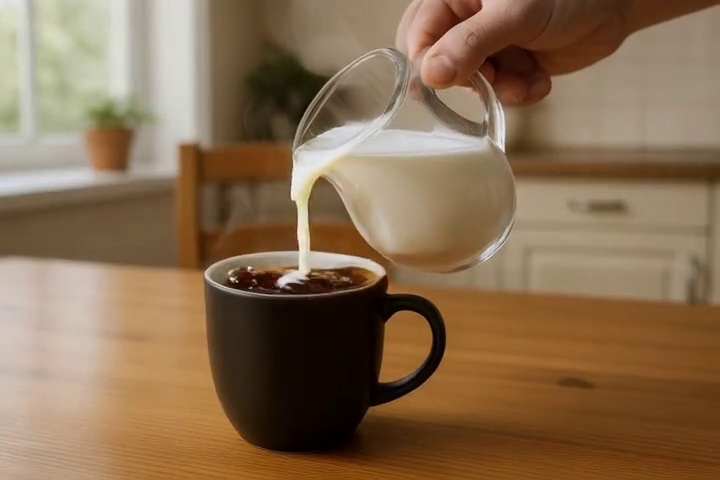} &
  \includegraphics[width=0.24\textwidth]{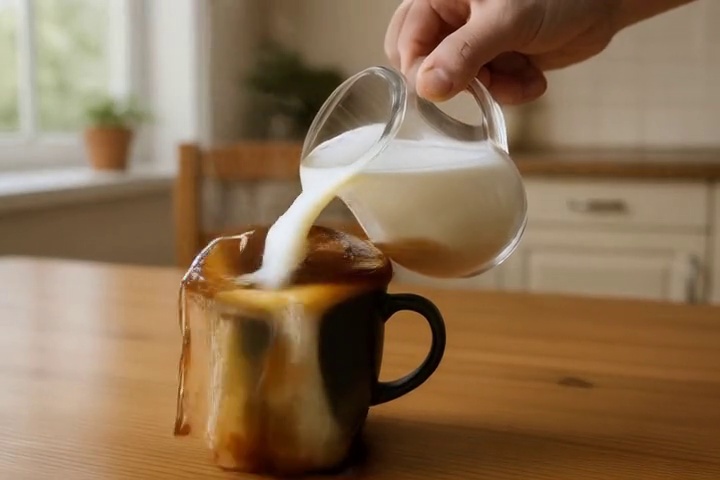} &
  \includegraphics[width=0.24\textwidth]{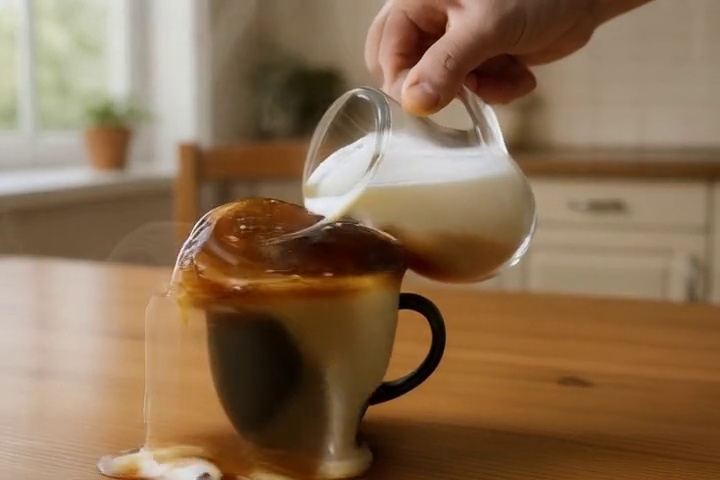} &
  \includegraphics[width=0.24\textwidth]{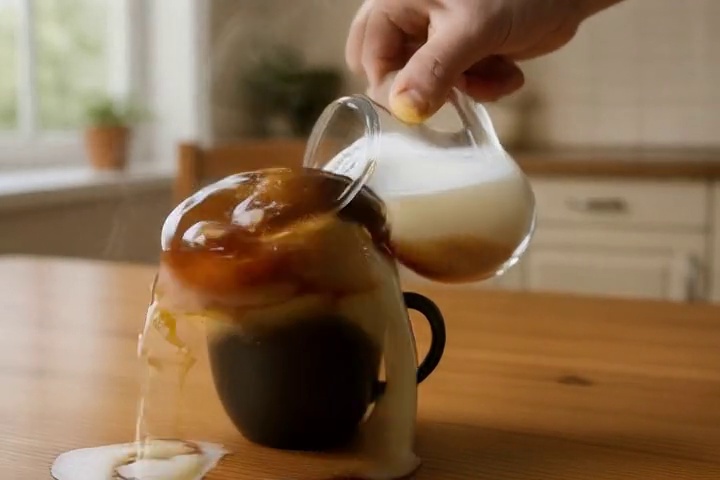} \\
\raisebox{2\height}{(d)} &
  \includegraphics[width=0.24\textwidth]{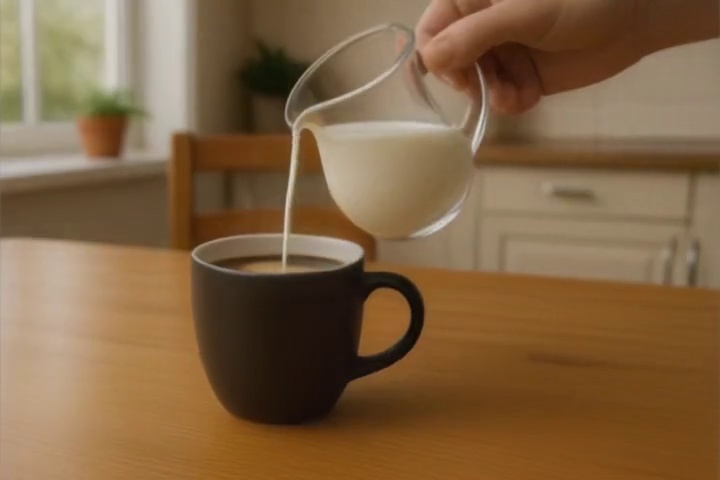} &
  \includegraphics[width=0.24\textwidth]{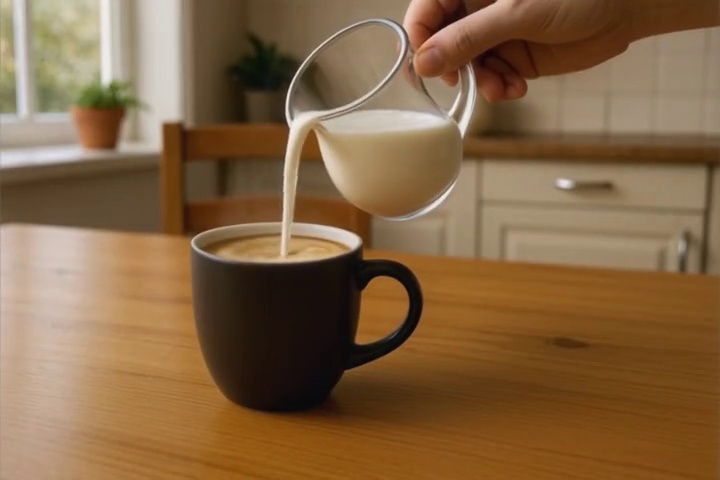} &
  \includegraphics[width=0.24\textwidth]{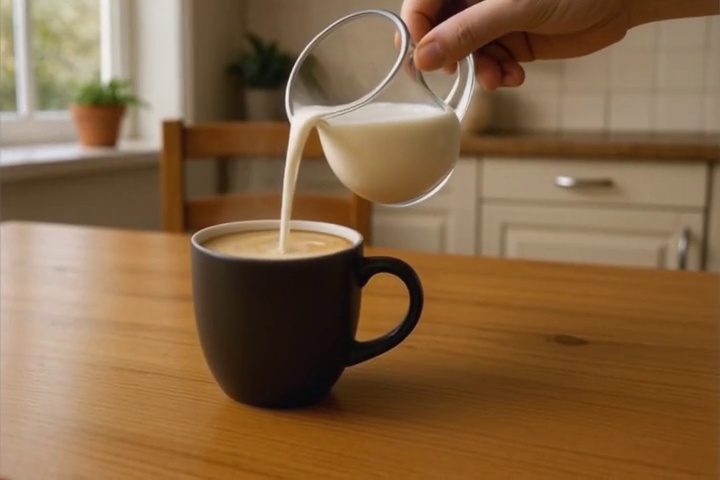} &
  \includegraphics[width=0.24\textwidth]{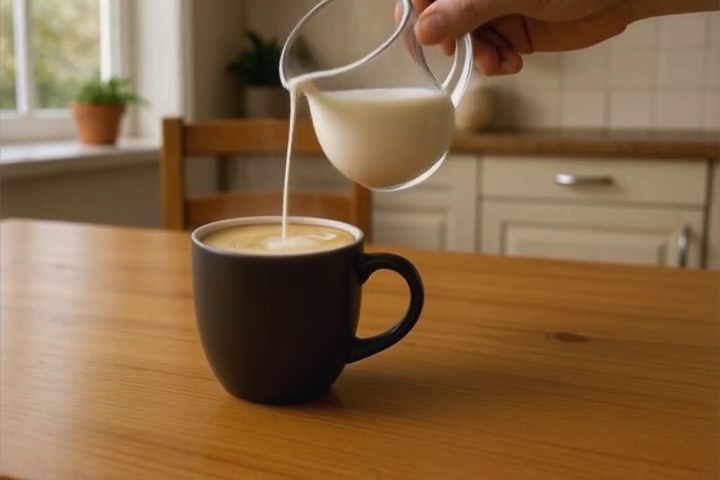} \\
\end{tabular}
\end{minipage}%
\hfill%
\begin{minipage}[t]{0.49\textwidth}
\centering
An ice cream cone is left out in the sun. \\[2.7mm]

\begin{tabular}{c c c c c}
\includegraphics[width=0.24\textwidth]{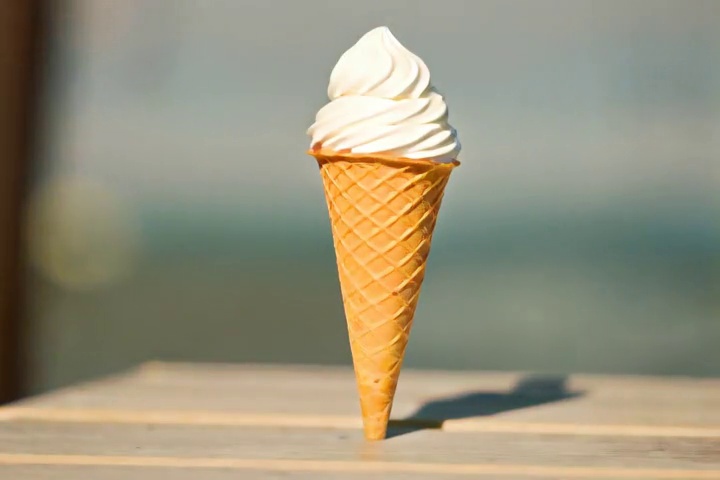} &
\includegraphics[width=0.24\textwidth]{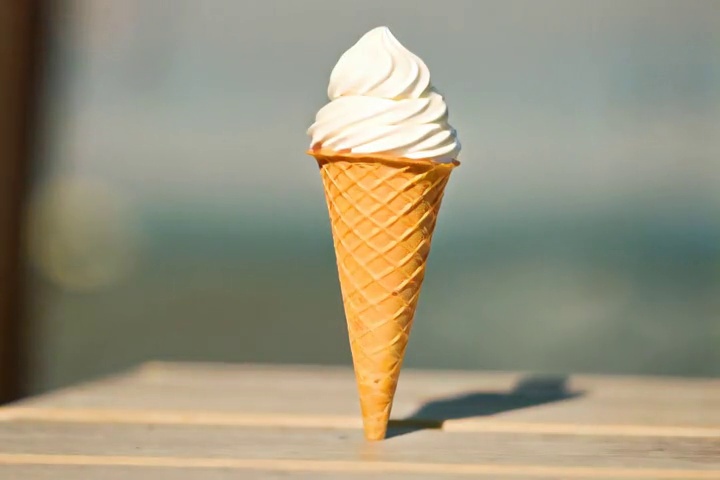} &
\includegraphics[width=0.24\textwidth]{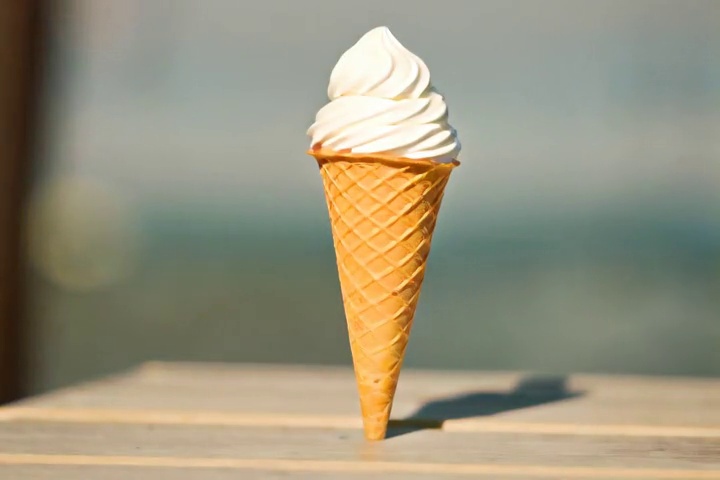} &
\includegraphics[width=0.24\textwidth]{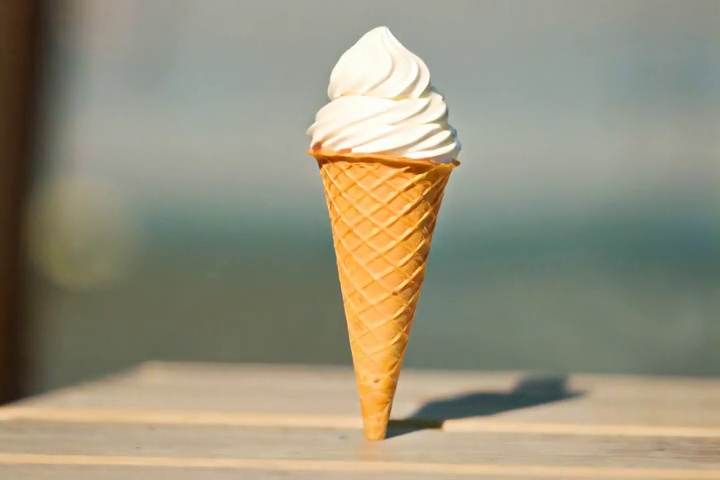} \\
\includegraphics[width=0.24\textwidth]{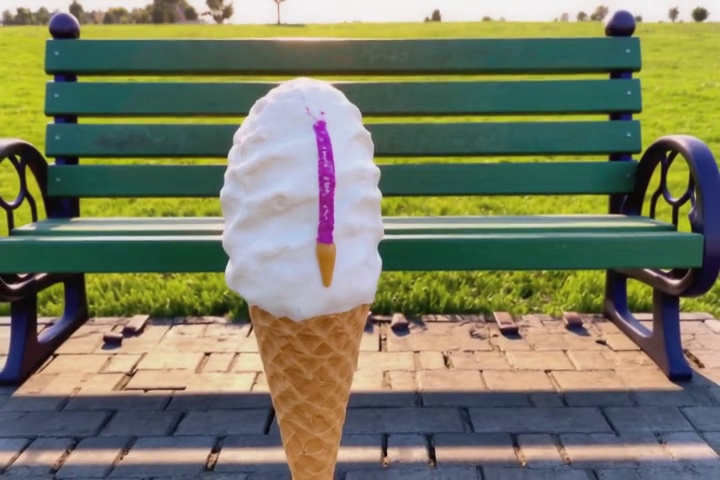} &
\includegraphics[width=0.24\textwidth]{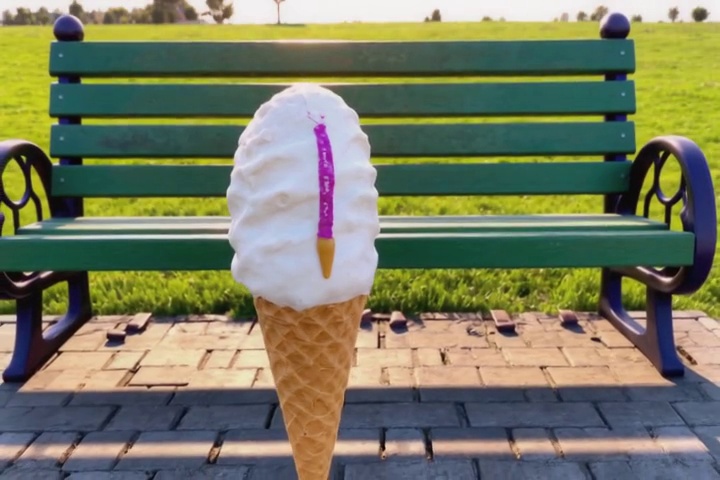} &
\includegraphics[width=0.24\textwidth]{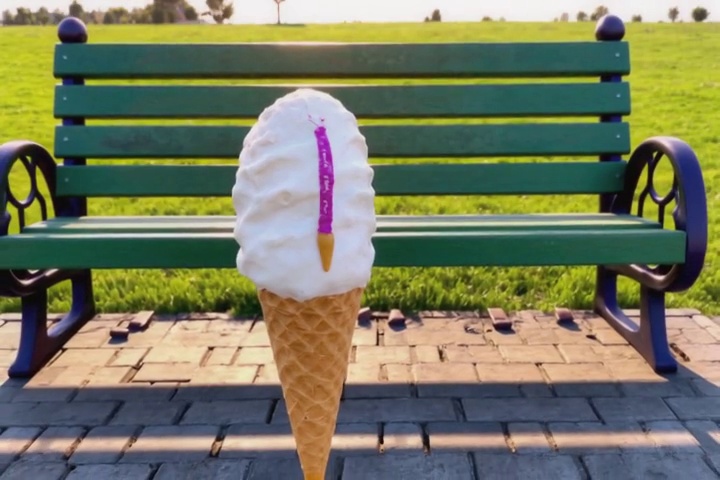} &
\includegraphics[width=0.24\textwidth]{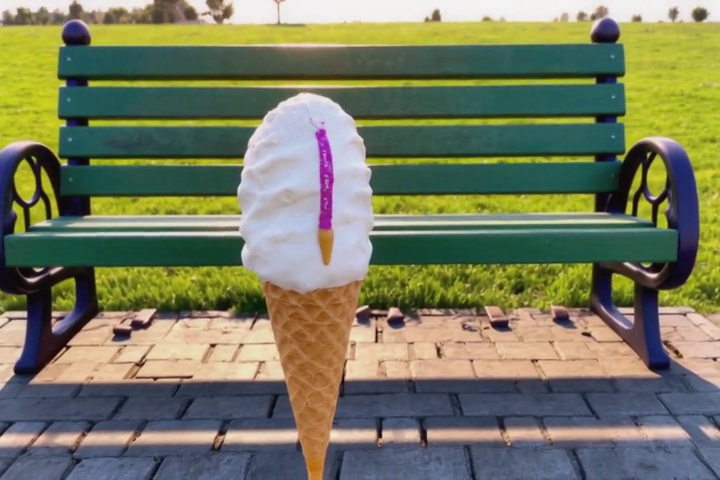} \\
\includegraphics[width=0.24\textwidth]{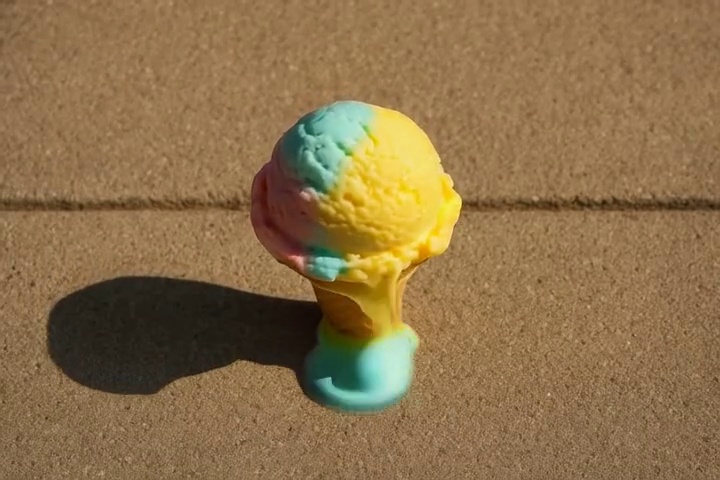} &
\includegraphics[width=0.24\textwidth]{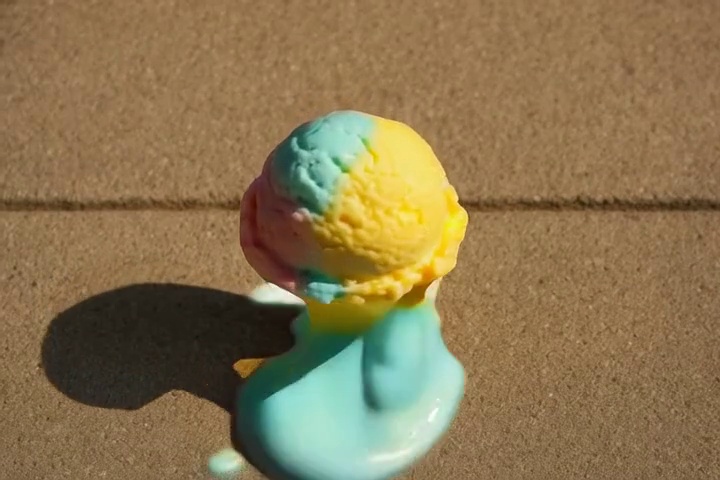} &
\includegraphics[width=0.24\textwidth]{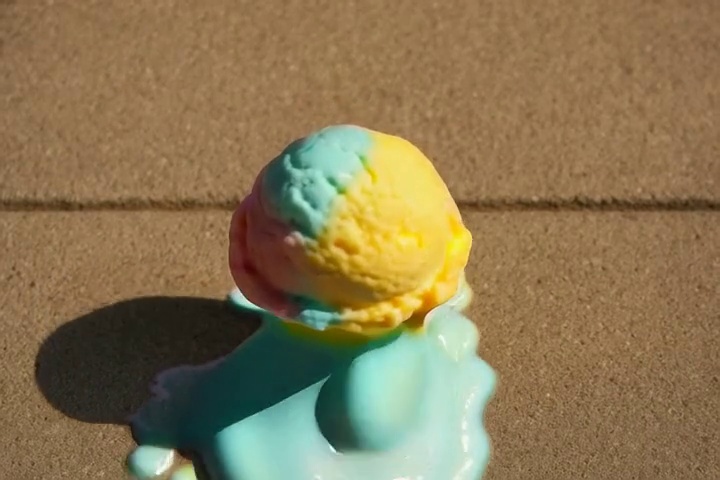} &
\includegraphics[width=0.24\textwidth]{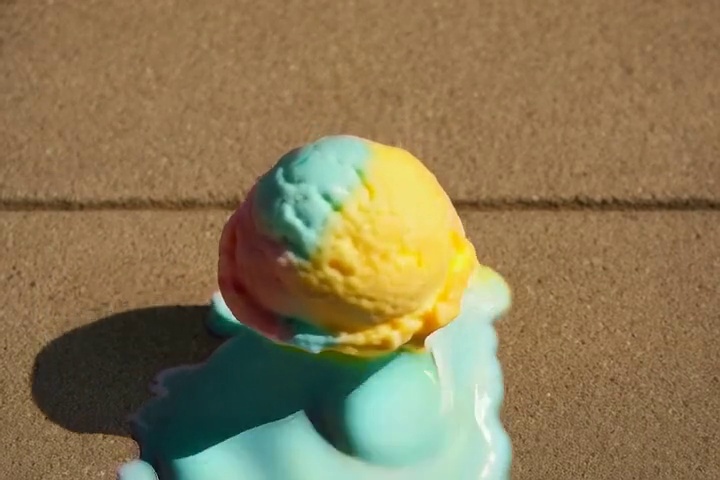} \\
\includegraphics[width=0.24\textwidth]{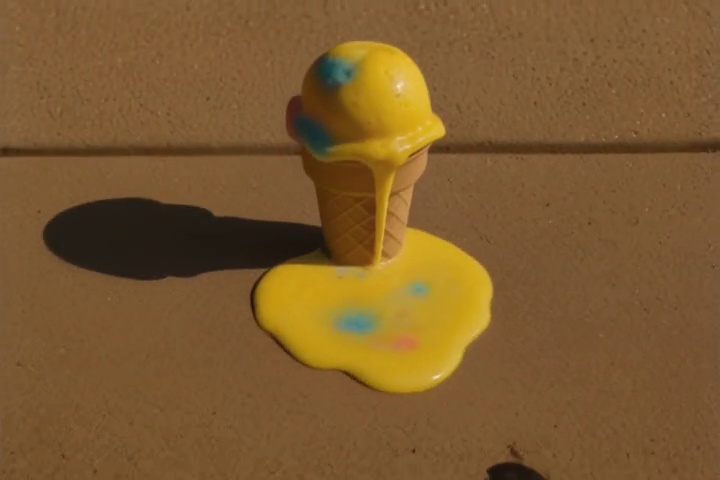} &
\includegraphics[width=0.24\textwidth]{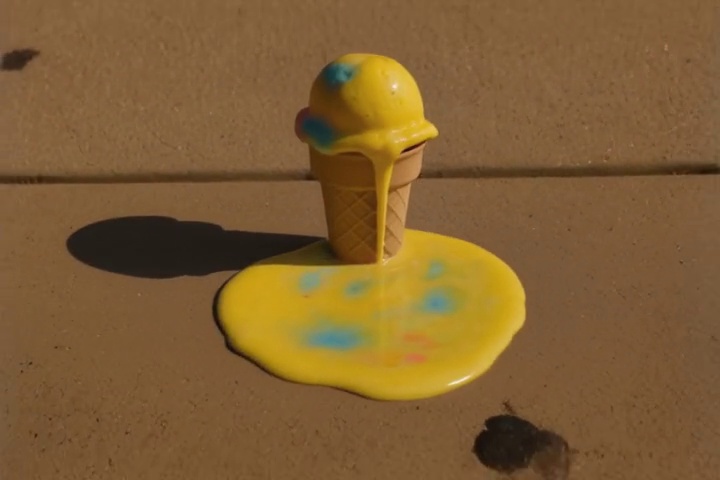} &
\includegraphics[width=0.24\textwidth]{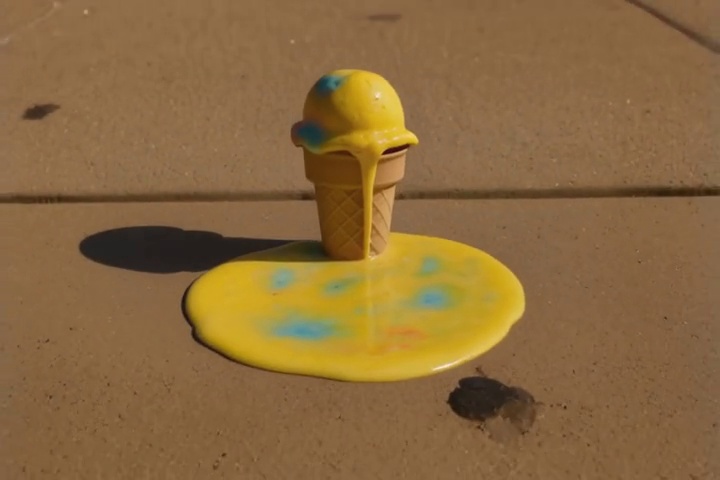} &
\includegraphics[width=0.24\textwidth]{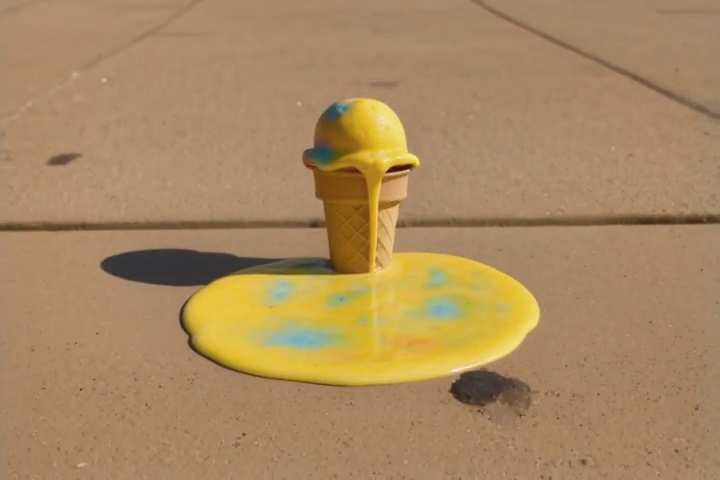} \\
\end{tabular}
\end{minipage}

\caption{Qualitative comparisons on physical video generation tasks. \textbf{Left:} liquid interaction scenario (milk poured into coffee). \textbf{Right:} thermal state transition scenario (ice cream melting in the sun). (a) Wan2.1-T2V-1.3B; (b) LTX-Video; (c) VLIPP; (d) CausalMotion (ours). Our method produces more physically consistent object interactions and state transitions across both scenarios.}
\label{fig:qualitative_results_combined}
\end{figure*}

Figure~\ref{fig:qualitative_results_combined} presents qualitative comparisons on two representative physical video generation scenarios: liquid interaction (\textit{left}) and thermal state transition (\textit{right}).

In the liquid interaction example, existing methods often generate physically implausible behaviors. LTX-Video produces semantically correct scenes but violates basic physical constraints, such as the coffee liquid surface rising outside the cup. Wan2.1-T2V-1.3B captures the pouring action but fails to model clear liquid state transitions. VLIPP generates unrealistic liquid accumulation above the cup instead of naturally flowing along the cup boundary. In contrast, our method explicitly models object motion and interactions, producing continuous and physically plausible transitions in both liquid height and color blending.

In the thermal transition example, both Wan2.1-T2V-1.3B and LTX-Video struggle to capture realistic melting dynamics, showing limited state changes over time. Although VLIPP depicts melted ice cream accumulating on the ground, the upper scoop remains largely unchanged and fails to exhibit realistic melting behavior. Our method generates a gradual and consistent melting process that better reflects real-world physical dynamics.

\subsection{Ablation studies}

To understand the contribution of each component in our physical grounding framework, we conduct ablation studies by progressively removing key modules: (i) keyframe reasoning, (ii) trajectory-aware physical state planning, and (iii) latent trajectory guidance. Results are reported in Table~\ref{tab:ablation}.

\textbf{Effect of trajectory-aware physical state planning.}
When removing trajectory physical state modeling, we drop trajectory planning and interpolate keyframes uniformly, and performance drops moderately.
We observe consistent degradation in mechanics and material, suggesting that explicit modeling of motion dynamics and object interactions further refines physical realism. Performance on thermal slightly improves, which can be attributed to reduced constraints allowing the generative model to better match appearance-based cues, albeit at the cost of physical consistency.

\textbf{Effect of latent trajectory guidance.}
While keyframe alignment still ensures coarse temporal consistency, the absence of continuous trajectory constraints leads to weaker frame-to-frame coherence. This is particularly evident in optics and mechanics, where fine-grained motion consistency is crucial. We observe a slight improvement in material, suggesting that strong guidance may occasionally over-constrain appearance variation.

The full model achieves slightly lower scores in material scenarios. We hypothesize that these categories often involve gradual appearance changes rather than significant spatial motion or geometric transformations. For example, in prompts such as \textit{“A timelapse records the oxidation process of a bright and new metal bedframe kept in a damp location over several years,”} the object remains largely stationary while its appearance progressively changes over time. In such cases, simple temporal interpolation between keyframes is often sufficient to produce smooth transitions, and additional trajectory-based latent constraints may introduce unnecessary restrictions. 

\begin{table}[htbp]
\centering
\begin{tabular}{lccccc}
\toprule
\textbf{Model Variant} & \textbf{Mechanics $\uparrow$} & \textbf{Optics $\uparrow$} & \textbf{Thermal$\uparrow$} & \textbf{Material$\uparrow$} & \textbf{Average$\uparrow$} \\
\midrule
Full model  & \best{0.608} & \best{0.713} & 0.678 & 0.608 &  \best{0.654}\\
-w/o keyframe reasoning          & 0.425          & 0.600          & 0.467          & 0.358 & 0.471\\
-w/o trajectory physical state      & 0.583          & 0.700          & \best{0.689}         & 0.592 & 0.642\\
-w/o latent trajectory guidance          & 0.594          & 0.680          & 0.678          & \best{0.625} & 0.645\\
\bottomrule
\end{tabular}
\caption{Quantitative results for ablation study on PhyGenbench}
\label{tab:ablation}
\vspace{-4mm}
\end{table}

\begin{figure}[htbp]
  \centering
  \begin{subfigure}[b]{0.48\textwidth}
    \centering
    \vspace{1mm}
    \includegraphics[width=0.18\textwidth]{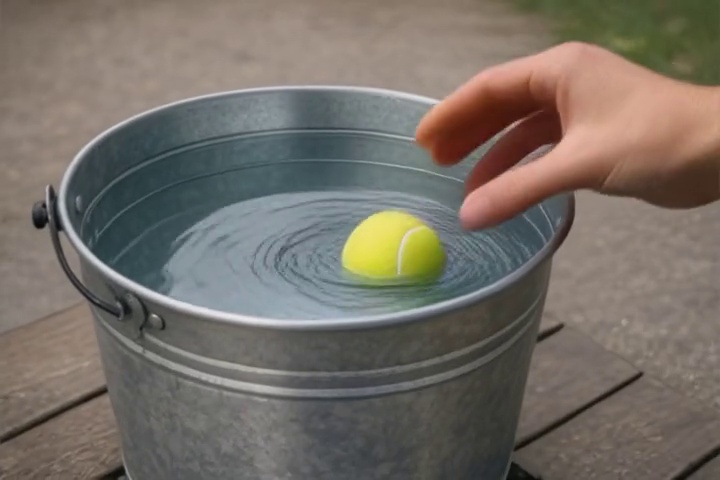}%
    \includegraphics[width=0.18\textwidth]{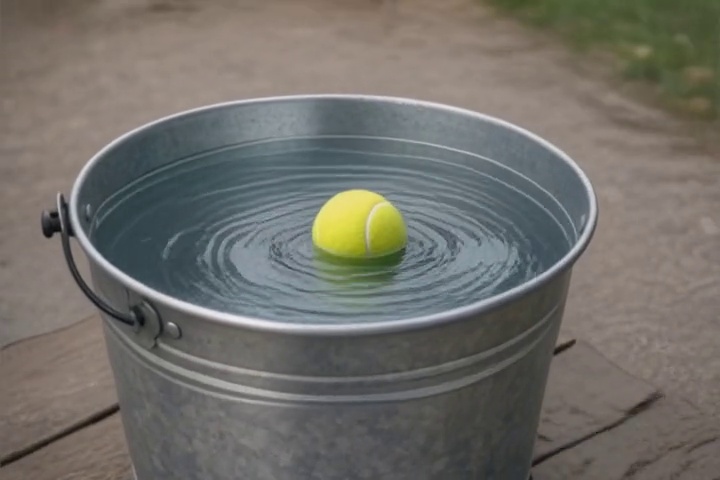}%
    \includegraphics[width=0.18\textwidth]{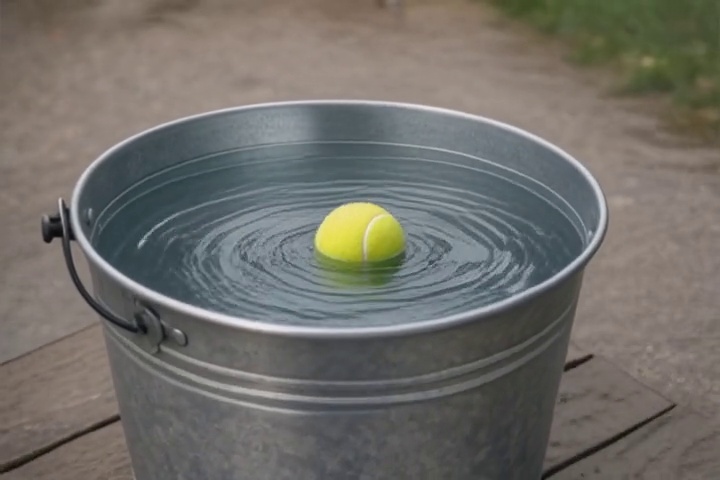}%
    \includegraphics[width=0.18\textwidth]{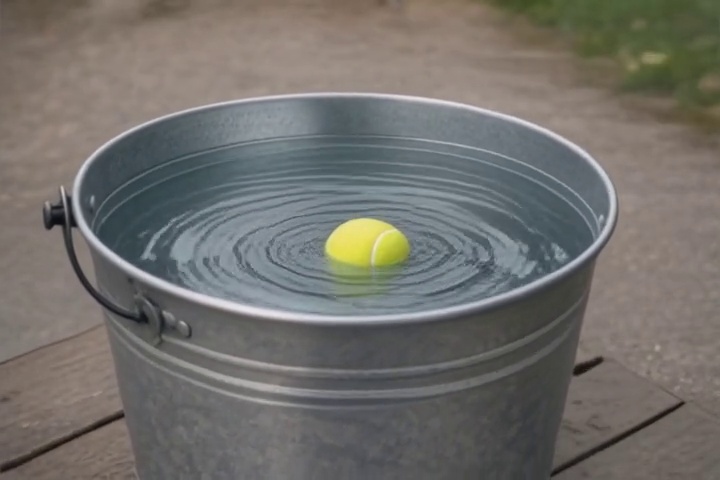}%
    \includegraphics[width=0.18\textwidth]{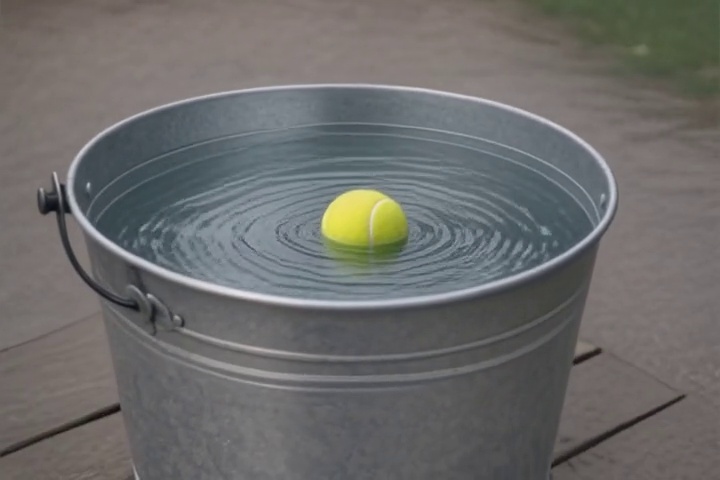}%
    
    \textit{Full model} \\
    \vspace{2mm}
    \includegraphics[width=0.18\textwidth]{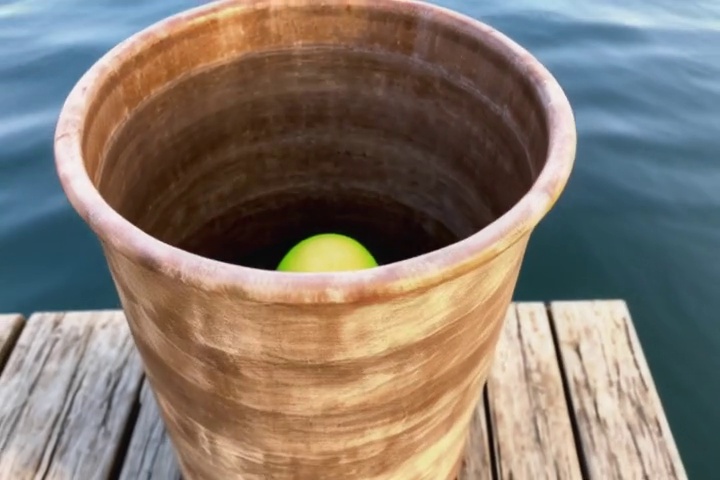}%
    \includegraphics[width=0.18\textwidth]{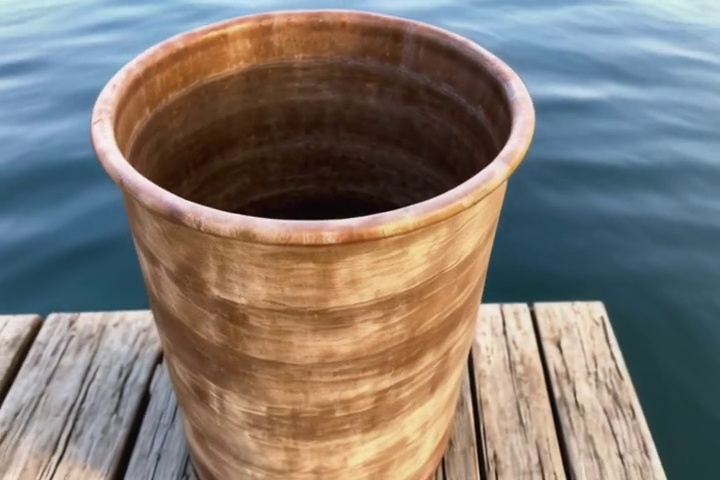}%
    \includegraphics[width=0.18\textwidth]{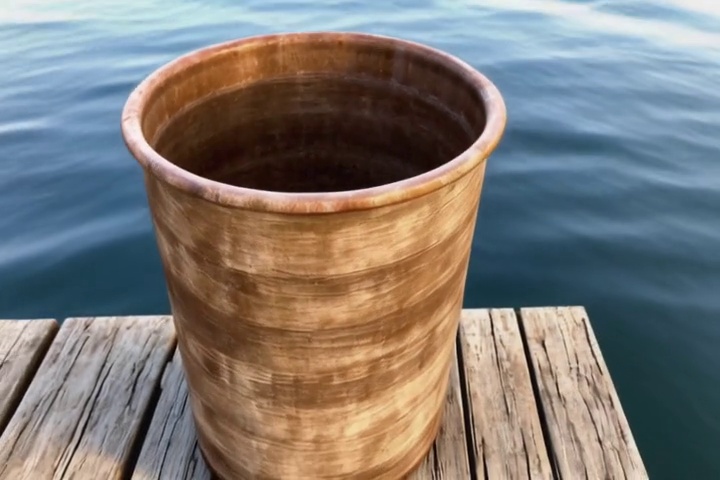}%
    \includegraphics[width=0.18\textwidth]{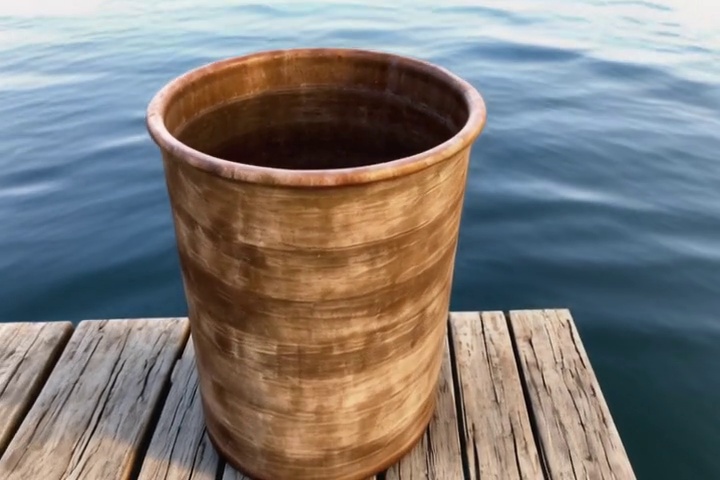}%
    \includegraphics[width=0.18\textwidth]{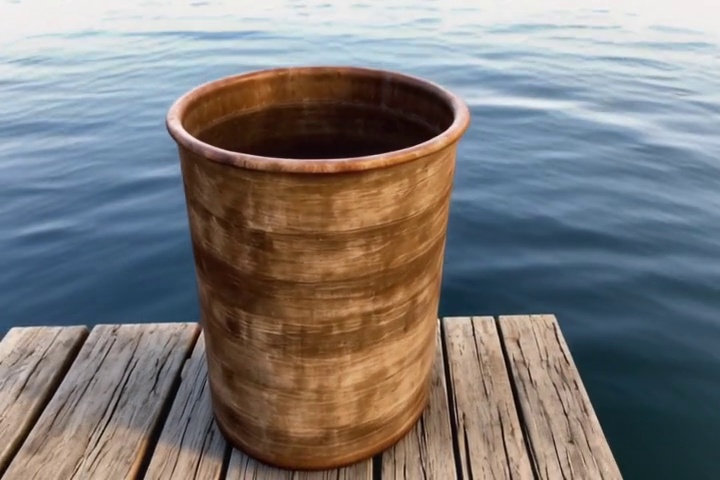}%
    
    \textit{w/o keyframe reasoning} \\
    \subcaption{A tennis ball is gently placed on the surface of a bucket filled with water.\\ \textcolor{blue!40}{(Force)}}
  \end{subfigure}
  \hfill
  \begin{subfigure}[b]{0.48\textwidth}
    \centering
    \vspace{1mm}
    \includegraphics[width=0.18\textwidth]{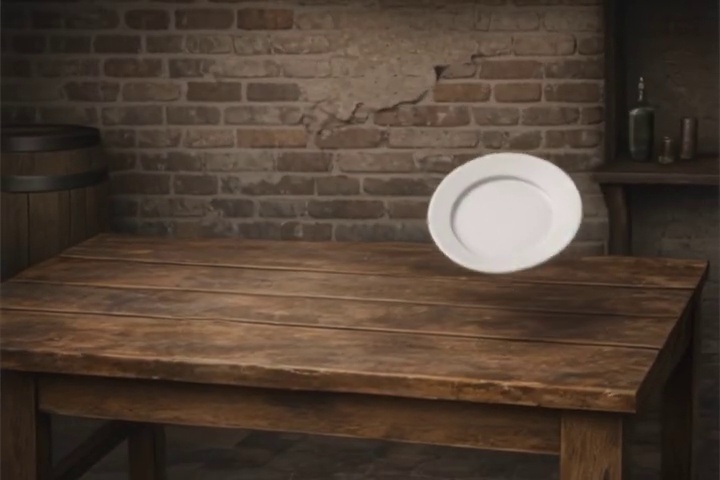}%
    \includegraphics[width=0.18\textwidth]{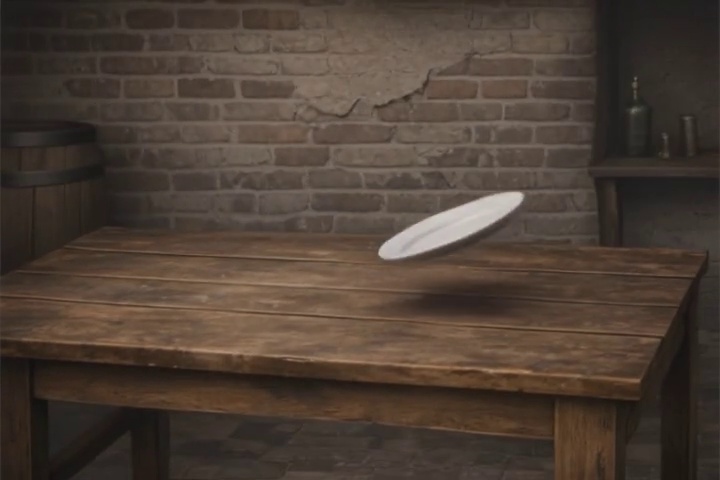}%
    \includegraphics[width=0.18\textwidth]{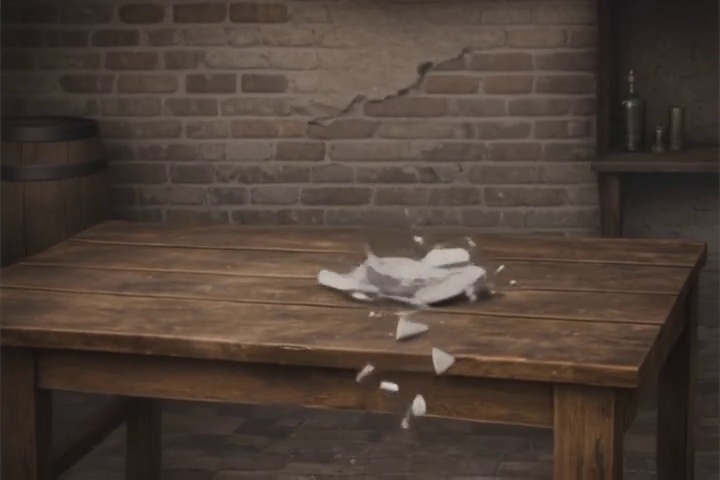}%
    \includegraphics[width=0.18\textwidth]{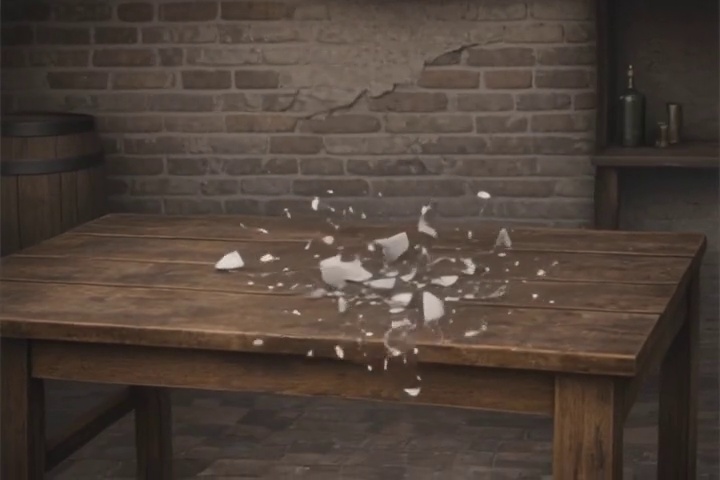}%
    \includegraphics[width=0.18\textwidth]{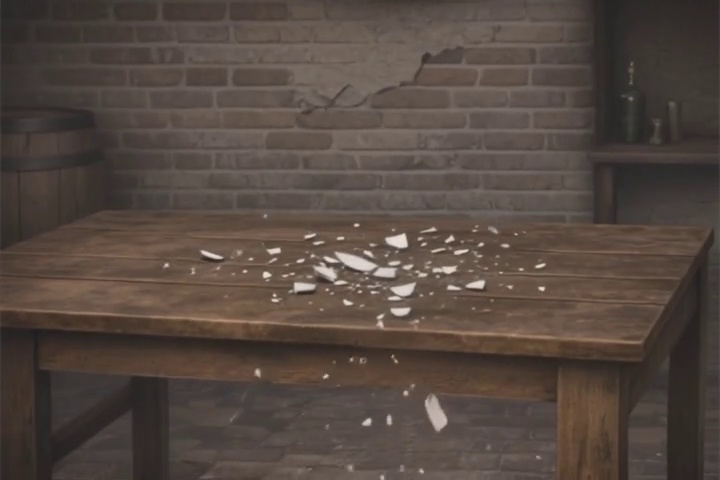}%
    
    \textit{Full model} \\
    \vspace{2mm}
    \includegraphics[width=0.18\textwidth]{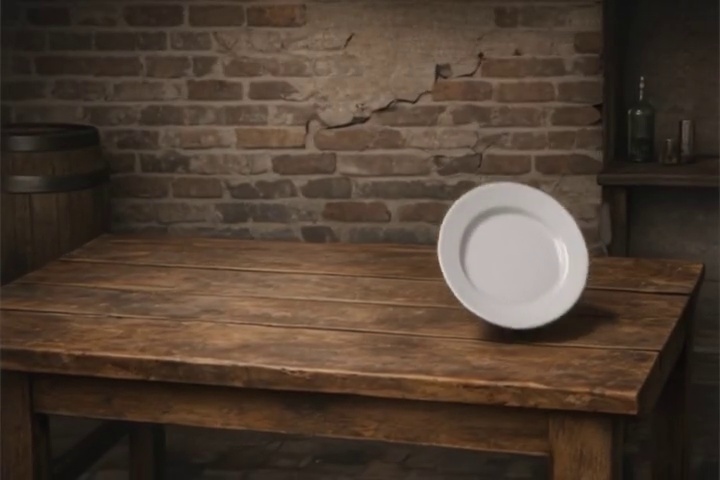}%
    \includegraphics[width=0.18\textwidth]{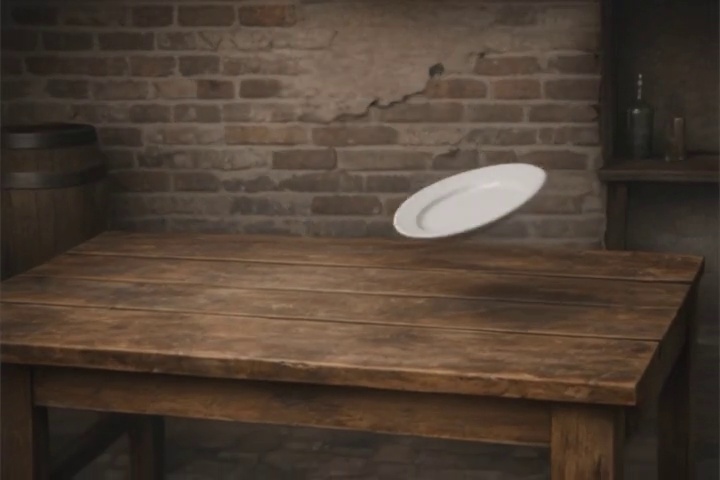}%
    \includegraphics[width=0.18\textwidth]{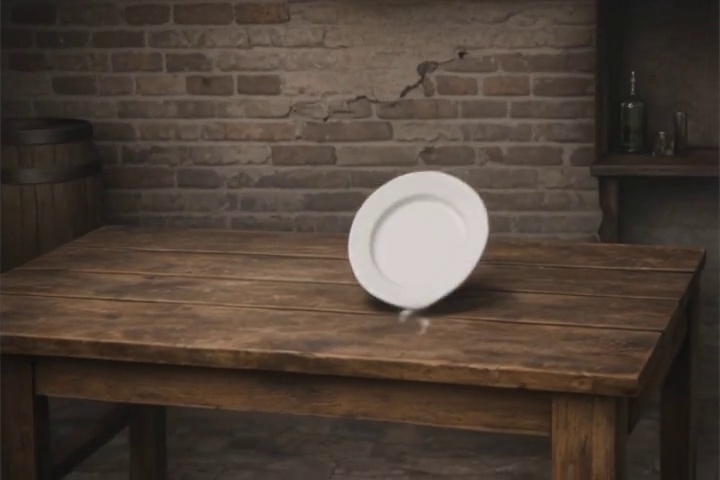}%
    \includegraphics[width=0.18\textwidth]{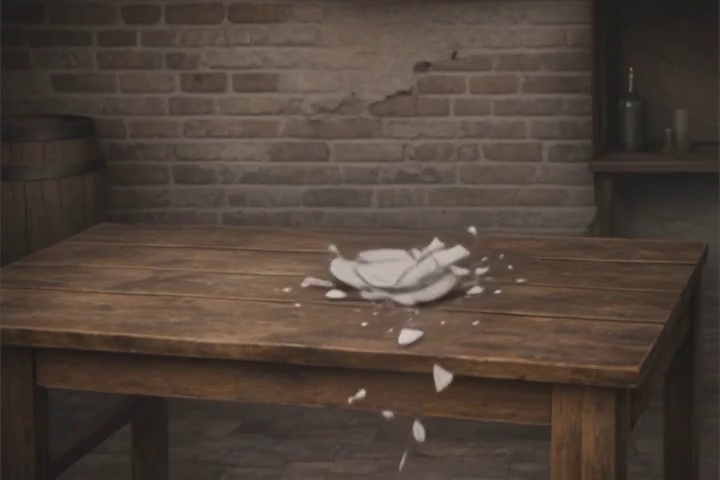}%
    \includegraphics[width=0.18\textwidth]{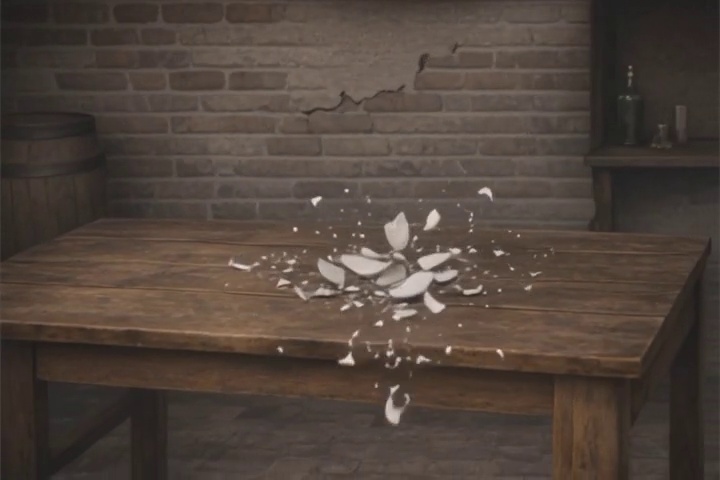}%
    
    \textit{w/o trajectory physical state} \\
    \subcaption{A weak, frail porcelain plate is flung with significant speed at a robust, wooden table, where it collides upon impact. \textcolor{blue!40}{(Physical Properties)}}
  \end{subfigure}
  
  
  \begin{subfigure}[b]{0.48\textwidth}
    \centering
    \vspace{1mm}
    \includegraphics[width=0.18\textwidth]{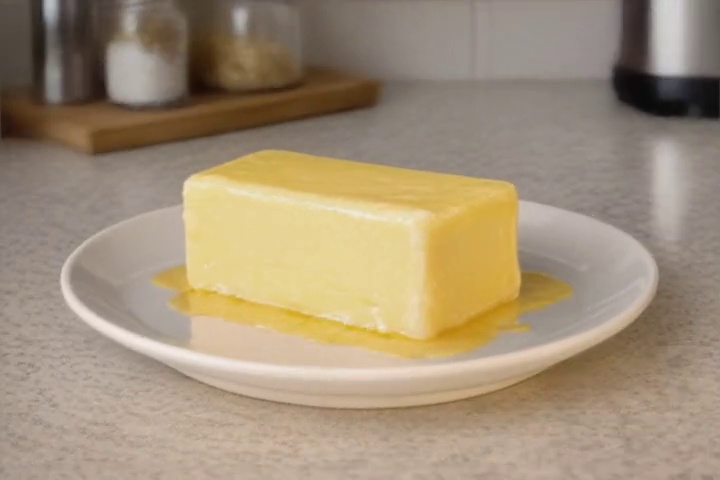}%
    \includegraphics[width=0.18\textwidth]{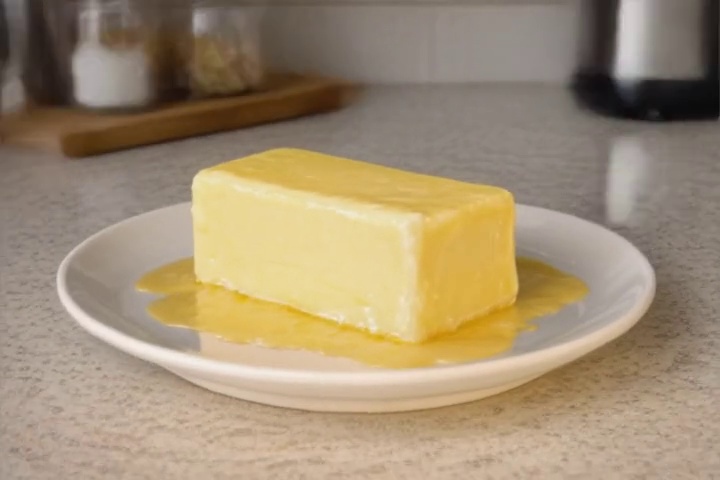}%
    \includegraphics[width=0.18\textwidth]{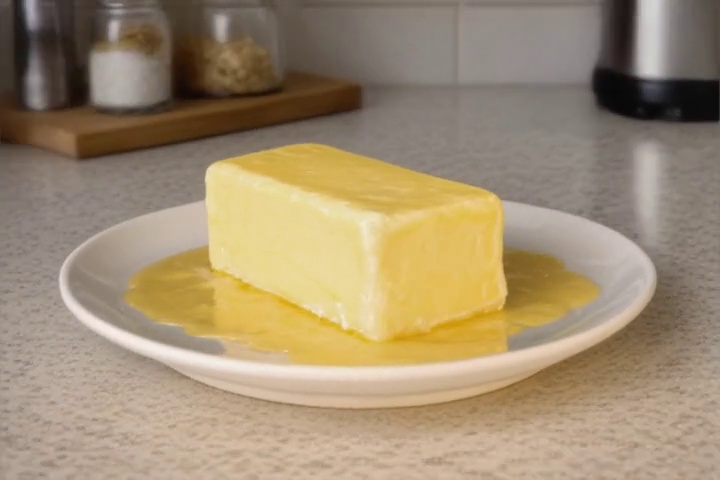}%
    \includegraphics[width=0.18\textwidth]{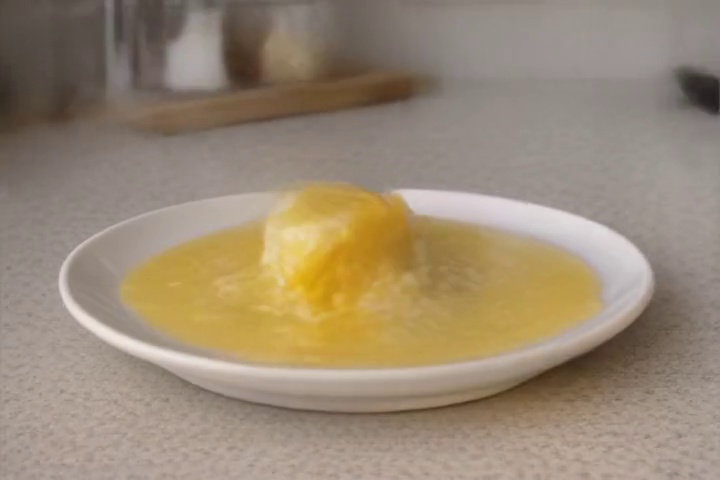}%
    \includegraphics[width=0.18\textwidth]{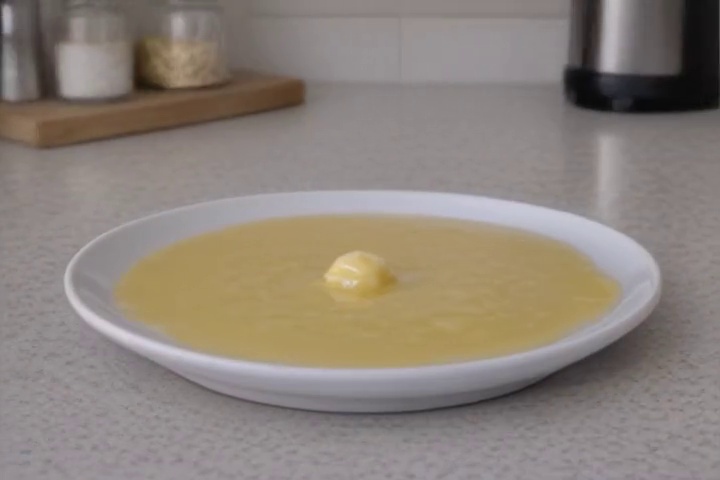}%
    
    \textit{Full model} \\
    \vspace{2mm}
    
    \includegraphics[width=0.18\textwidth]{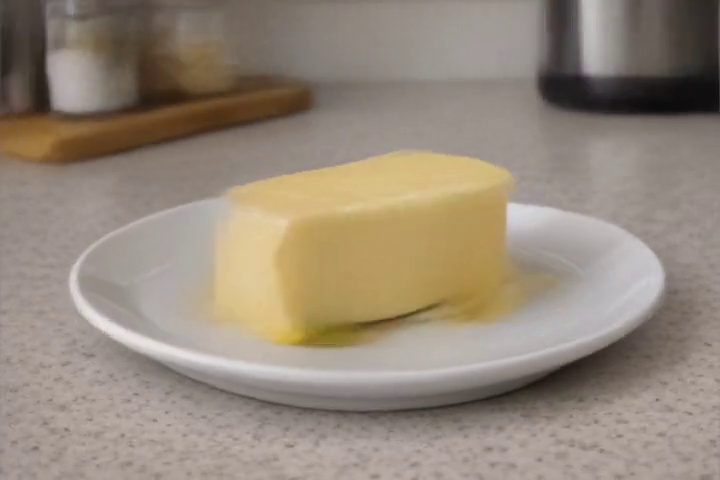}%
    \includegraphics[width=0.18\textwidth]{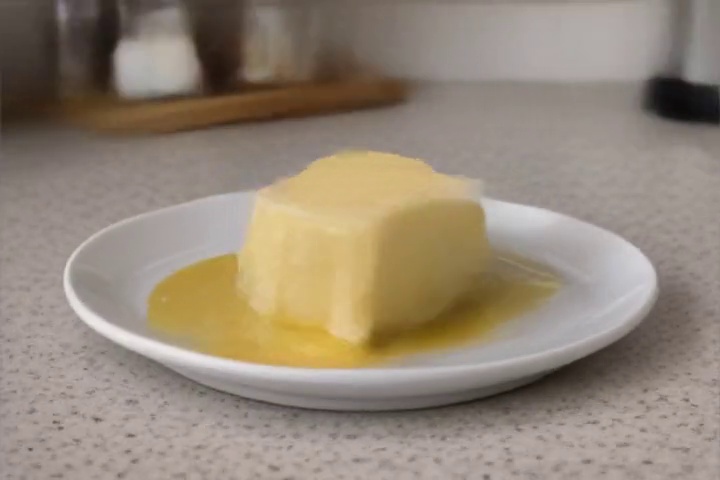}%
    \includegraphics[width=0.18\textwidth]{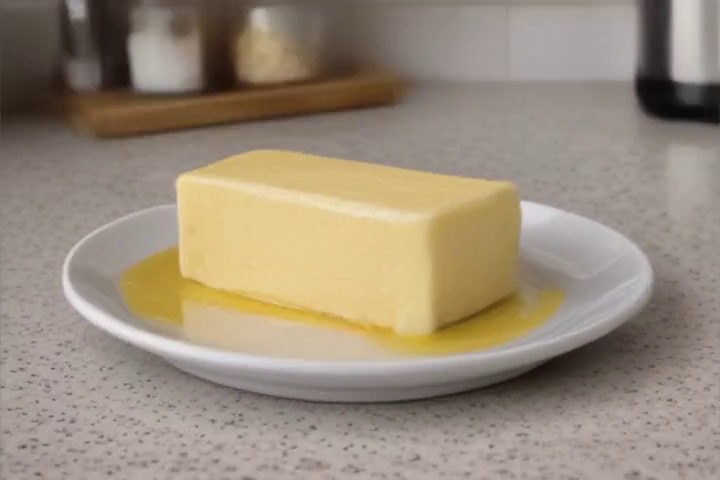}%
    \includegraphics[width=0.18\textwidth]{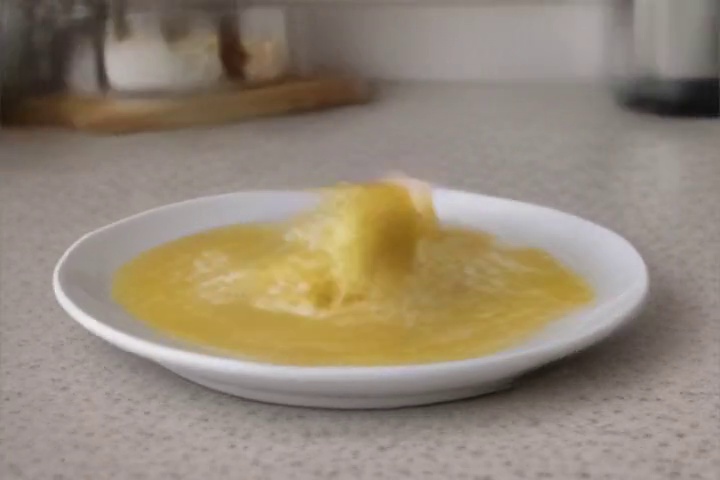}%
    \includegraphics[width=0.18\textwidth]{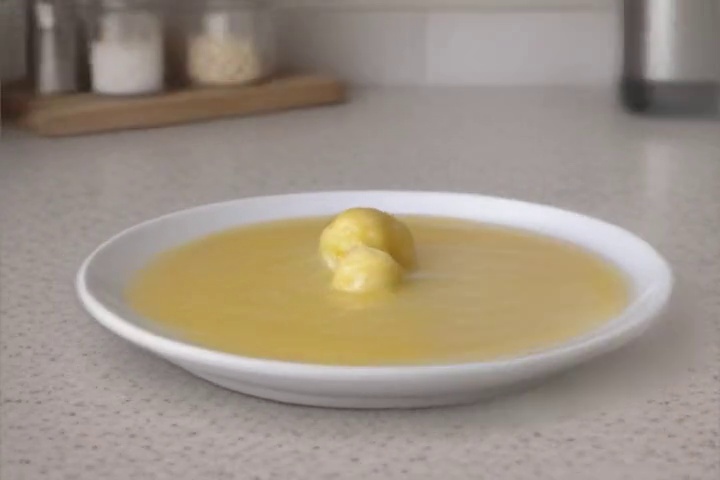}%
    
    \textit{w/o latent guidance} \\
    \subcaption{A timelapse captures the gradual transformation of butter as the temperature rises significantly. \\\textcolor{blue!40}{(Heat)}}
  \end{subfigure}
  \hfill
  \begin{subfigure}[b]{0.48\textwidth}
    \centering
    \vspace{1mm}
    \includegraphics[width=0.18\textwidth]{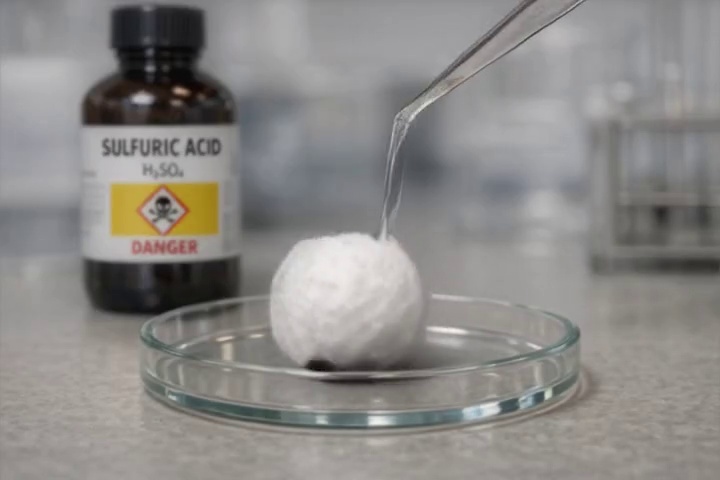}%
    \includegraphics[width=0.18\textwidth]{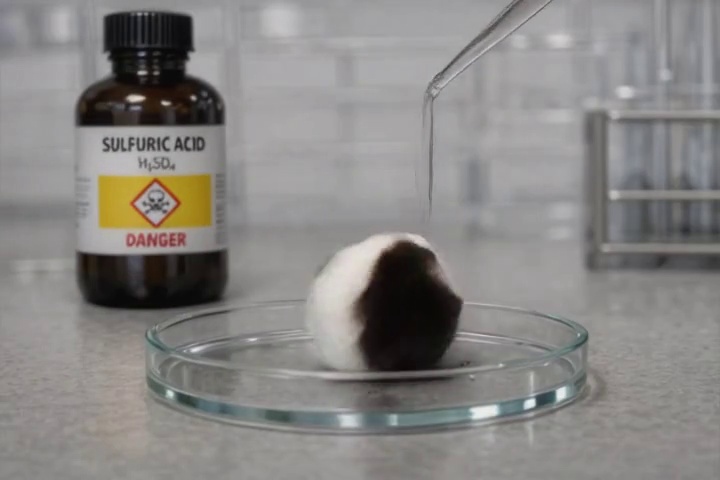}%
    \includegraphics[width=0.18\textwidth]{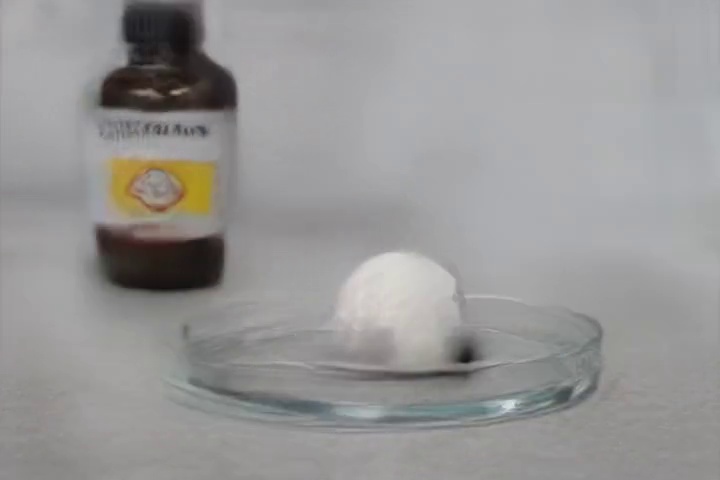}%
    \includegraphics[width=0.18\textwidth]{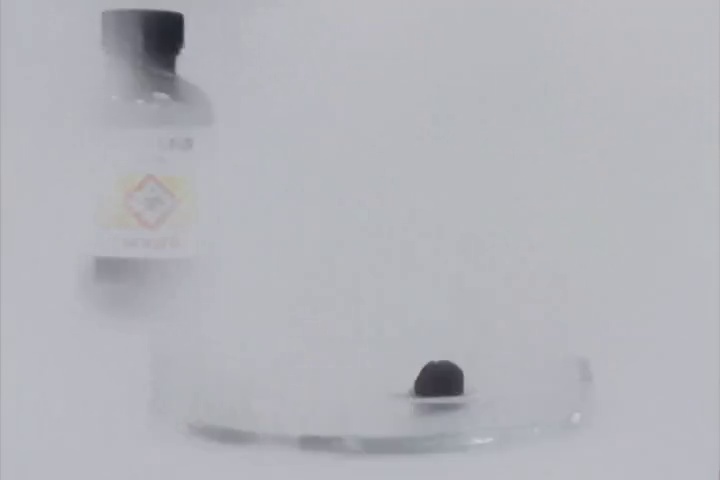}%
    \includegraphics[width=0.18\textwidth]{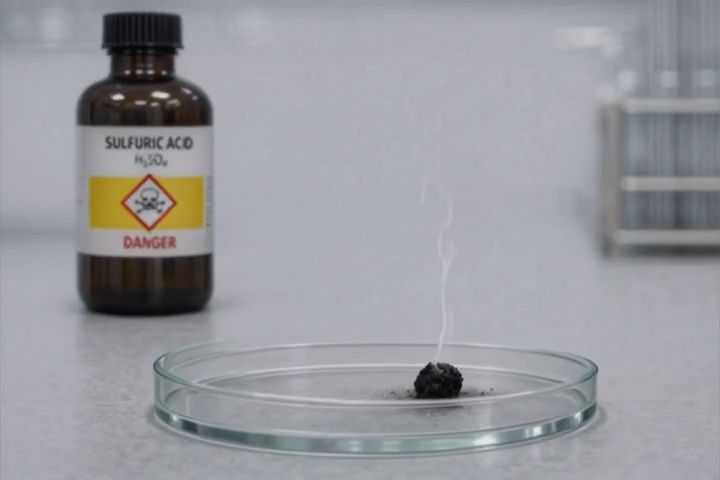}%
    
    \textit{Full model} \\
    \vspace{2mm}
    
    \includegraphics[width=0.18\textwidth]{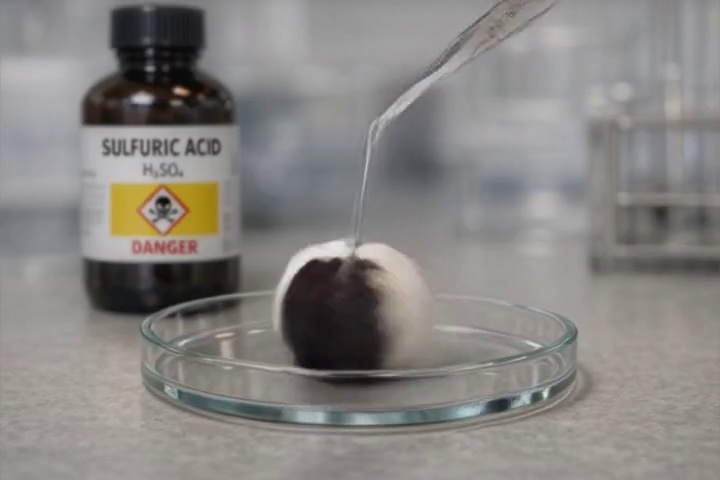}%
    \includegraphics[width=0.18\textwidth]{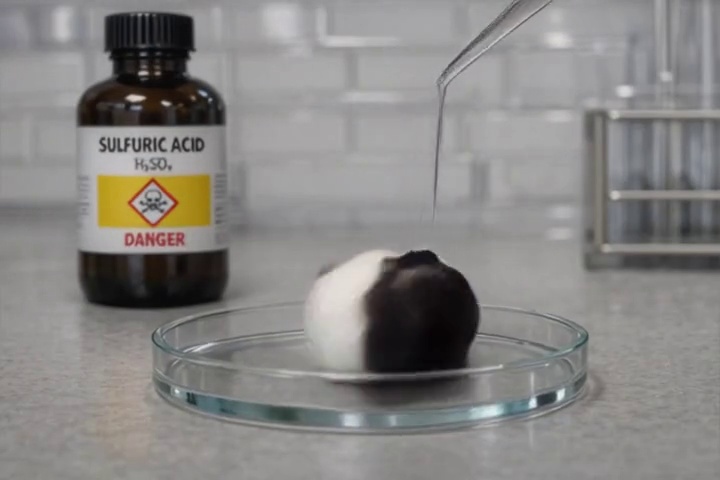}%
    \includegraphics[width=0.18\textwidth]{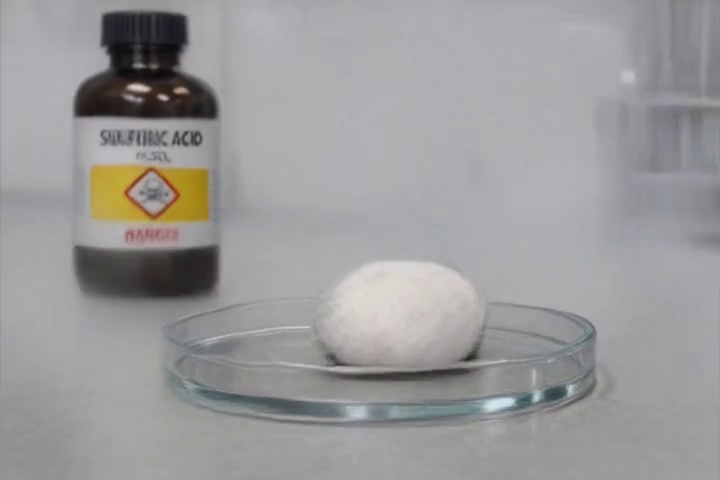}%
    \includegraphics[width=0.18\textwidth]{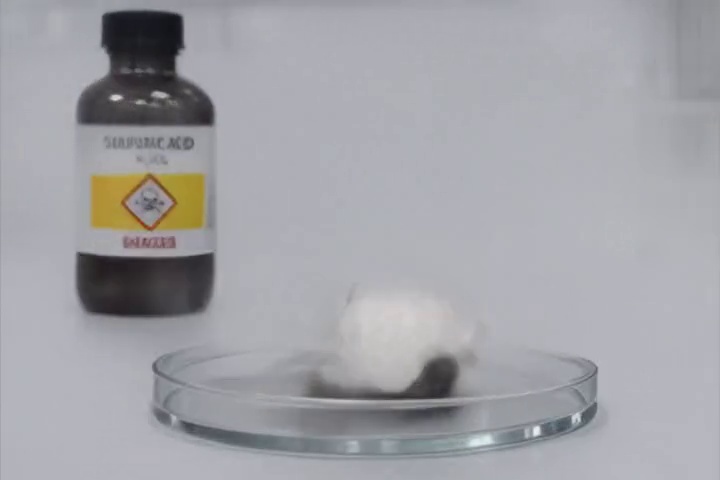}%
    \includegraphics[width=0.18\textwidth]{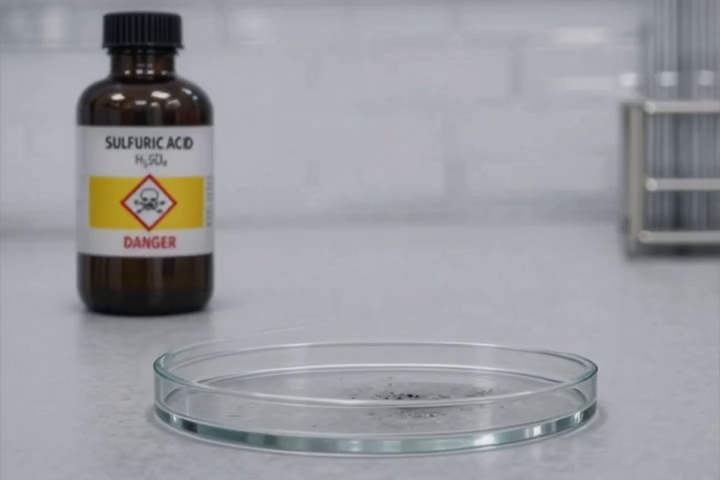}%
    
    \textit{w/o latent guidance} \\
    \subcaption{A timelapse captures the reaction as concentrated sulfuric acid is poured onto a cotton ball. \\ \textcolor{blue!40}{(Chemical Properties)}}
  \end{subfigure}
  
  \caption{
    Qualitative ablation results. Removing keyframe reasoning (a) leads to missing or incorrect rendering of the tennis ball. Removing trajectory physical state (b) results in unnatural collision dynamics of the plate. Disabling latent trajectory guidance (c) makes butter melting less smooth, and (d) causes the cotton ball to vanish abruptly instead of showing a chemical reaction. The full model maintains gradual transformations and physical consistency across all cases.
  }
  \label{fig:ablation_qualitative}
  \vspace{-4mm}
\end{figure}

Figure~\ref{fig:ablation_qualitative} further illustrates the impact of each component on visual quality and physical consistency. Removing keyframe reasoning leads to incorrect or missing intermediate states, indicating weak causal grounding. Without trajectory-aware physical state planning, object interactions become less realistic, particularly in collision scenarios. Disabling latent trajectory guidance mainly affects temporal smoothness, resulting in less coherent transitions across frames. In contrast, the full model consistently produces gradual, physically plausible dynamics, demonstrating that each component contributes to different aspects of the generation process.

\section{Conclusion}
In this work, we proposed CausalMotion, a training-free physical grounding framework that integrates VLM reasoning ability into video generation. By decoupling causal reasoning from visual synthesis, our approach introduces structured intermediate representations and physically grounded trajectories to guide diffusion models toward more coherent and physically plausible outputs. Extensive experiments demonstrate that our method consistently improves physical consistency across diverse scenarios, particularly in mechanics and long-horizon dynamics, while preserving or even enhancing perceptual quality. Our method suggests a new paradigm for video generation, opening up a flexible and scalable direction for integrating higher-level cognition into generative models.

\appendix

\section{Discussion of Trajectory Mapping}
\label{sec:discussion_traj_mapping}
In our framework, keyframe generation (Section~\ref{sec:visual_thought_reasoning}) and trajectory reasoning (Section~\ref{sec:trajectory_reasoning}) are performed in different representational spaces. Keyframes are generated through VLM-based visual reasoning, while trajectory planning operates on object-centric spatial representations initialized from the first frame $\mathit{img}_0$. As a result, the two processes are not explicitly aligned in either coordinate system or temporal parametrization.

\begin{figure}[htbp]
\centering
\begin{minipage}{0.48\linewidth}
    \centering
    \includegraphics[width=\linewidth]{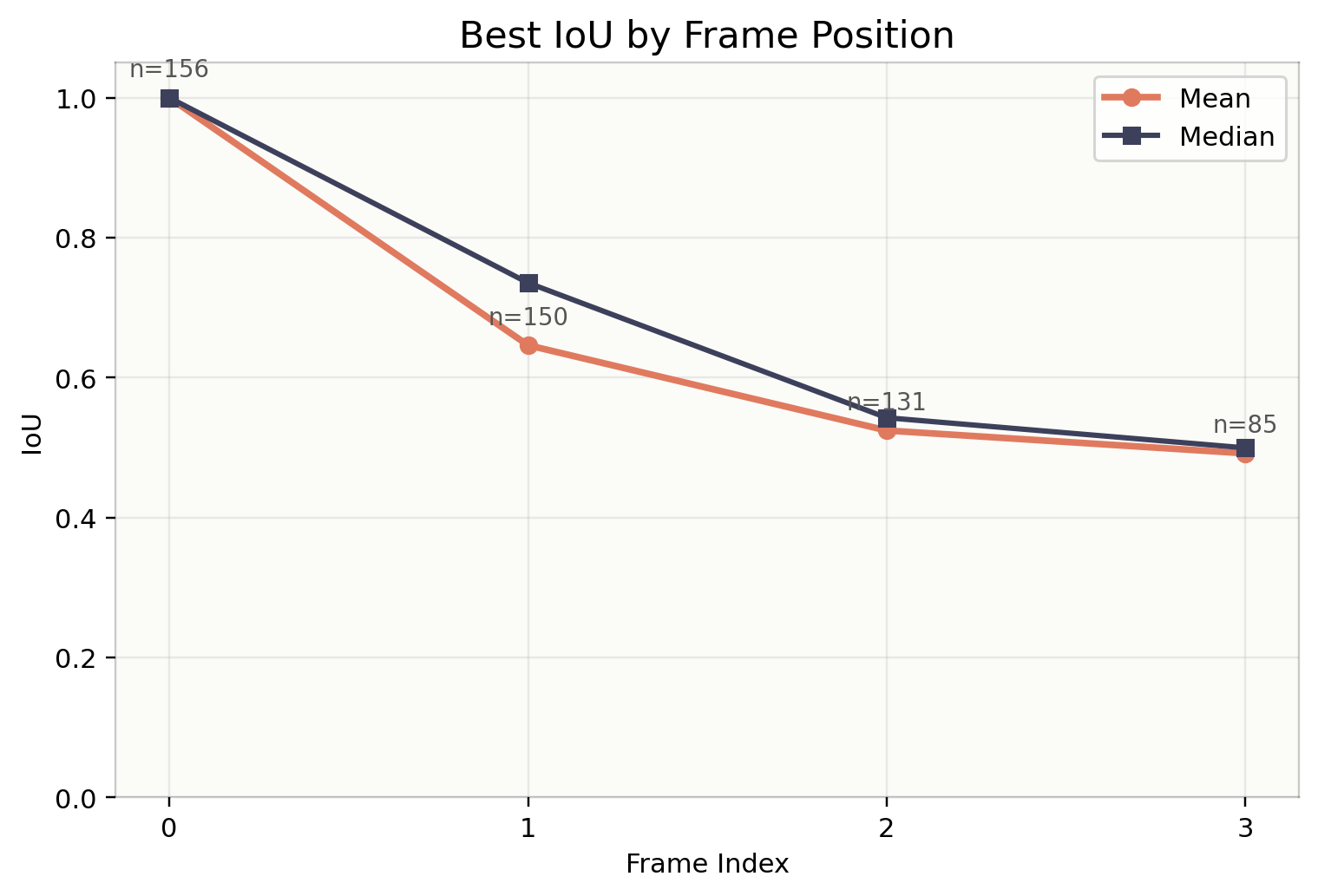}
    {\small (a) Mean frame-wise IoU over time across all samples.}
\end{minipage}
\hfill
\begin{minipage}{0.48\linewidth}
    \centering
    \includegraphics[width=\linewidth]{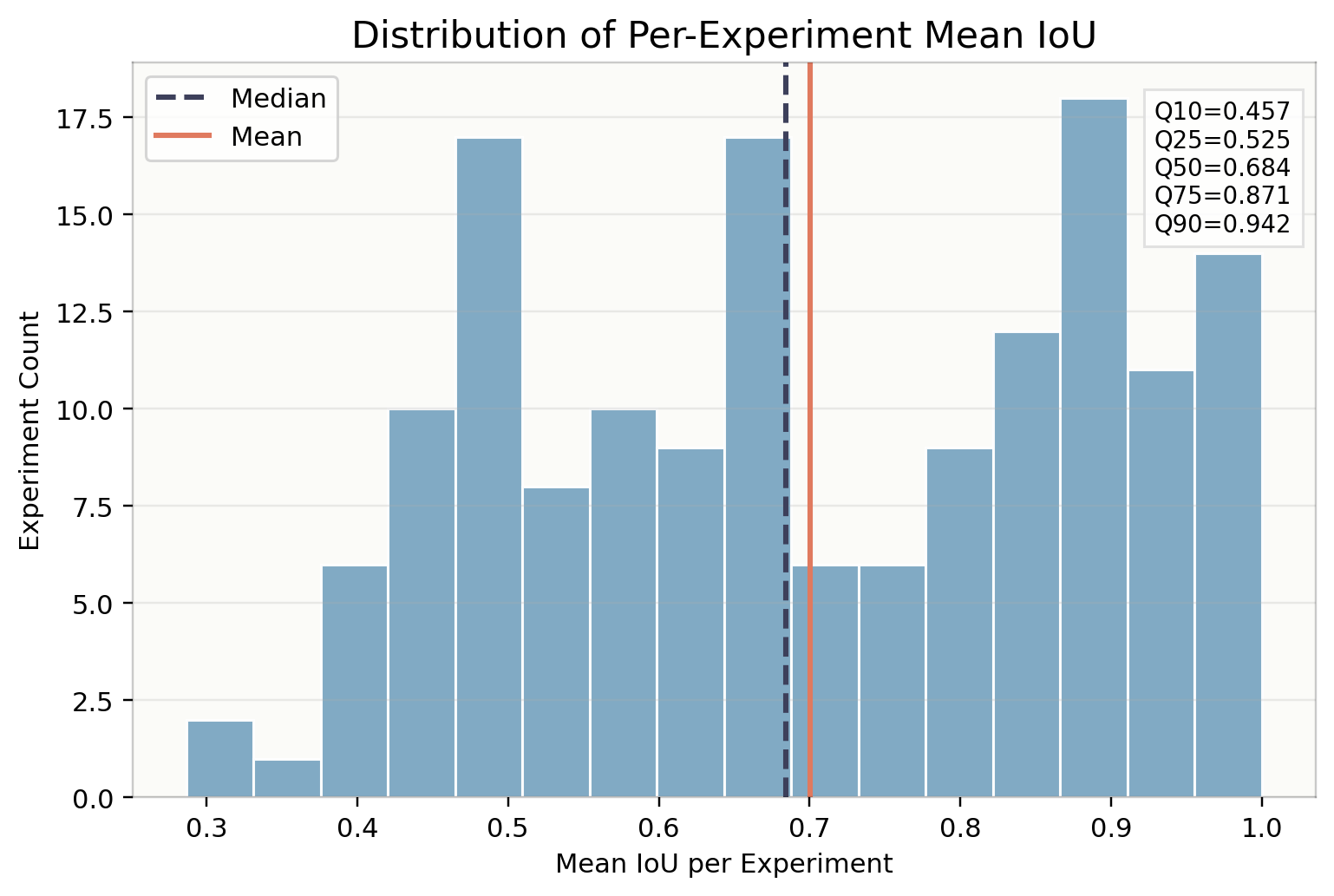}
    {\small (b) Distribution of per-experiment mean IoU.}
\end{minipage}
\caption{Alignment analysis between trajectory predictions and keyframe object states. (a) shows frame-wise IoU degradation over time, while (b) illustrates variance across different prompts.}
\label{fig:mapping_iou_main}
\end{figure}

To analyze the consistency between these two modules, we compute the Intersection-over-Union (IoU) between object bounding boxes from trajectory prediction and those extracted from generated keyframes during temporal alignment (Section~\ref{sec:time_align}). The frame-wise IoU statistics are shown in Figure~\ref{fig:mapping_iou_main}(a).

Despite being generated independently and potentially under different implicit coordinate systems, we observe that the predicted trajectories and keyframe object states exhibit consistent motion trends. In particular, we observe reasonably consistent spatial overlap trends across both representations, indicating that the VLM reasoning process captures coherent physical dynamics even without explicit spatial supervision.

Figure~\ref{fig:mapping_iou_main}(b) presents the distribution of per-experiment mean IoU values. While the overall average IoU remains relatively high, the distribution shows noticeable variance across prompts. This indicates that alignment quality depends on scene complexity, motion patterns, and the degree of object interaction.

Overall, our analysis shows that although the two modules operate independently, they converge to consistent motion dynamics in most cases, with failures primarily occurring in long-horizon or complex interaction scenarios.

\section{Time Cost}
Table~\ref{tab:time_cost} shows the average runtime of each stage in our method, providing a detailed breakdown of the overall computational cost.

\begin{table}[htbp]
    \centering
    \small
    \caption{Average runtime breakdown of different components in our framework.}
    \begin{tabularx}{\textwidth}{c c X}
        \toprule
        \textbf{Breakdown} & \textbf{Time Cost} & \textbf{Comments} \\
        \midrule
        Keyframe Generation & 271 sec & GPT-Image-1.5 API inference calls \\
        Physical State and Trajectory Planning & 69 sec & Qwen2.5-VL-72B API inference calls \\
        Trajectory-Guided Latent-Space Guidance & 108 sec & LTX-Video inference on 1 × NVIDIA A800 GPU \\
        \midrule
        Total & 448 sec & \\
        \bottomrule
    \end{tabularx}
    \label{tab:time_cost}
\end{table}

In total, our method requires approximately 448 seconds per sample. The main computational bottleneck comes from API-based keyframe generation rather than the trajectory guidance module itself. We note that this overhead can be further reduced by replacing external API calls with locally deployed VLMs or more efficient keyframe generation strategies.

\section{Intermediate Visualization}
\label{sec:supp_intermediate_visualization}

To improve the interpretability of our pipeline, we provide intermediate visualizations for representative examples, including generated keyframes, trajectory alignment, and final synthesis results.

\begin{figure}[htbp]
\centering
\setlength{\tabcolsep}{0pt}  
\renewcommand{\arraystretch}{0}  
\begin{tabular}{cc}
\includegraphics[width=0.5\linewidth, keepaspectratio]{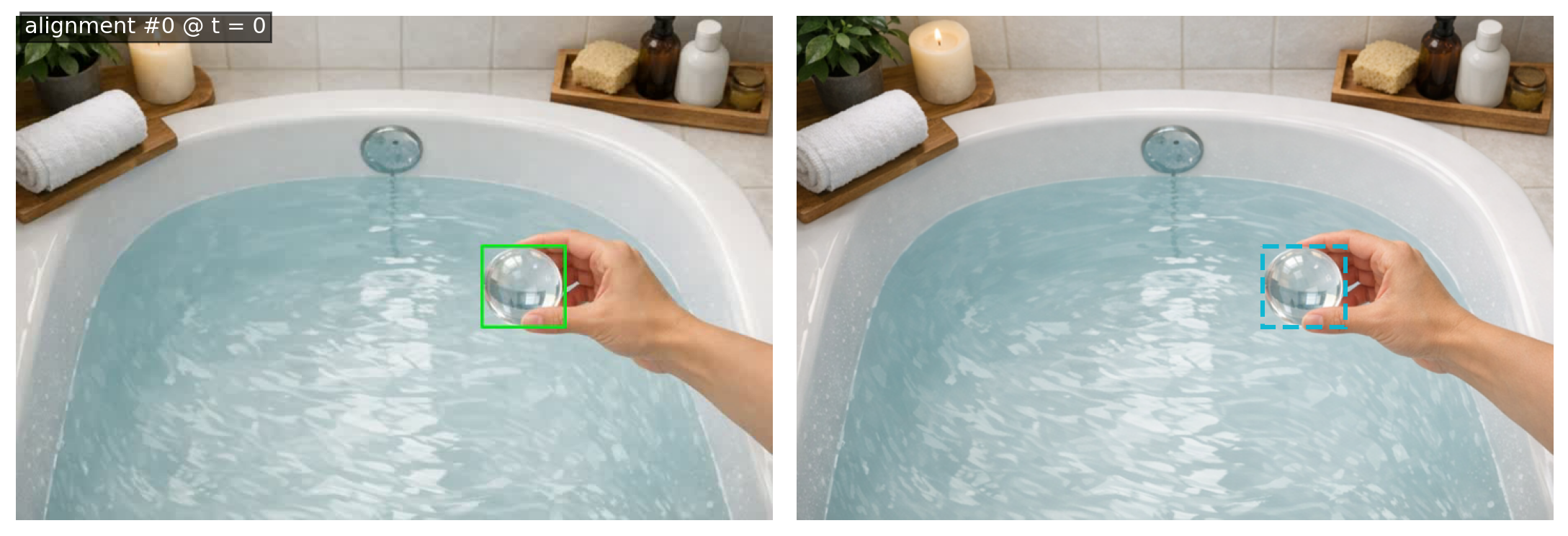} &
\includegraphics[width=0.5\linewidth, keepaspectratio]{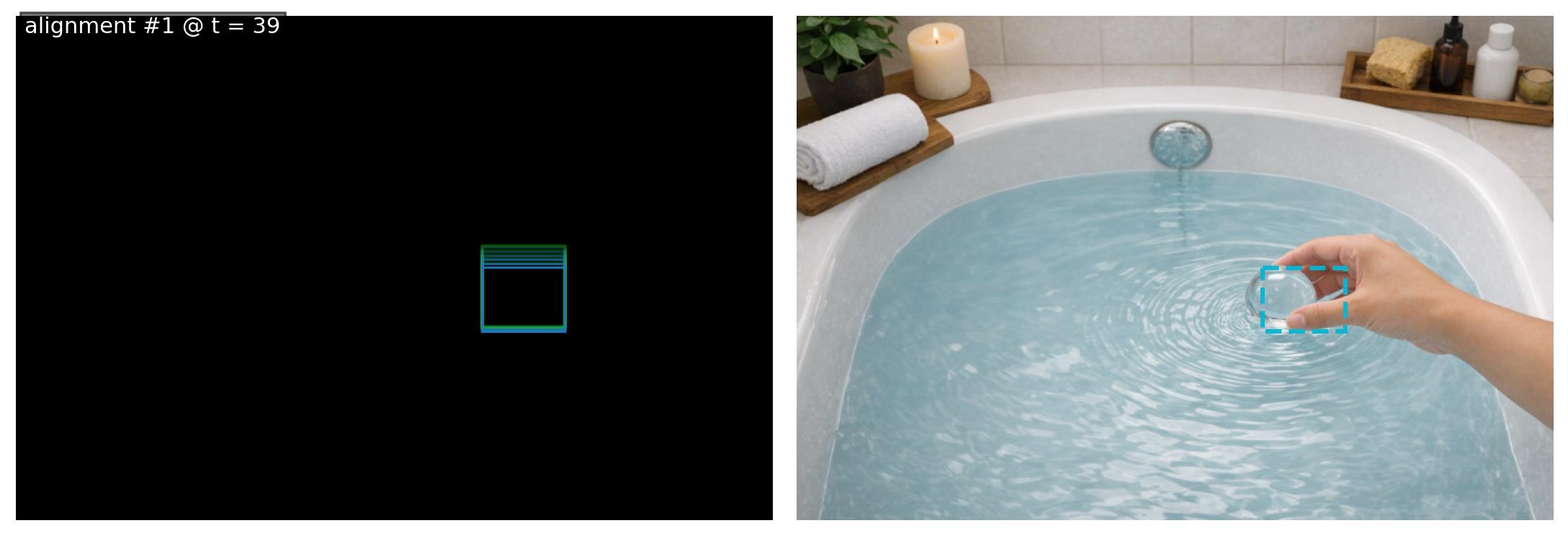} \\
\includegraphics[width=0.5\linewidth, keepaspectratio]{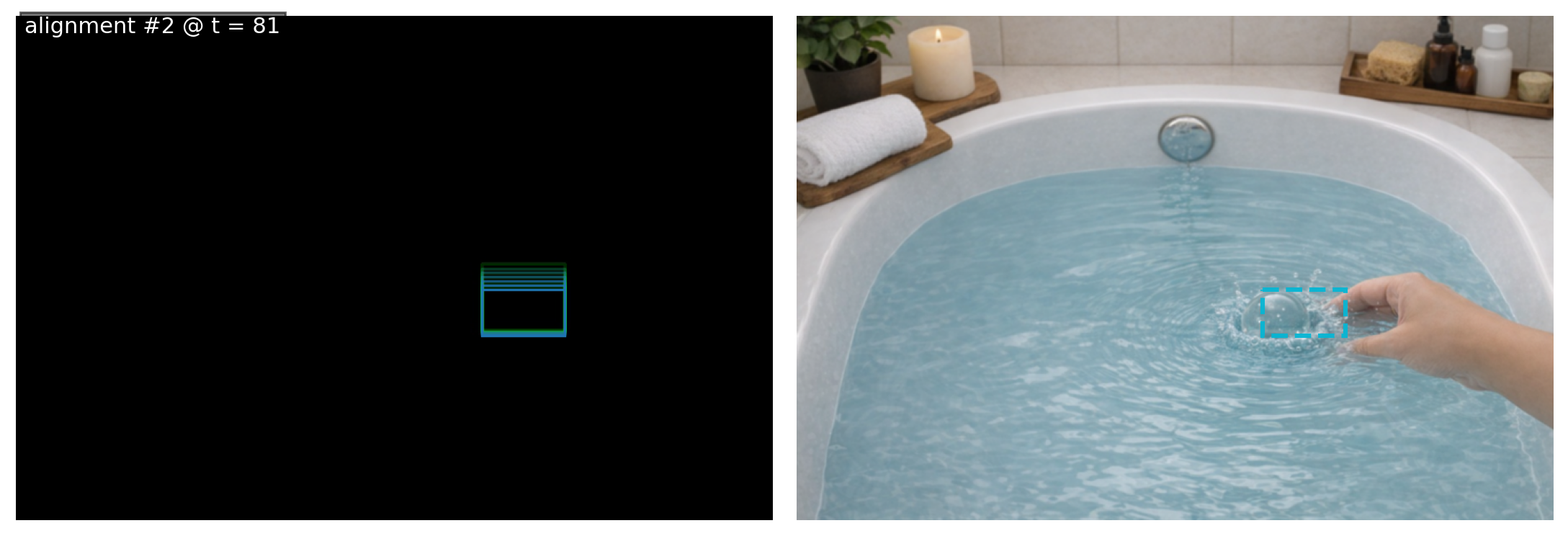} &
\includegraphics[width=0.5\linewidth, keepaspectratio]{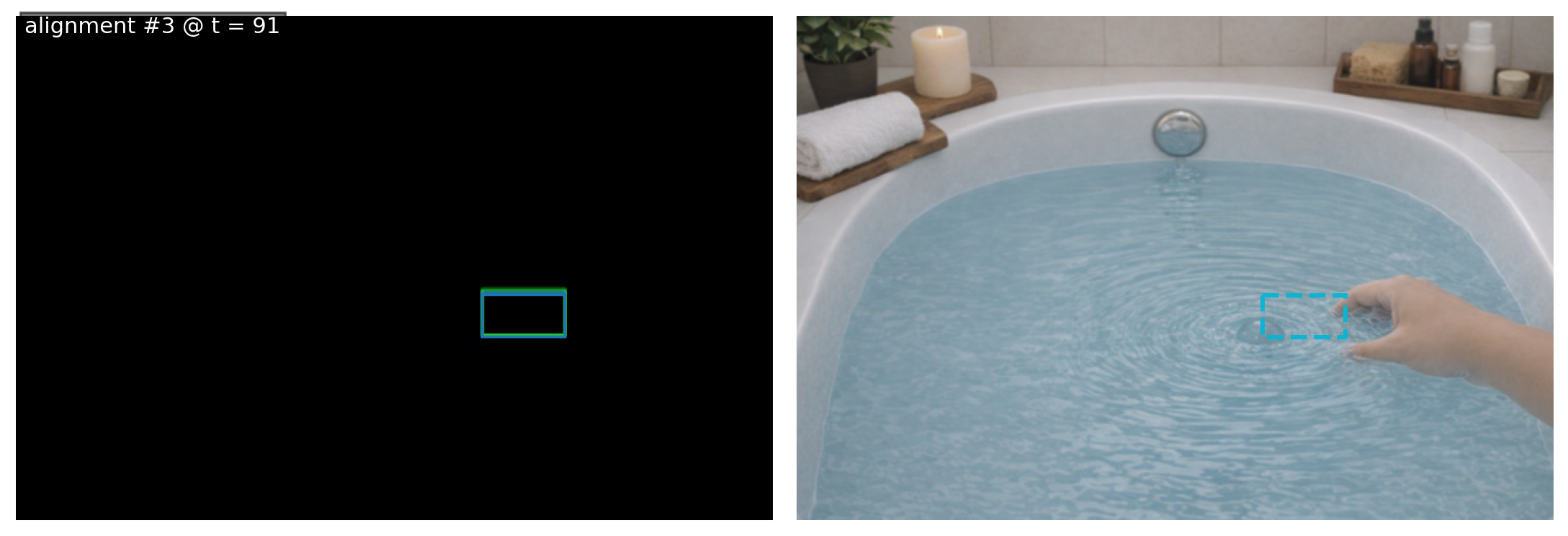} \\
\end{tabular}
\caption{Trajectory alignment visualization for the prompt: 
\textit{“A glass ball is gently placed on the surface of a bathtub filled with water.”} 
Each sub-panel represents one alignment step. The left image shows intermediate trajectory transitions between adjacent keyframes, while the right image shows the aligned keyframe. Blue dashed boxes denote aligned object states used for dense trajectory guidance.}
\label{fig:alignment_result_exp10}
\end{figure}

\begin{figure}[htbp]
\centering
\setlength{\tabcolsep}{0pt}  
\renewcommand{\arraystretch}{0}  
\begin{tabular}{cc}
\includegraphics[width=0.5\linewidth, keepaspectratio]{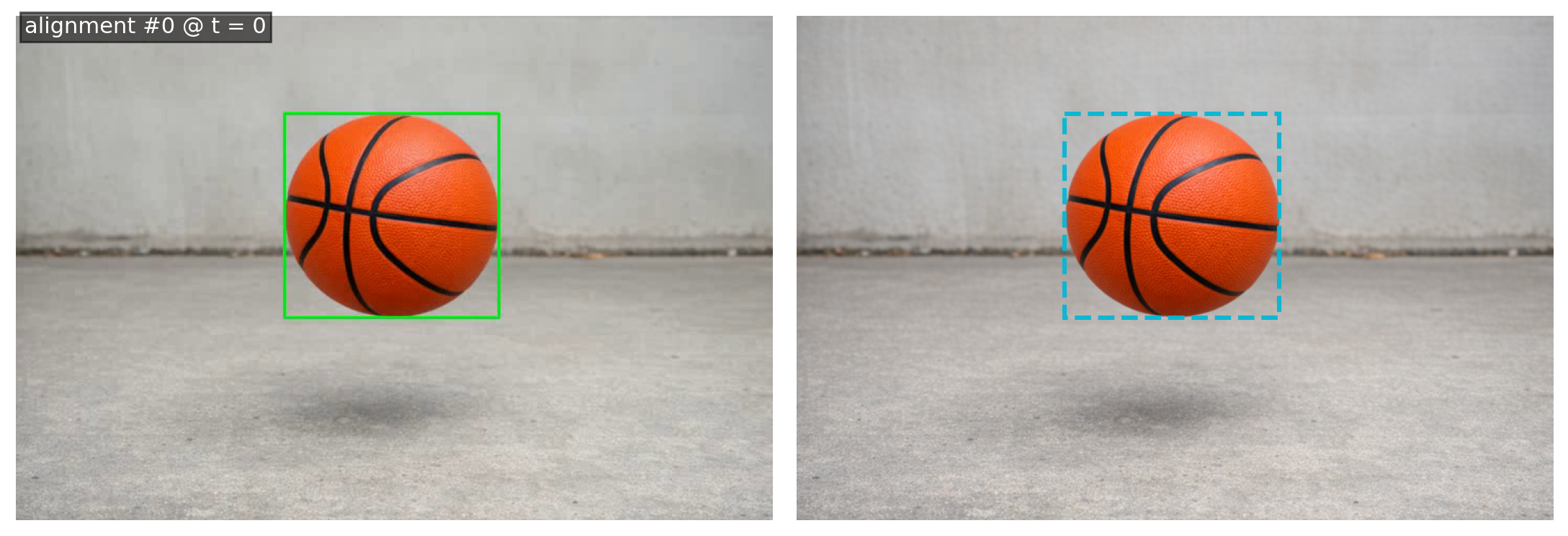} &
\includegraphics[width=0.5\linewidth, keepaspectratio]{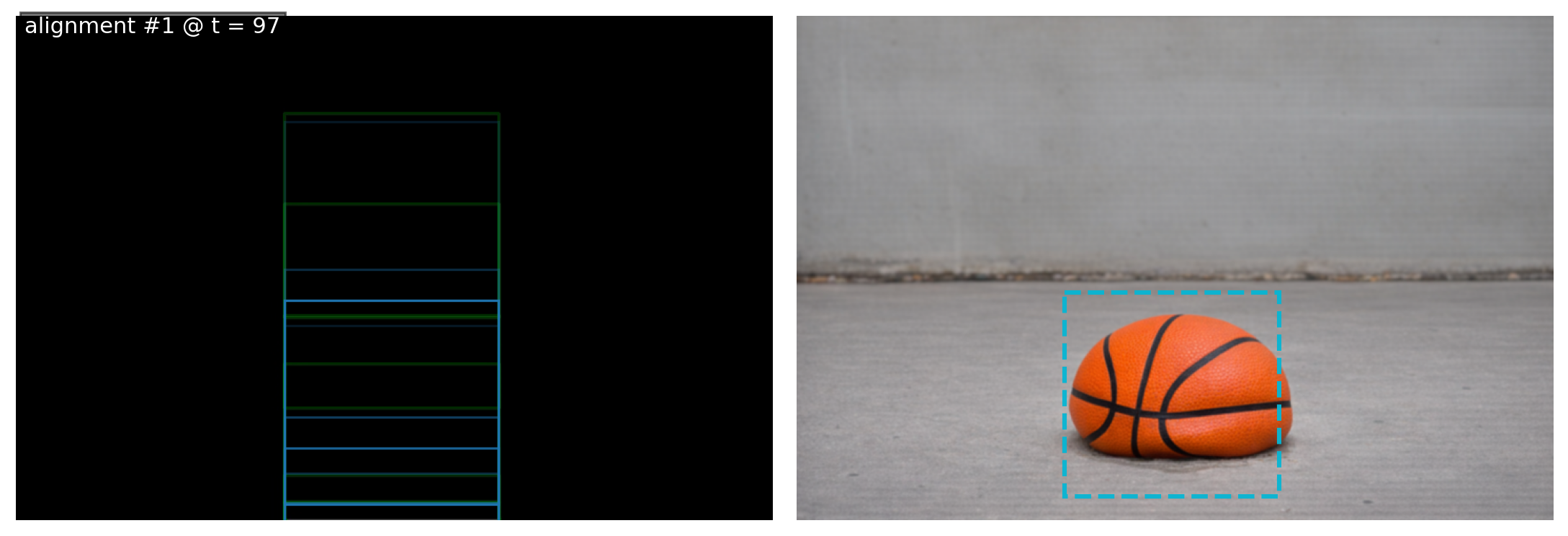} \\
\includegraphics[width=0.5\linewidth, keepaspectratio]{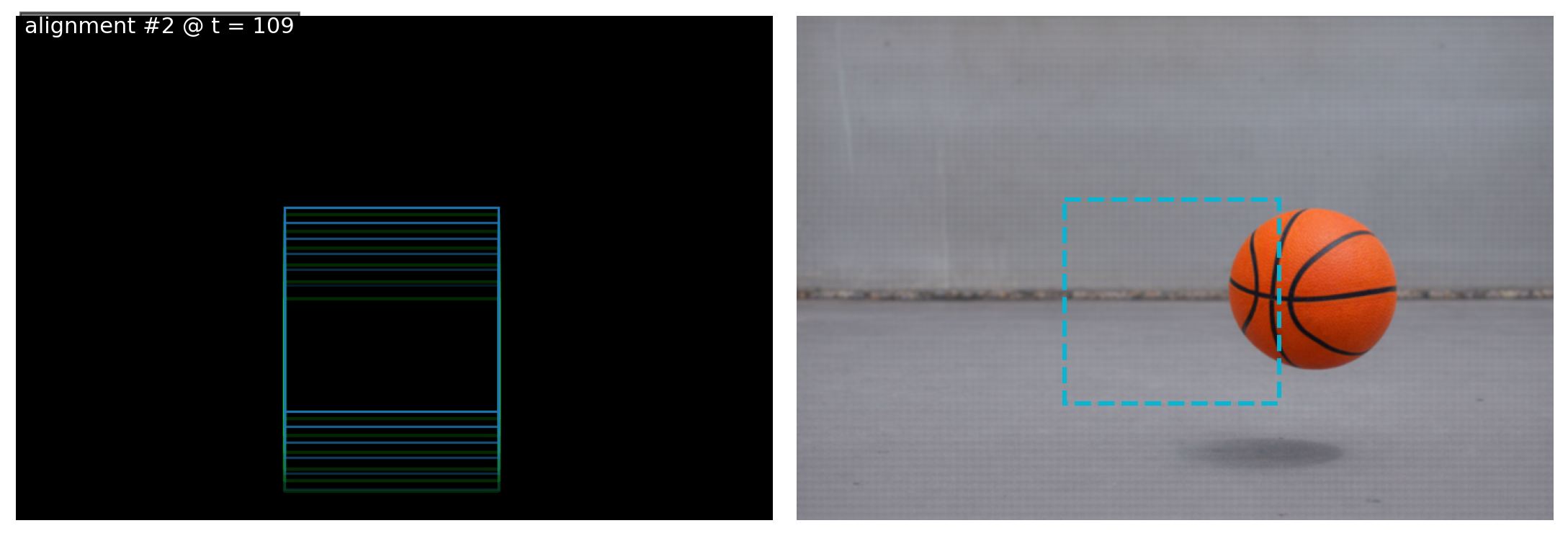} &
\includegraphics[width=0.5\linewidth, keepaspectratio]{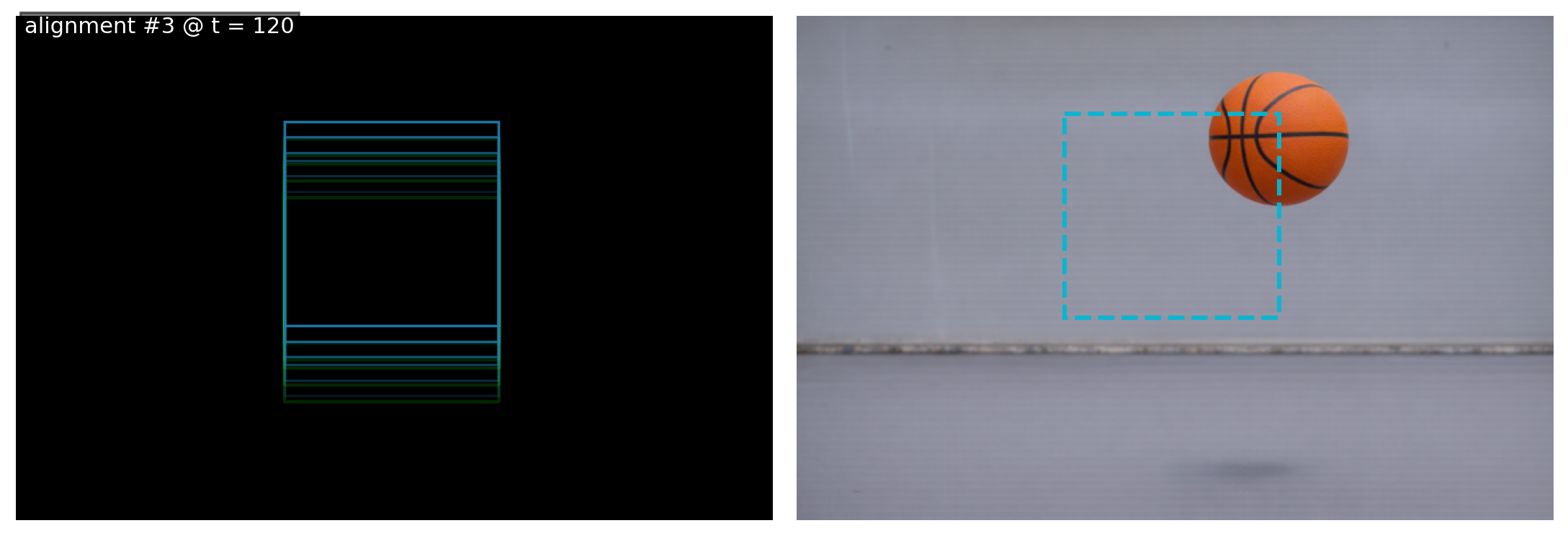} \\
\end{tabular}
\caption{Trajectory alignment visualization for the prompt: 
\textit{“A vibrant, elastic basketball is thrown forcefully toward the ground.”} 
The example illustrates how our method aligns sparse keyframes with dense trajectories to model physically consistent collision and rebound dynamics.}
\label{fig:alignment_result_exp27}
\end{figure}

Figures~\ref{fig:alignment_result_exp10} and~\ref{fig:alignment_result_exp27} present two examples of our trajectory alignment process under different physical scenarios: object-water interaction and collision dynamics. Each figure is organized as a 2$\times$2 grid, where each sub-panel corresponds to one alignment step between sparse keyframes and the final video timeline. Within each sub-panel, the \textbf{left} image visualizes intermediate motion segments between consecutive keyframes, where overlaid bounding boxes represent the predicted object trajectories over time. The \textbf{right} image shows the corresponding aligned keyframe at the selected timestamp, where blue dashed boxes indicate the aligned object states used for trajectory guidance. 

Figure~\ref{fig:alignment_result_exp10} demonstrates a gradual object-water interaction process, where the glass ball progressively approaches and contacts the water surface. Figure~\ref{fig:alignment_result_exp27} shows a fast collision scenario, where the basketball follows a physically plausible bouncing trajectory after impacting the ground. These examples illustrate how our alignment module bridges sparse keyframe reasoning with dense motion guidance, enabling temporally smooth and physically consistent video generation.

\section{Limitations}
Nevertheless, our approach still has limitations. The quality of trajectory planning depends on the reasoning capability of the VLM, and errors may accumulate in complex or highly interactive scenarios. Our trajectory alignment mechanism primarily relies on explicit object motion. Its benefits become limited in scenarios where objects primarily undergo appearance-only changes.

\section{Future Work}
Although our current framework is designed for physical video generation, the proposed keyframe reasoning and trajectory-guided generation paradigm has broader potential beyond physical scenarios. Future work could extend this framework to more general video generation tasks that require long-horizon temporal planning, such as human activities, multi-agent interactions, and story-driven video generation.

\section{Broader Impact}

Our work improves the physical plausibility and controllability of video generation, which can benefit applications such as scientific simulation, education, visual effects, and robotics prototyping. By enabling more structured and interpretable generation, it may also contribute to safer and more controllable generative systems.

However, like other video generation technologies, our method could be misused to create misleading or synthetic media (e.g., deepfakes), potentially contributing to misinformation or manipulation. Additionally, errors in physical reasoning could lead to incorrect interpretations in safety-critical applications.

To mitigate these risks, future work could explore safeguards such as watermarking, usage restrictions, and integration with detection systems for synthetic media. We encourage responsible use aligned with ethical guidelines.

\section{More qualitative results}
To better illustrate the result of our method, we provide more examples compared with strong baseline VLIPP and PhyGDPO.
\begin{figure}[!htb]
\centering
\setlength{\tabcolsep}{0pt}
\renewcommand{\arraystretch}{0}

A wooden pencil is carefully dipped into a glass of crystal-clear water, showing the intriguing visual shifts and reflections caused by the interaction between the pencil and the liquid. \\

\begin{tabular}{c c c c c}
\includegraphics[width=0.18\textwidth]{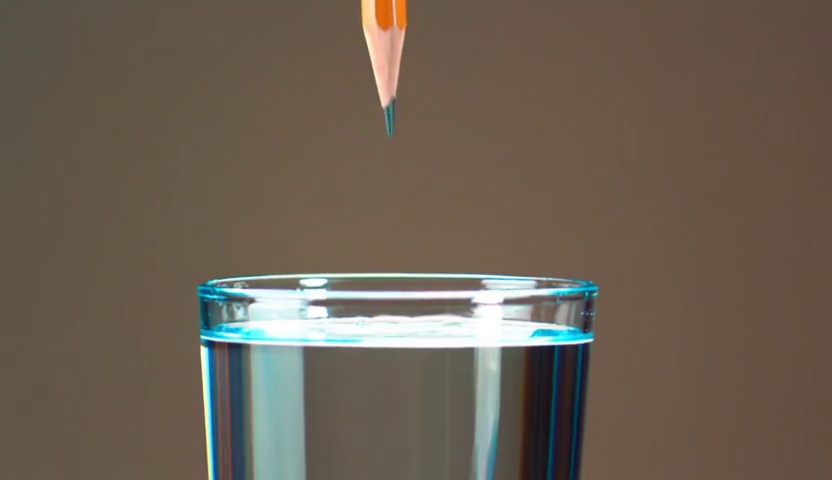} &
\includegraphics[width=0.18\textwidth]{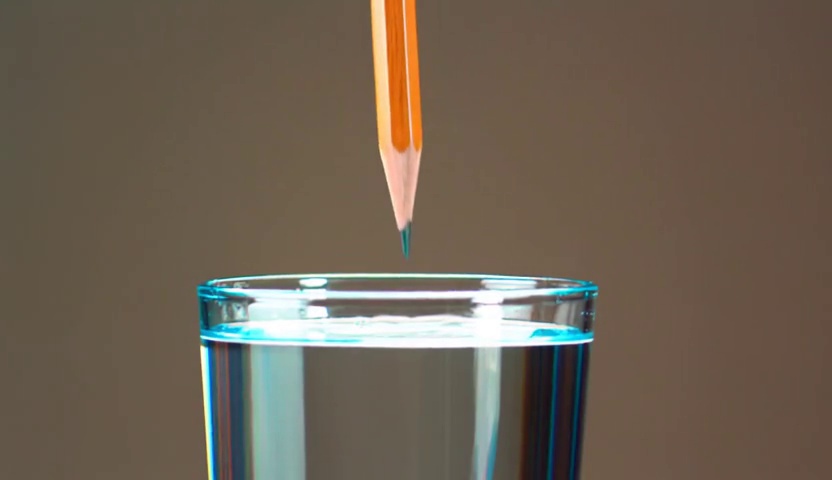} &
\includegraphics[width=0.18\textwidth]{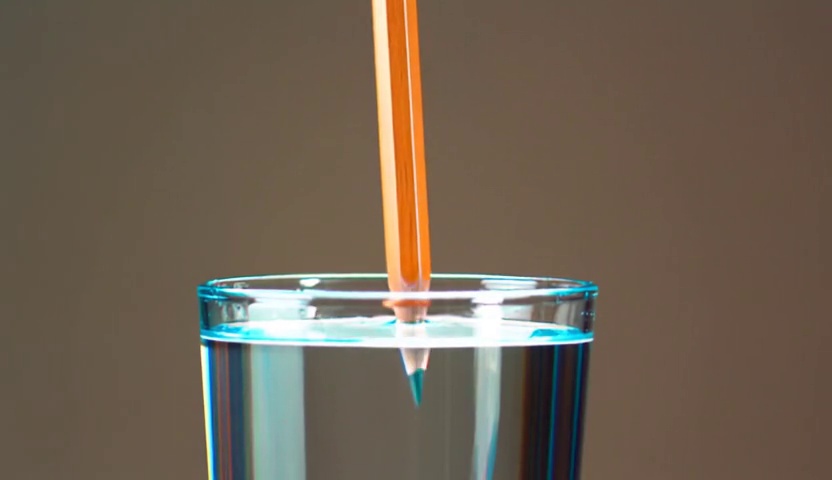} &
\includegraphics[width=0.18\textwidth]{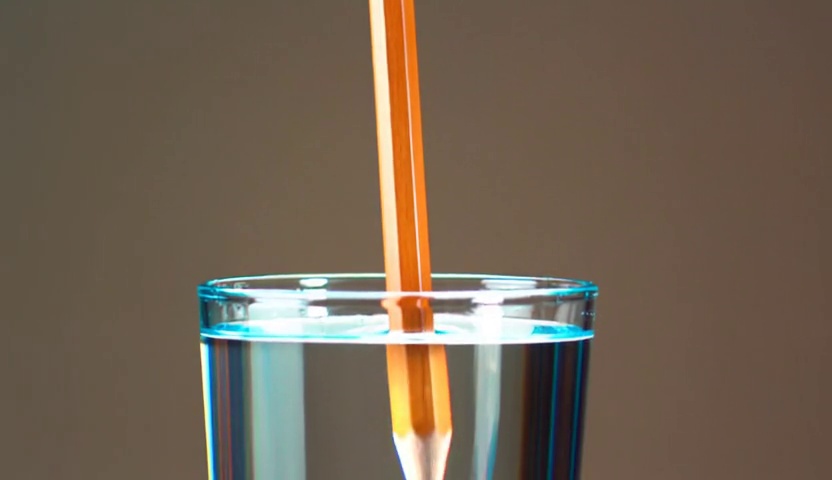} &
\includegraphics[width=0.18\textwidth]{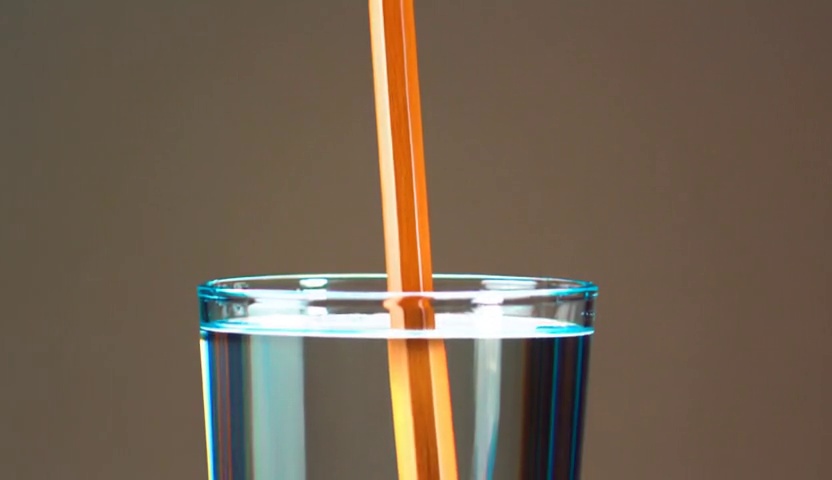} \\

\includegraphics[width=0.18\textwidth]{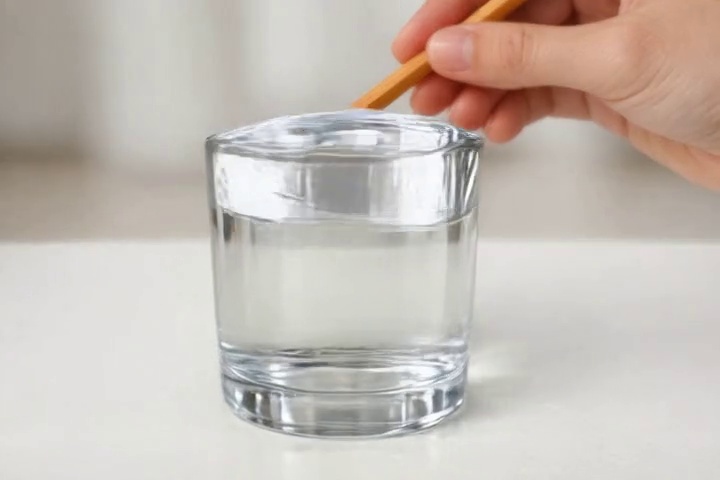} &
\includegraphics[width=0.18\textwidth]{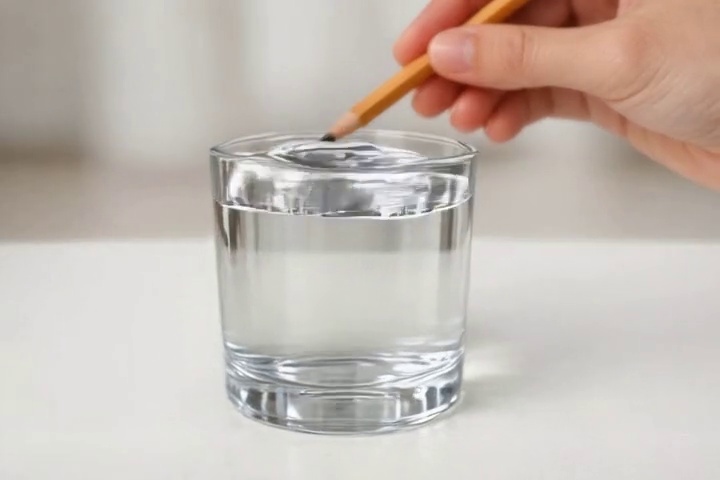} &
\includegraphics[width=0.18\textwidth]{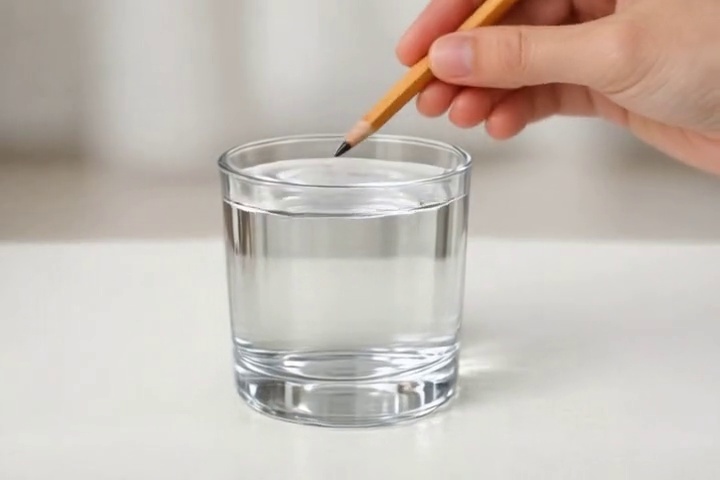} &
\includegraphics[width=0.18\textwidth]{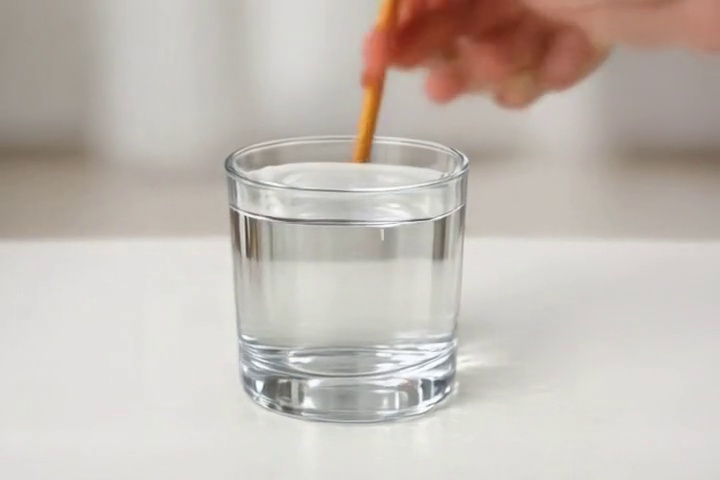} &
\includegraphics[width=0.18\textwidth]{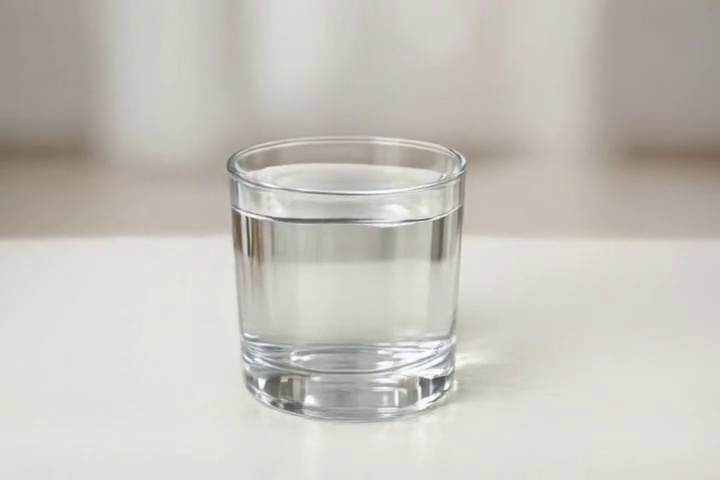} \\

\includegraphics[width=0.18\textwidth]{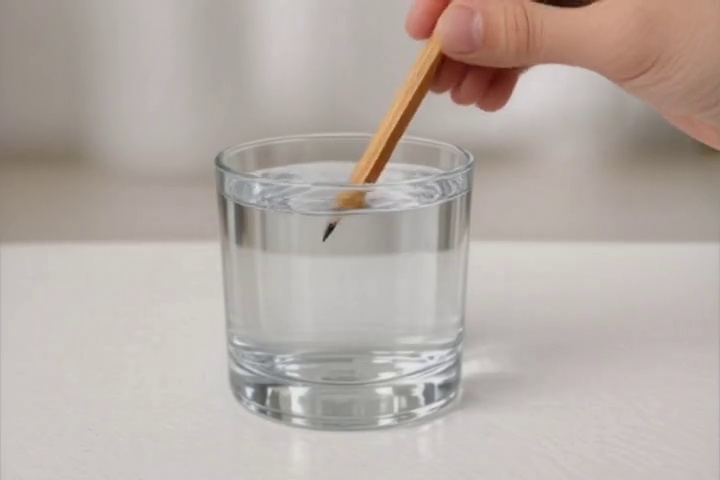} &
\includegraphics[width=0.18\textwidth]{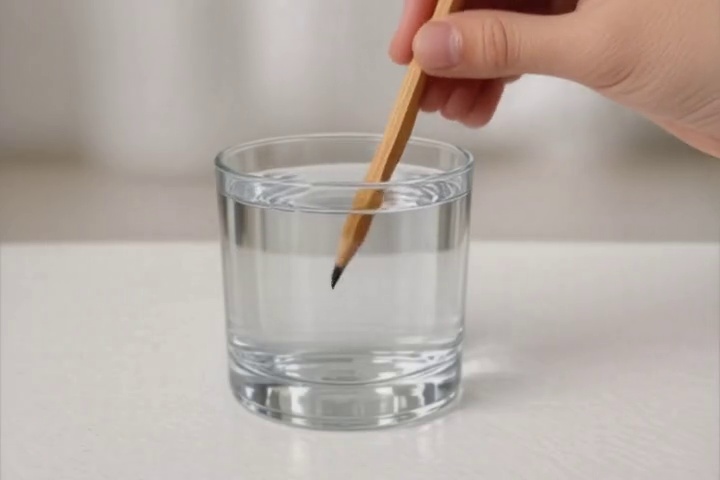} &
\includegraphics[width=0.18\textwidth]{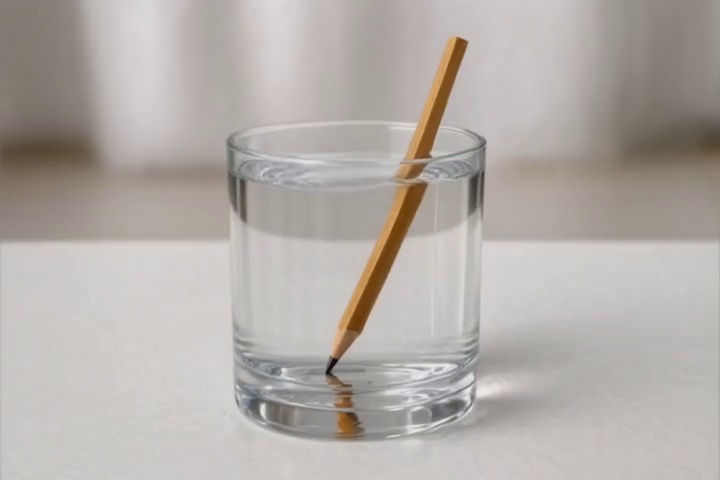} &
\includegraphics[width=0.18\textwidth]{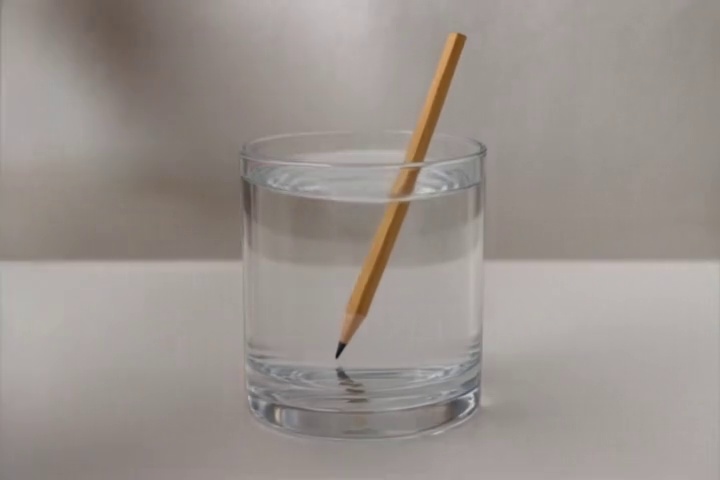} &
\includegraphics[width=0.18\textwidth]{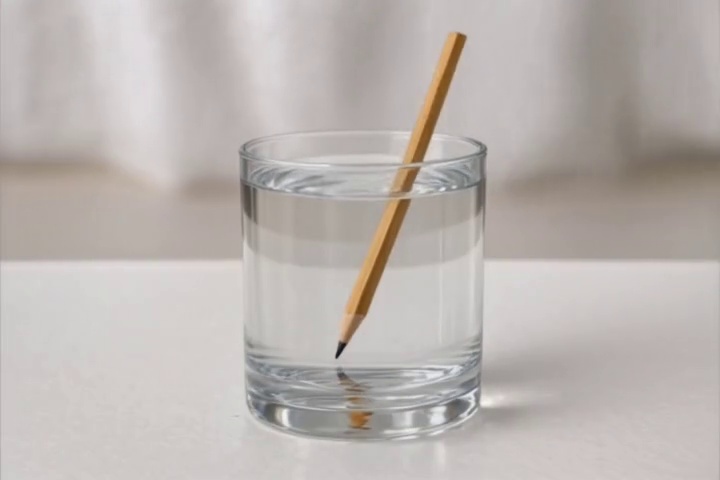} \\

\end{tabular}

\vspace{2mm}

A small burning ball of paper was thrown into a pile of dry paper.  \\

\begin{tabular}{c c c c c}
\includegraphics[width=0.18\textwidth]{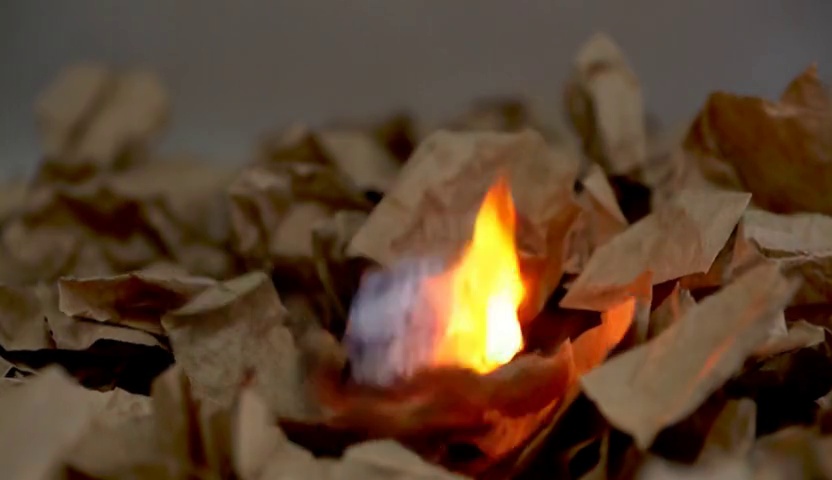} &
\includegraphics[width=0.18\textwidth]{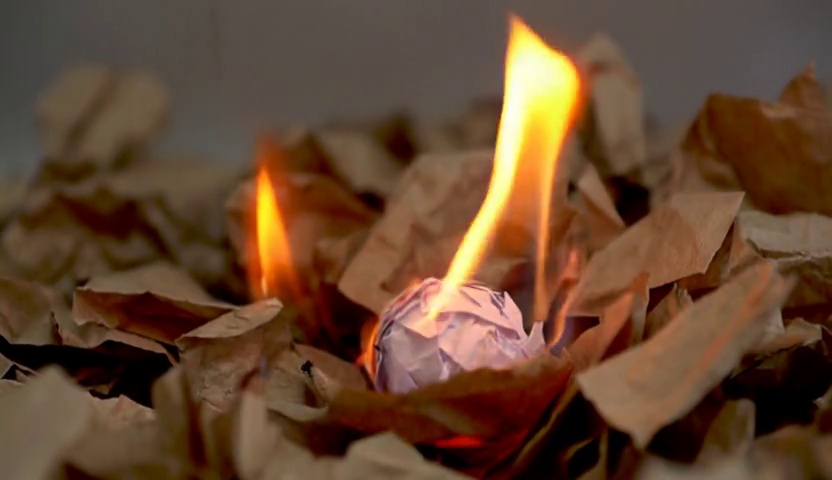} &
\includegraphics[width=0.18\textwidth]{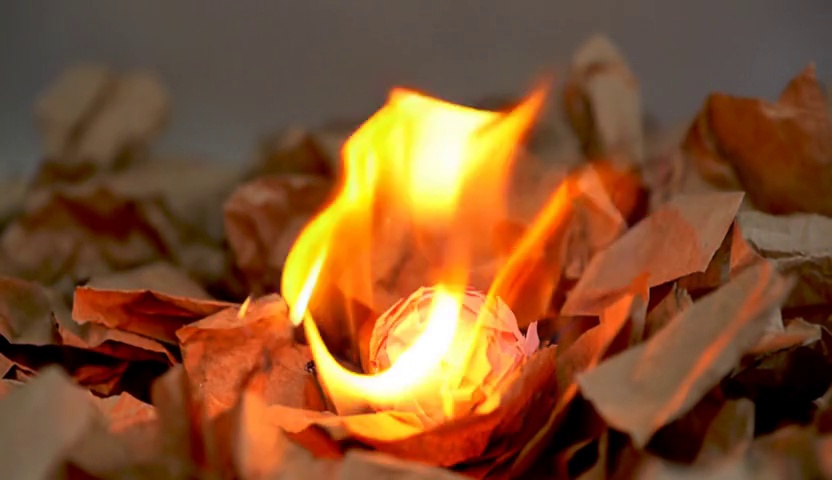} &
\includegraphics[width=0.18\textwidth]{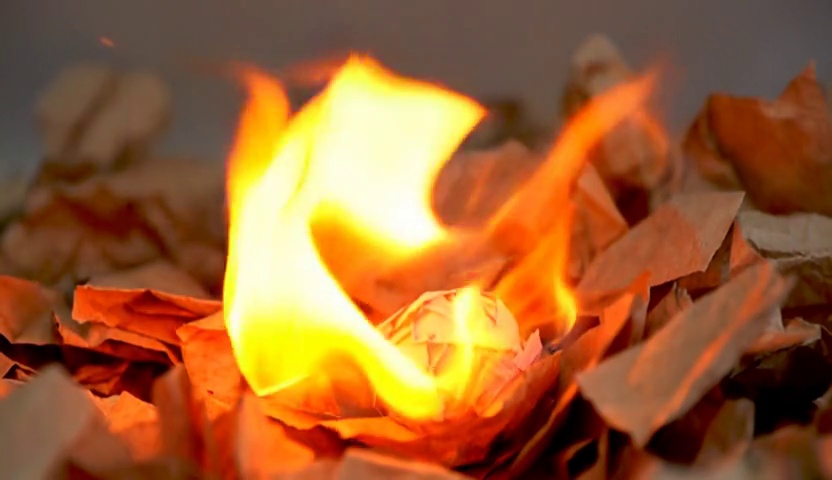} &
\includegraphics[width=0.18\textwidth]{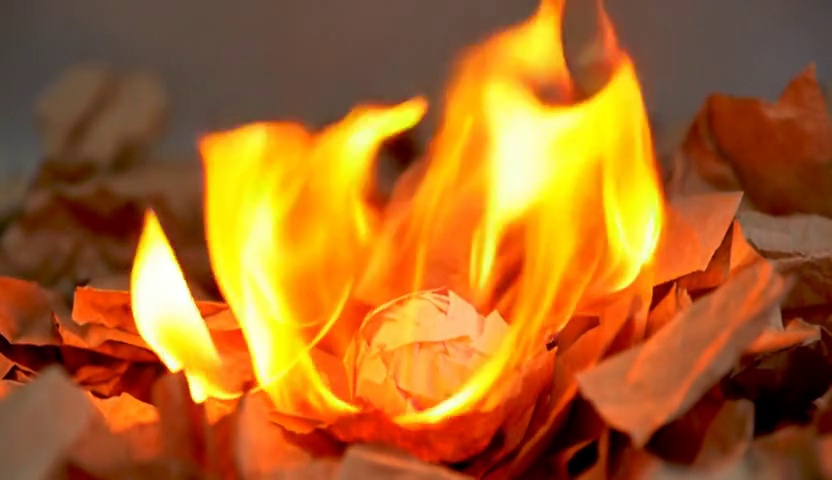} \\

\includegraphics[width=0.18\textwidth]{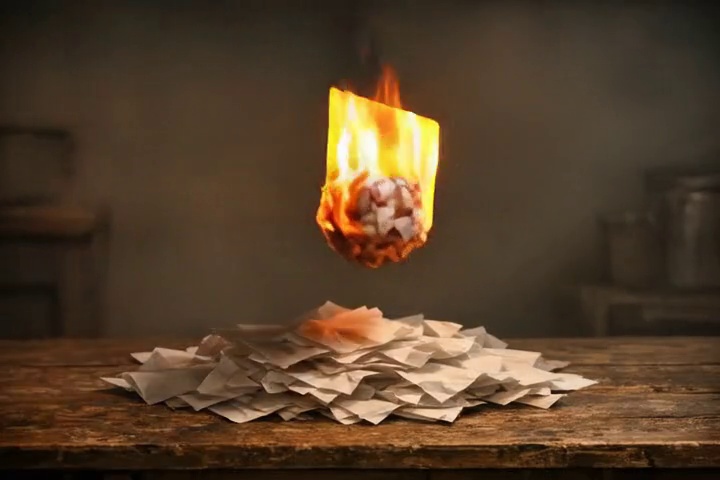} &
\includegraphics[width=0.18\textwidth]{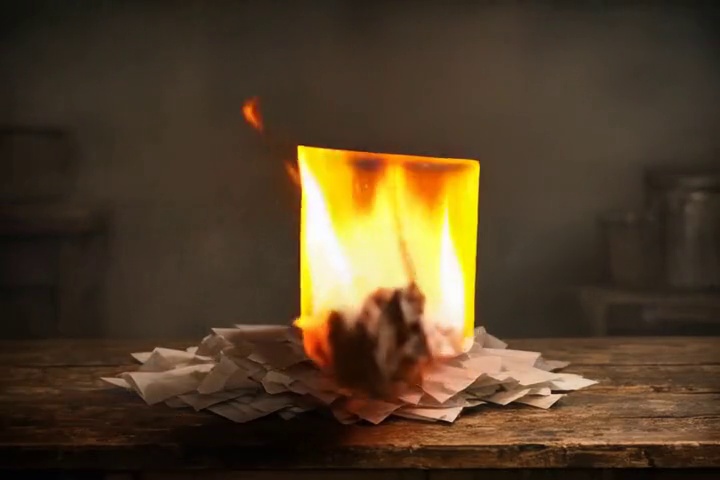} &
\includegraphics[width=0.18\textwidth]{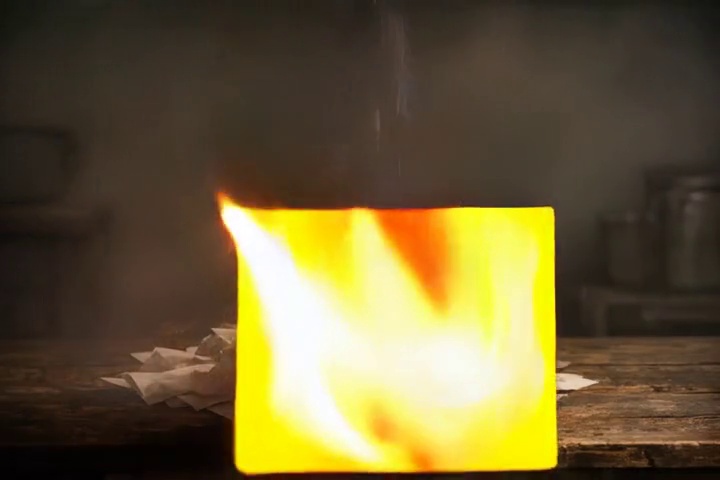} &
\includegraphics[width=0.18\textwidth]{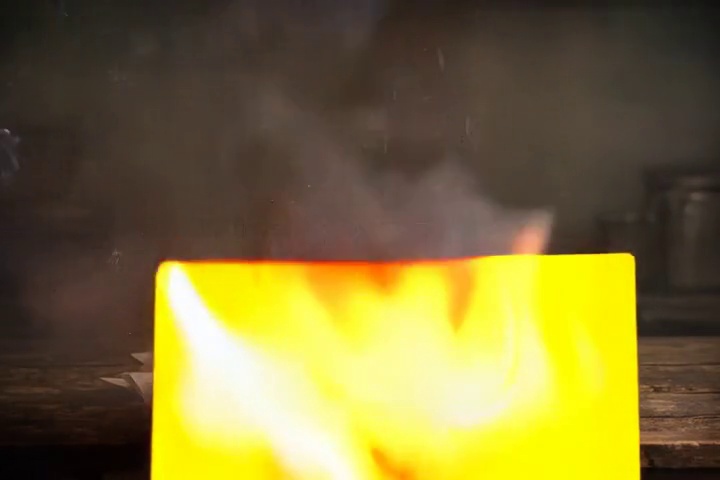} &
\includegraphics[width=0.18\textwidth]{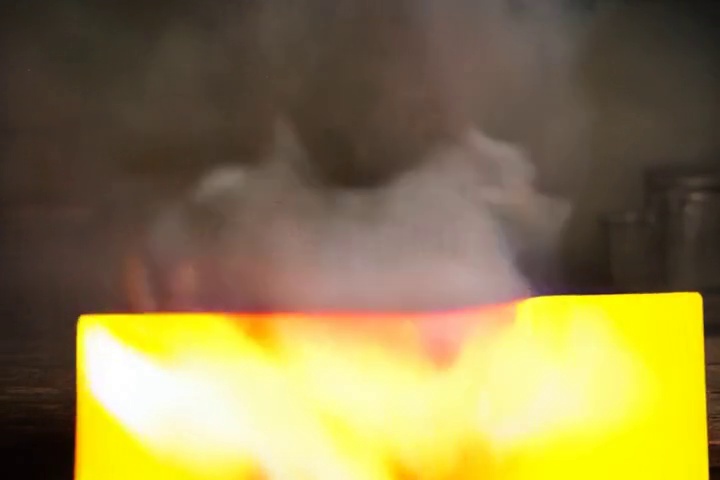} \\

\includegraphics[width=0.18\textwidth]{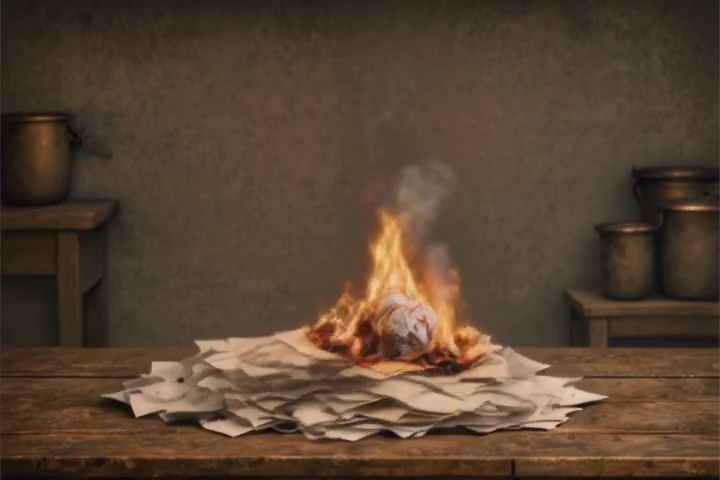} &
\includegraphics[width=0.18\textwidth]{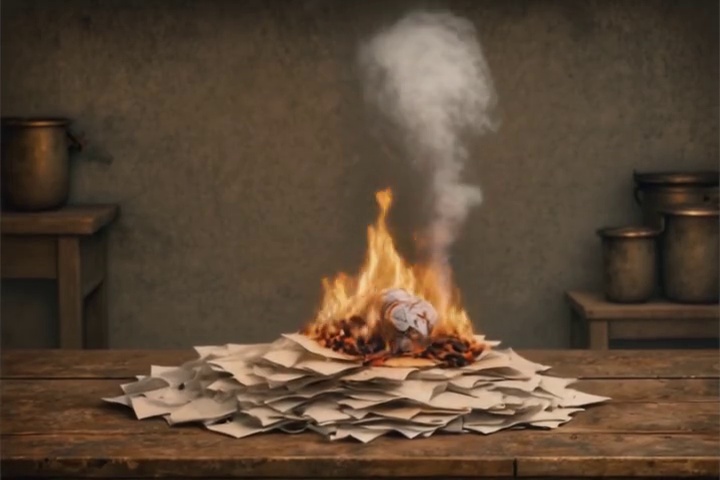} &
\includegraphics[width=0.18\textwidth]{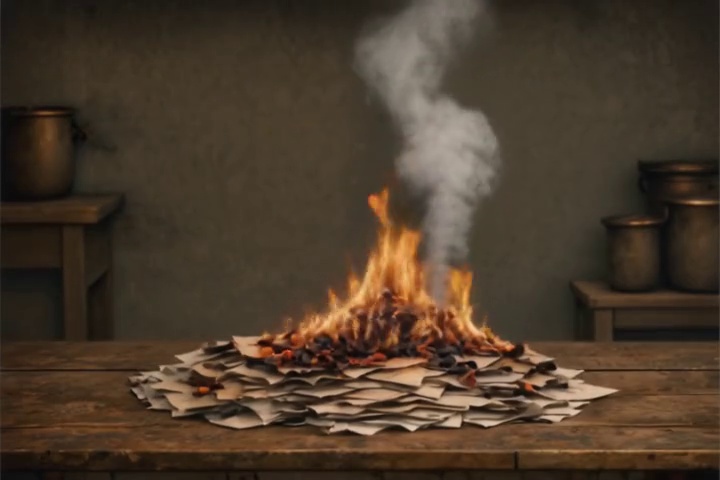} &
\includegraphics[width=0.18\textwidth]{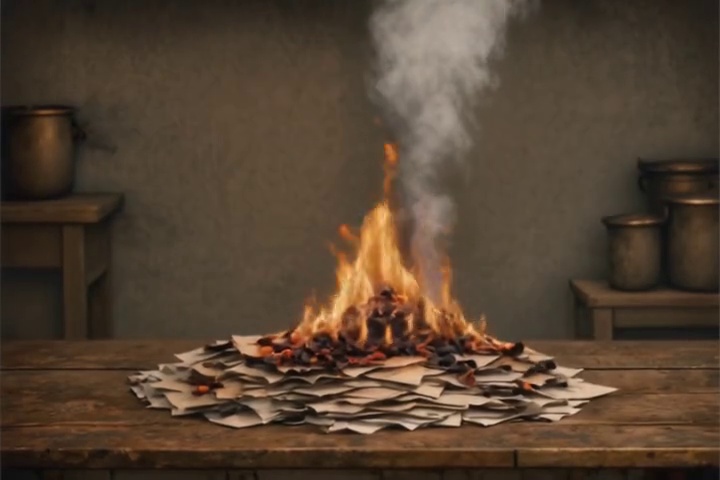} &
\includegraphics[width=0.18\textwidth]{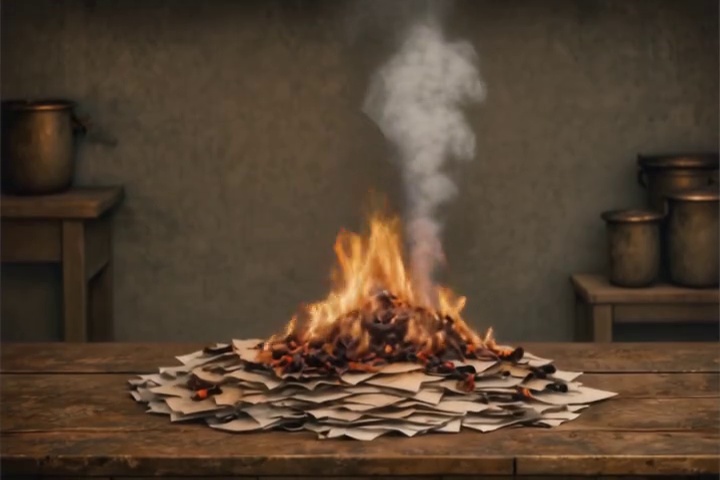} \\

\end{tabular}

\caption{Additional qualitative comparisons between PhyGDPO, VLIPP, and our method.}
\label{fig:sup_qualitative_results_1}
\end{figure}

\begin{figure}[!htb]
\centering
\setlength{\tabcolsep}{0pt}
\renewcommand{\arraystretch}{0}

A timelapse captures the gradual transformation of butter as the temperature rises significantly. \\

\begin{tabular}{c c c c c}

\includegraphics[width=0.18\textwidth]{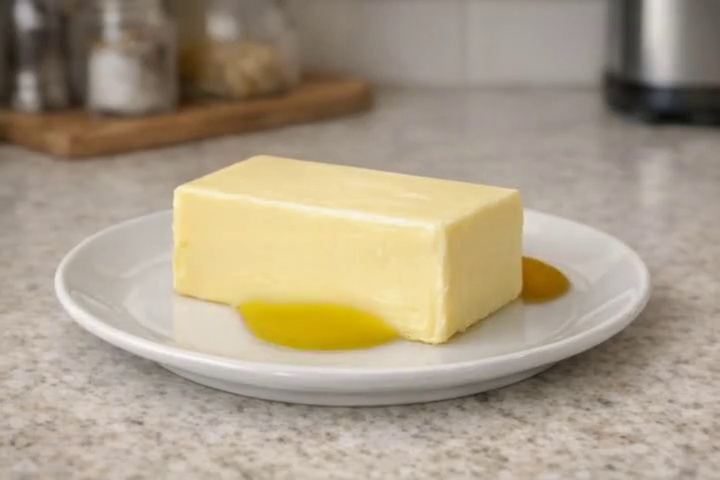} &
\includegraphics[width=0.18\textwidth]{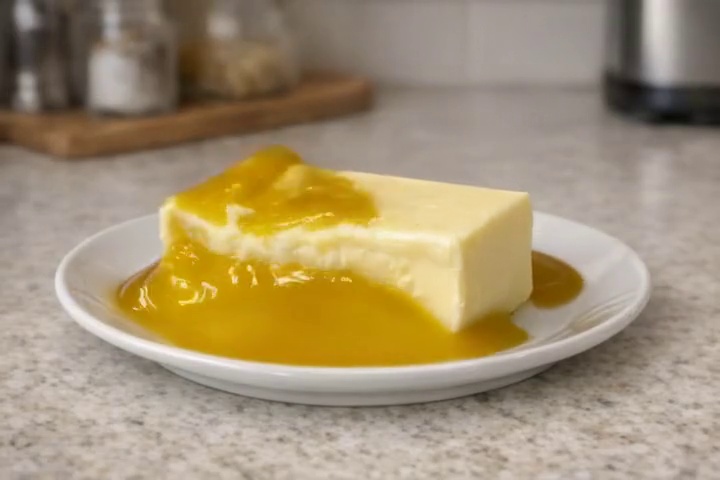} &
\includegraphics[width=0.18\textwidth]{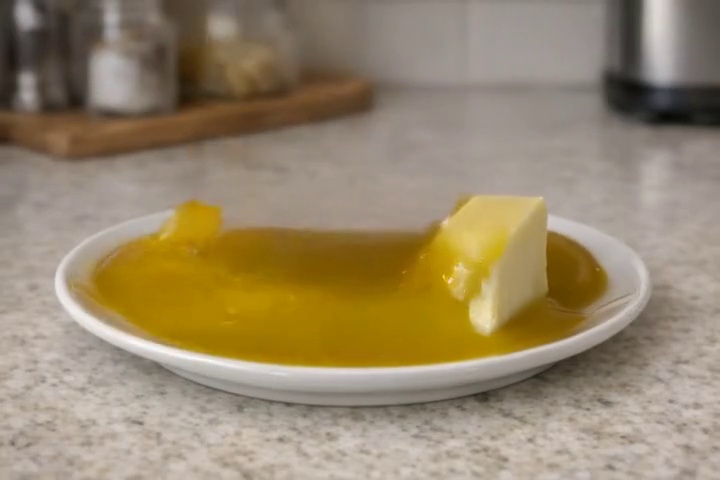} &
\includegraphics[width=0.18\textwidth]{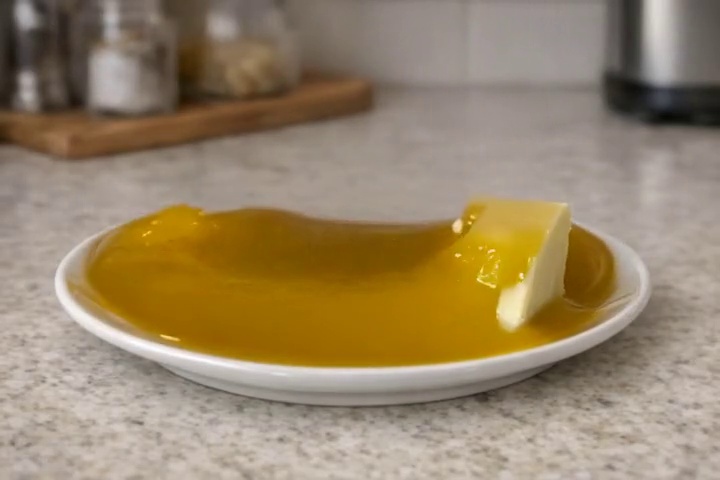} &
\includegraphics[width=0.18\textwidth]{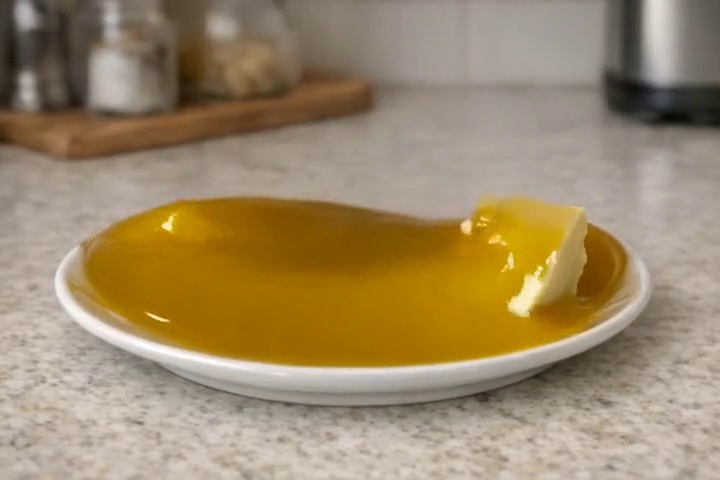} \\

\includegraphics[width=0.18\textwidth]{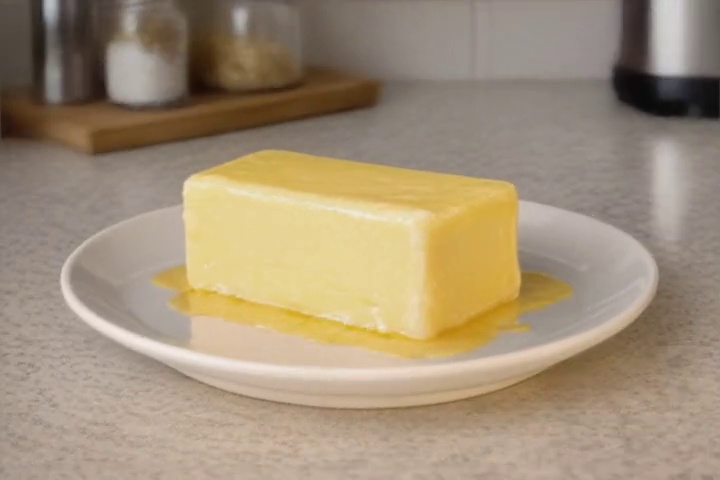} &
\includegraphics[width=0.18\textwidth]{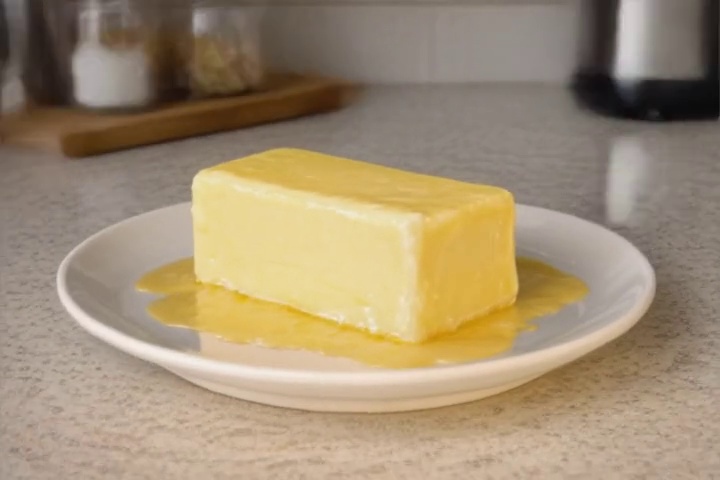} &
\includegraphics[width=0.18\textwidth]{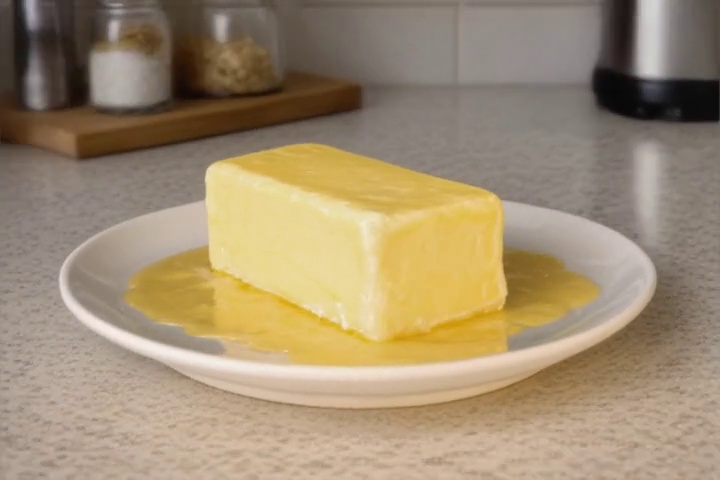} &
\includegraphics[width=0.18\textwidth]{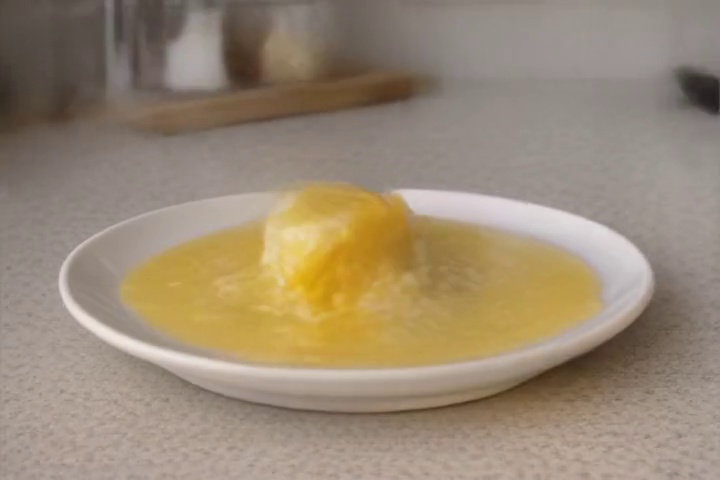} &
\includegraphics[width=0.18\textwidth]{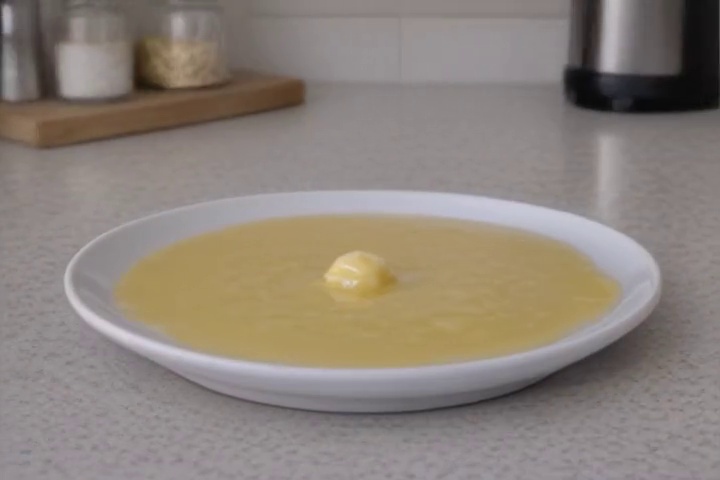} \\

\end{tabular}

\vspace{2mm}

A glistening dewdrop is sliding gracefully across the smooth surface of a waxed apple, accentuating its shape as it moves.  \\

\begin{tabular}{c c c c c}

\includegraphics[width=0.18\textwidth]{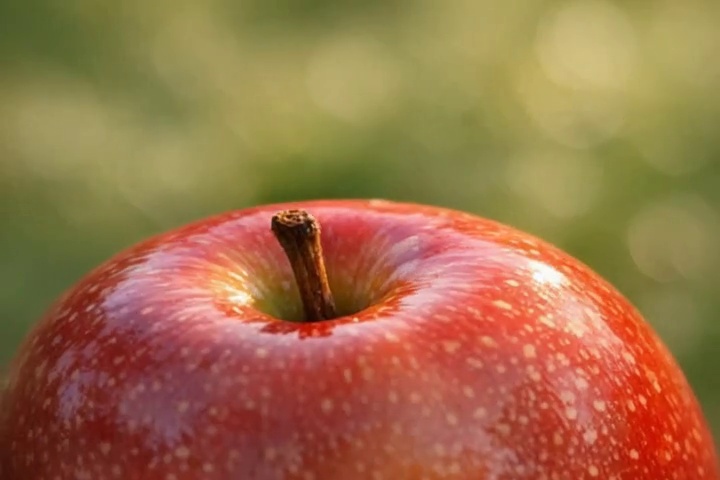} &
\includegraphics[width=0.18\textwidth]{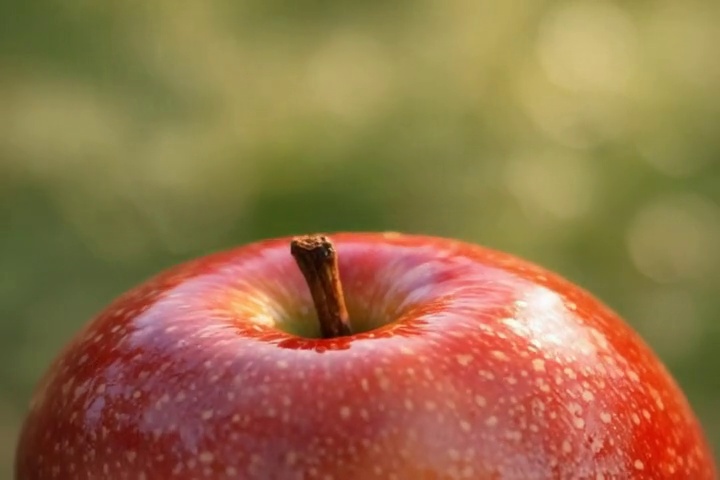} &
\includegraphics[width=0.18\textwidth]{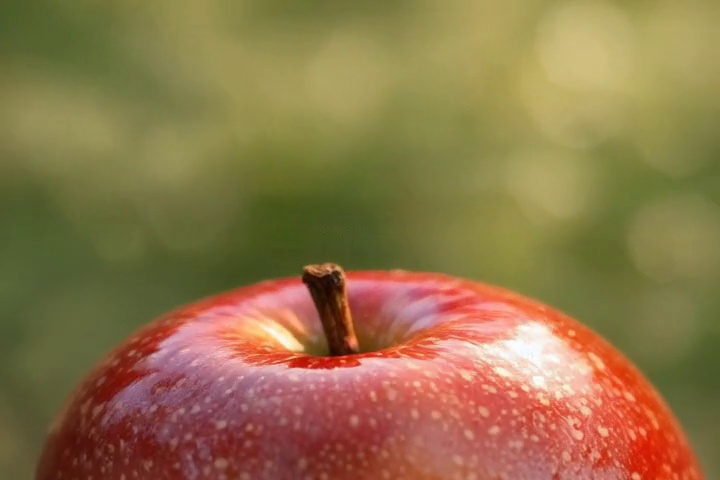} &
\includegraphics[width=0.18\textwidth]{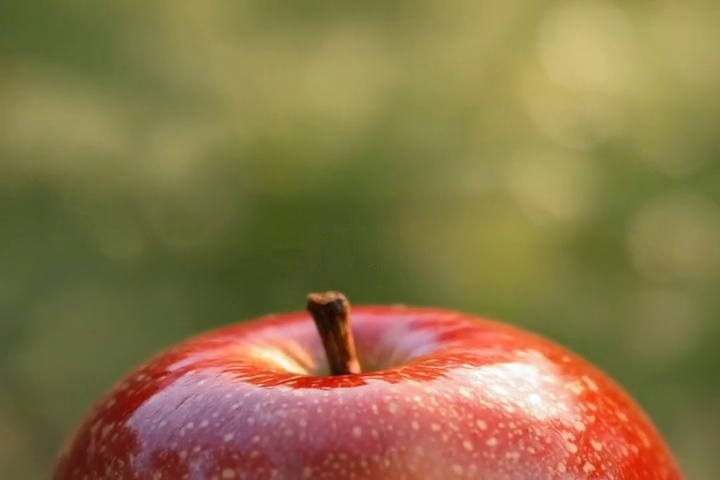} &
\includegraphics[width=0.18\textwidth]{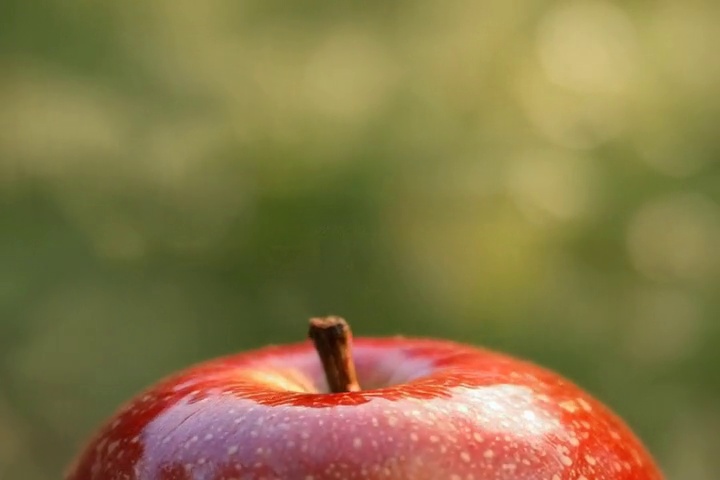} \\

\includegraphics[width=0.18\textwidth]{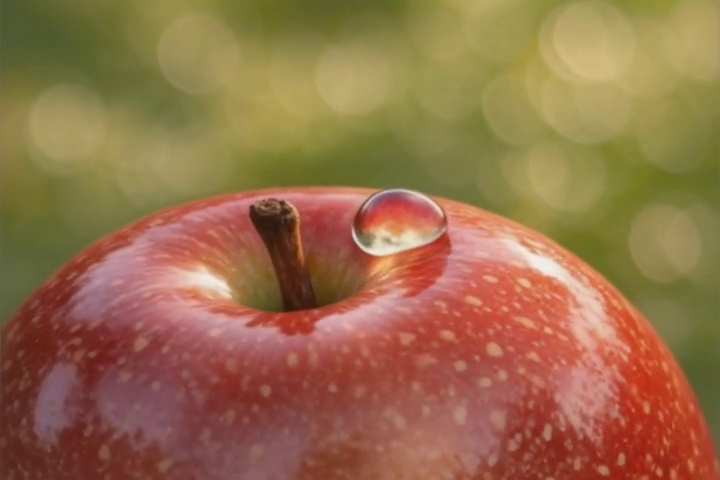} &
\includegraphics[width=0.18\textwidth]{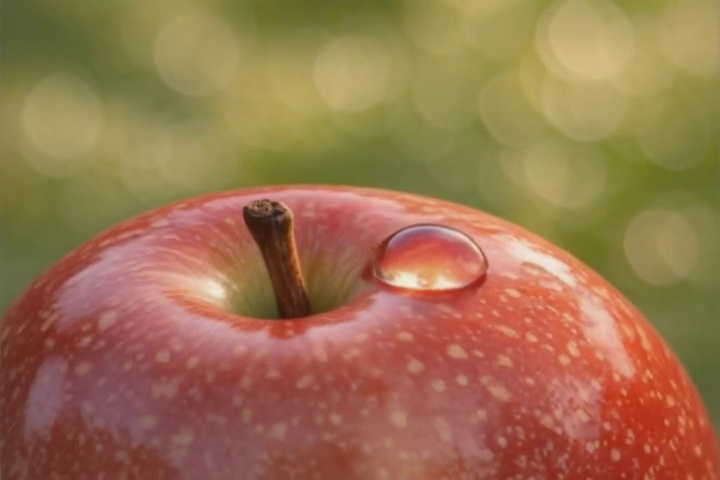} &
\includegraphics[width=0.18\textwidth]{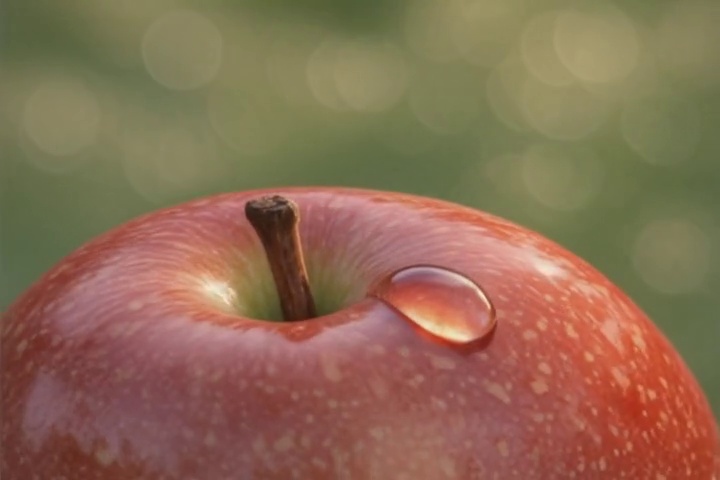} &
\includegraphics[width=0.18\textwidth]{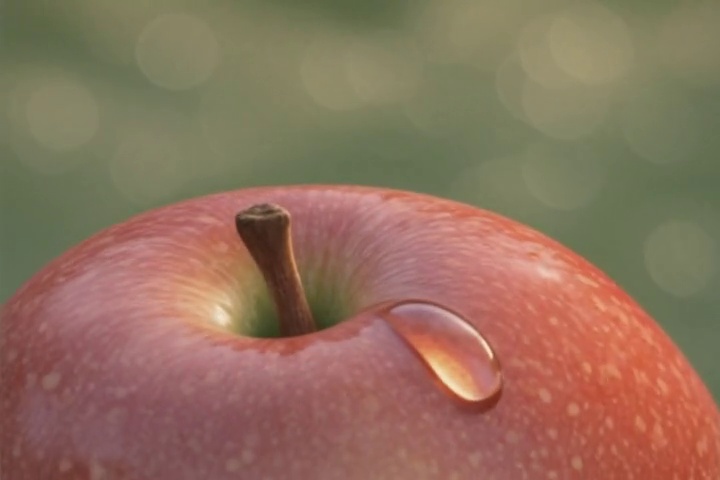} &
\includegraphics[width=0.18\textwidth]{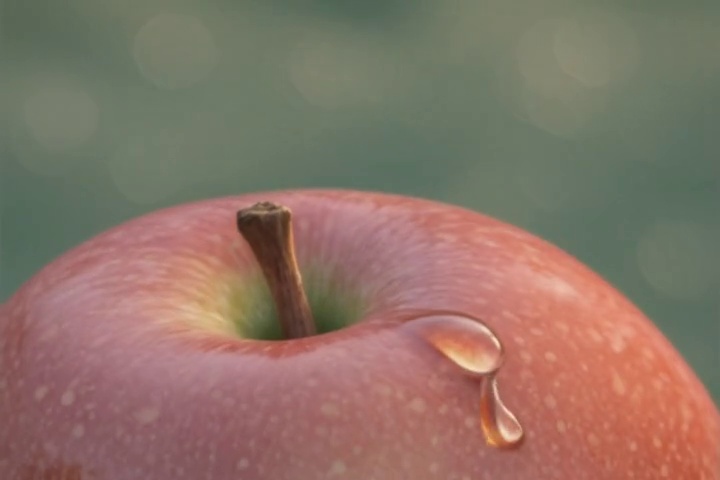} \\

\end{tabular}

\vspace{2mm}

A weak, frail porcelain plate is flung with significant speed at a robust, wooden table, where it collides upon impact.  \\

\begin{tabular}{c c c c c}

\includegraphics[width=0.18\textwidth]{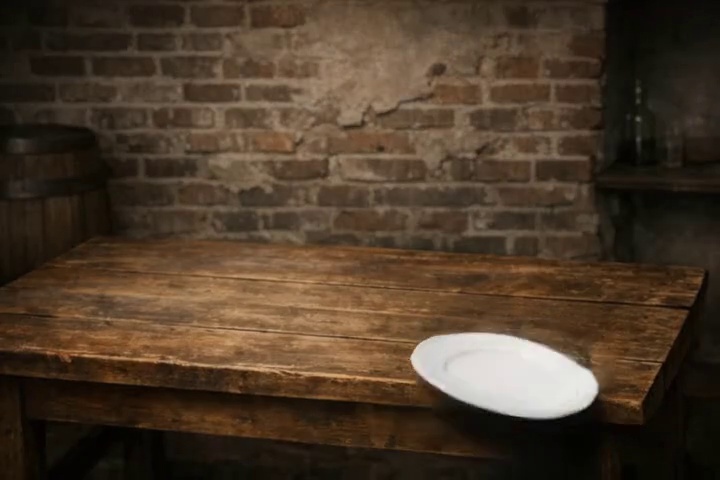} &
\includegraphics[width=0.18\textwidth]{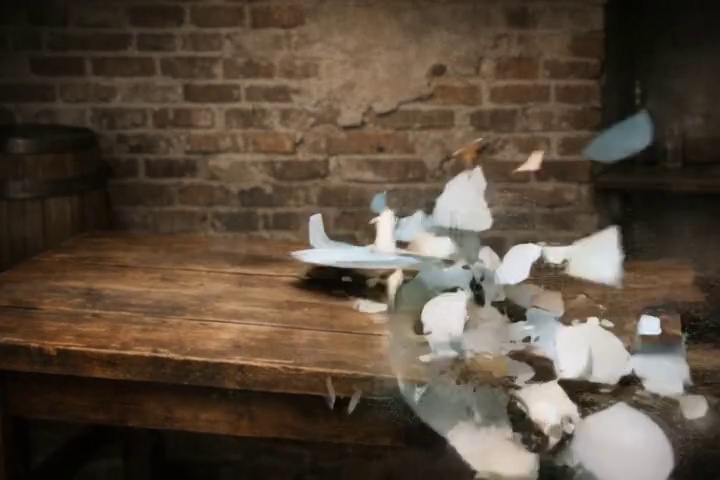} &
\includegraphics[width=0.18\textwidth]{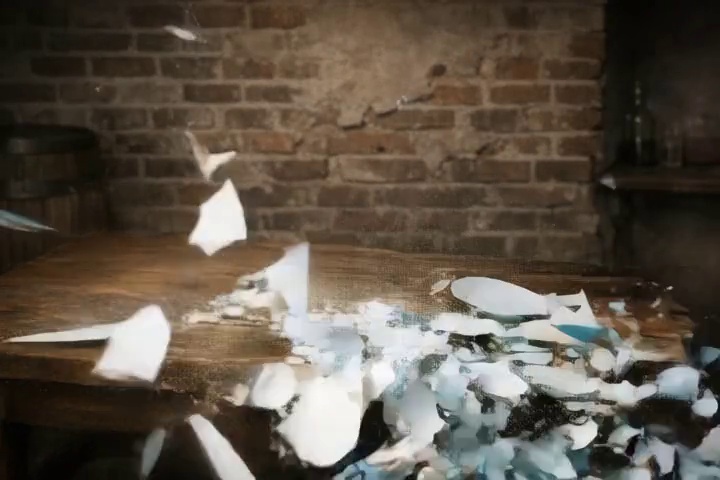} &
\includegraphics[width=0.18\textwidth]{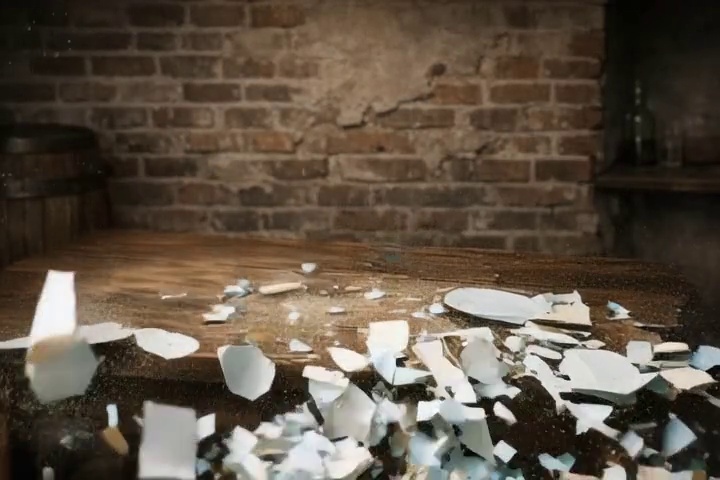} &
\includegraphics[width=0.18\textwidth]{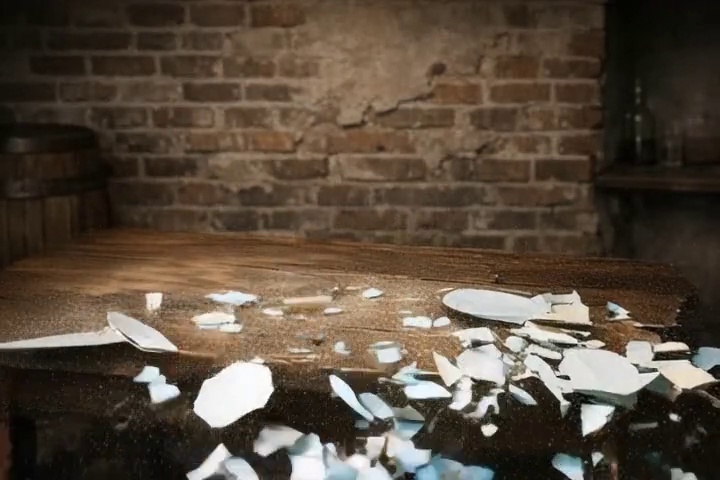} \\

\includegraphics[width=0.18\textwidth]{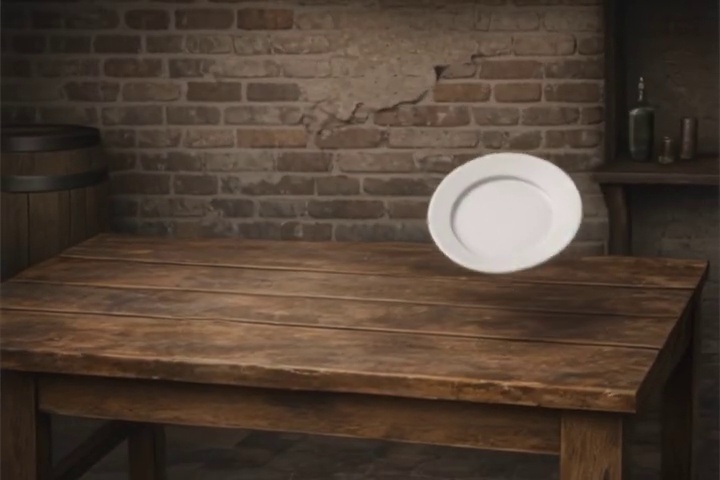} &
\includegraphics[width=0.18\textwidth]{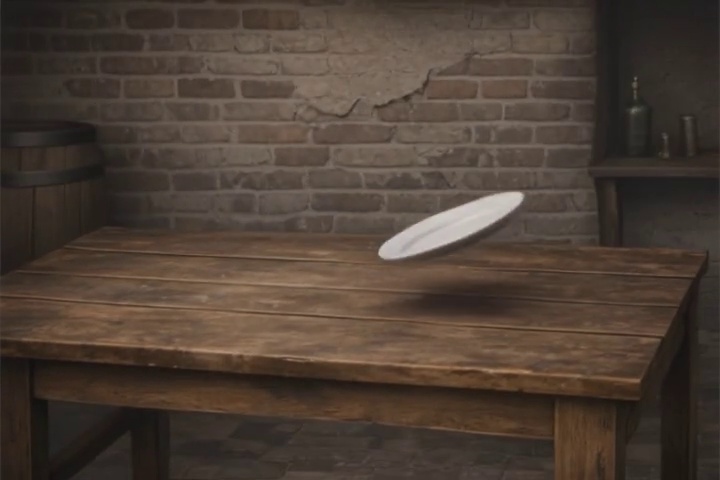} &
\includegraphics[width=0.18\textwidth]{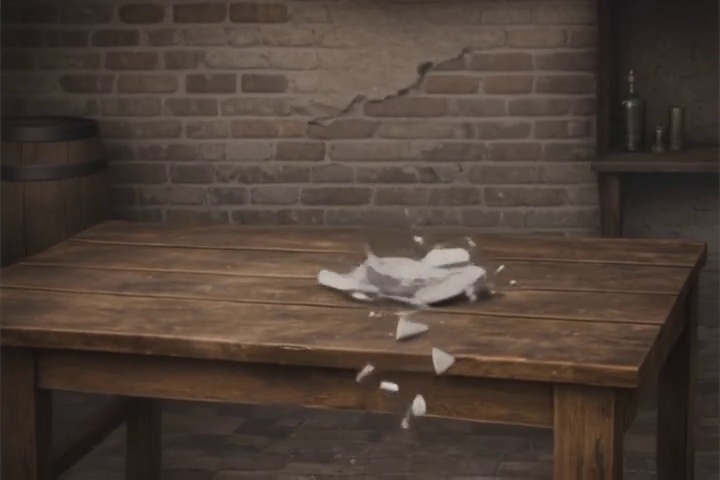} &
\includegraphics[width=0.18\textwidth]{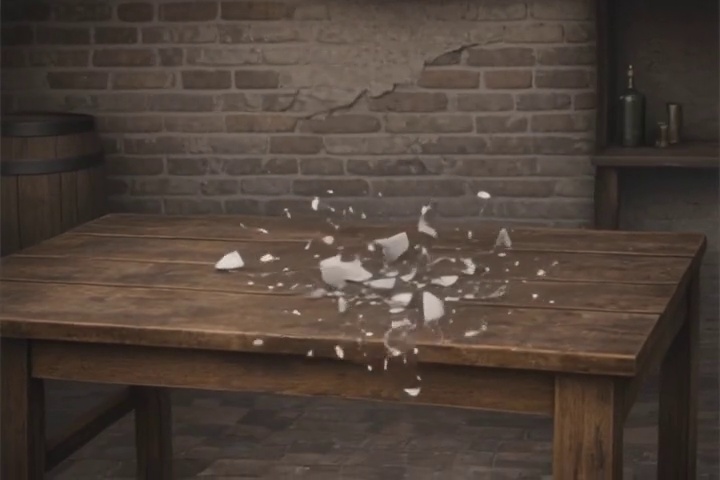} &
\includegraphics[width=0.18\textwidth]{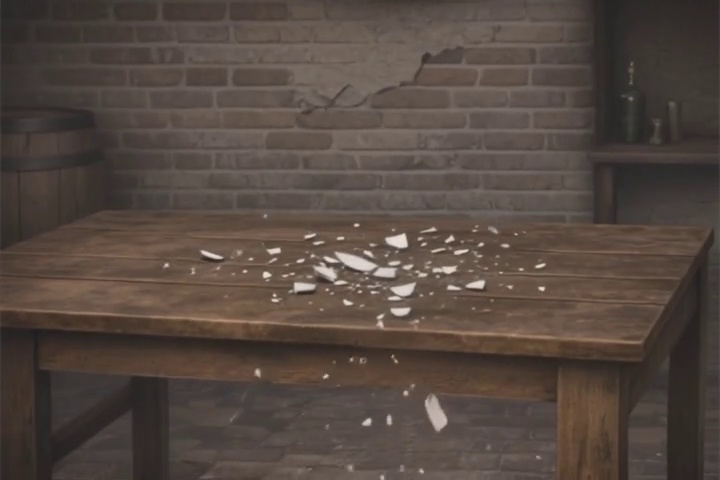} \\

\end{tabular}

Concentrated sulfuric acid is poured onto a wooden table.  \\

\begin{tabular}{c c c c c}

\includegraphics[width=0.18\textwidth]{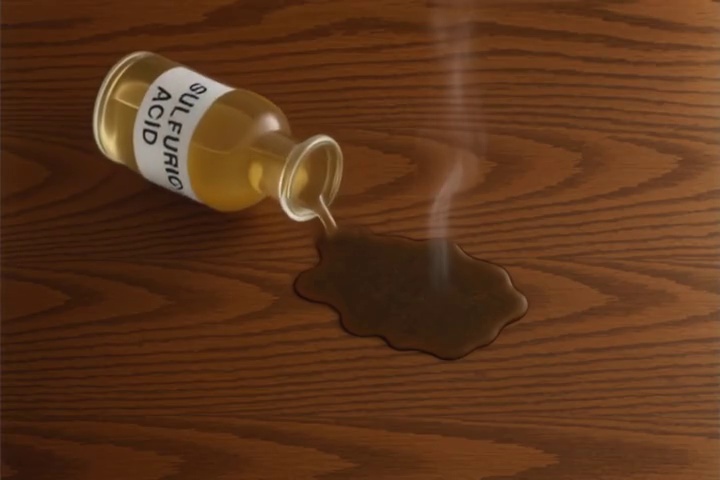} &
\includegraphics[width=0.18\textwidth]{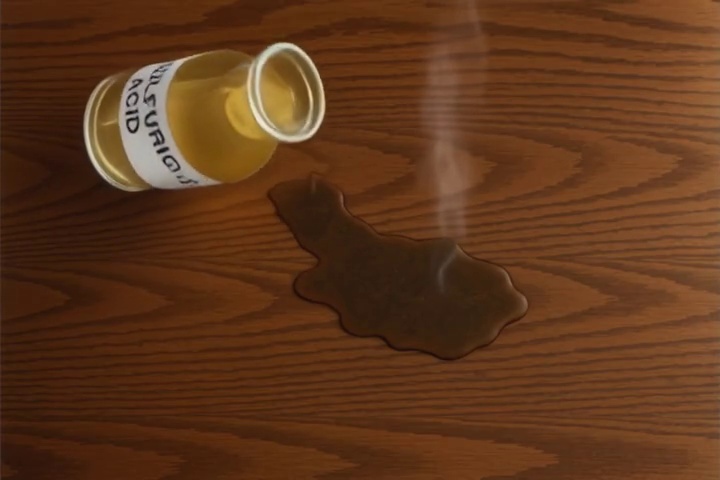} &
\includegraphics[width=0.18\textwidth]{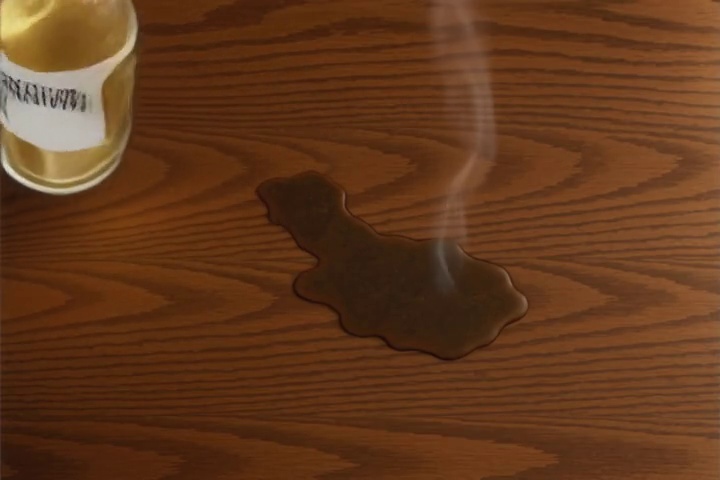} &
\includegraphics[width=0.18\textwidth]{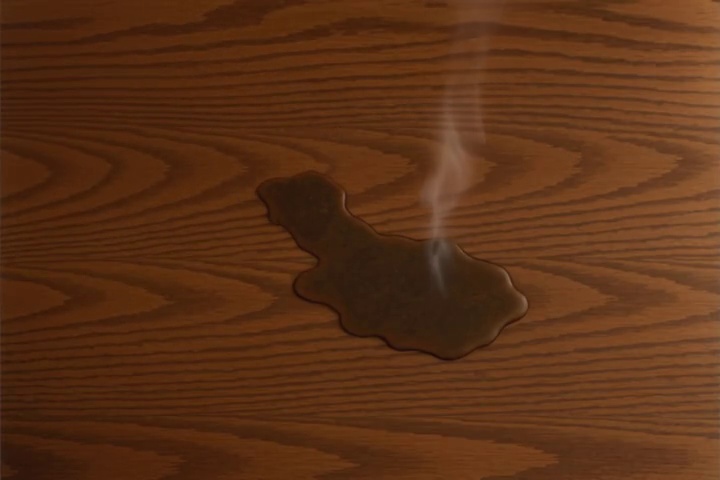} &
\includegraphics[width=0.18\textwidth]{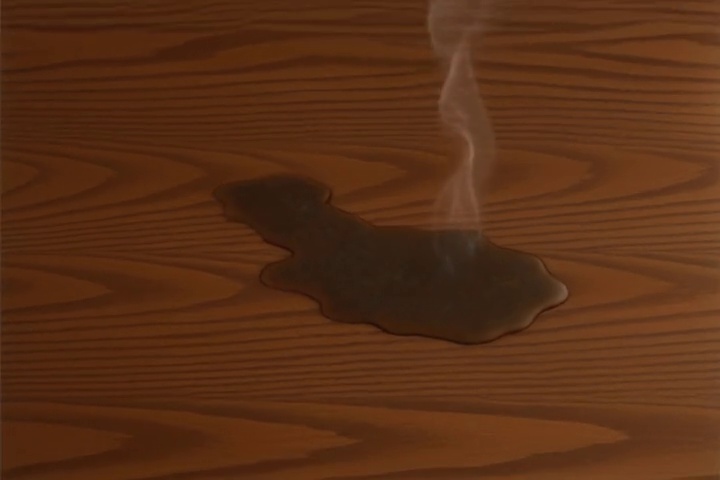} \\

\end{tabular}

\vspace{2mm}

A chameleon eats a flying insect.  \\

\begin{tabular}{c c c c c}

\includegraphics[width=0.18\textwidth]{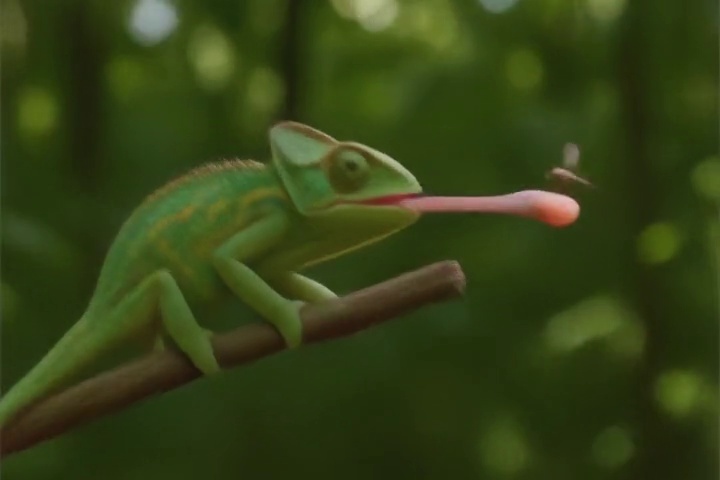} &
\includegraphics[width=0.18\textwidth]{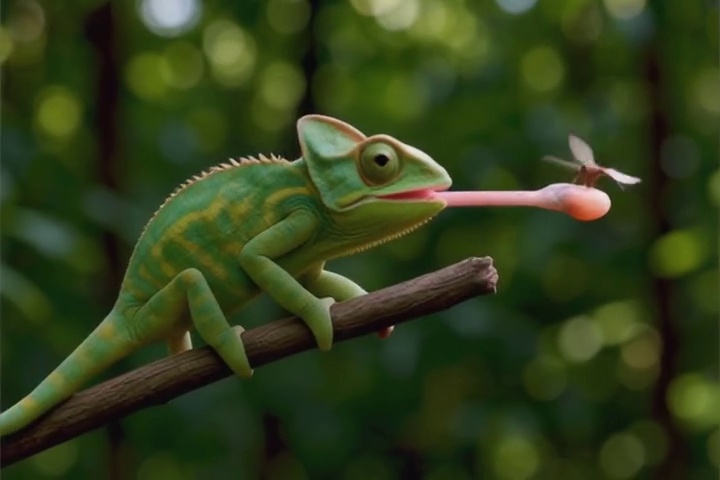} &
\includegraphics[width=0.18\textwidth]{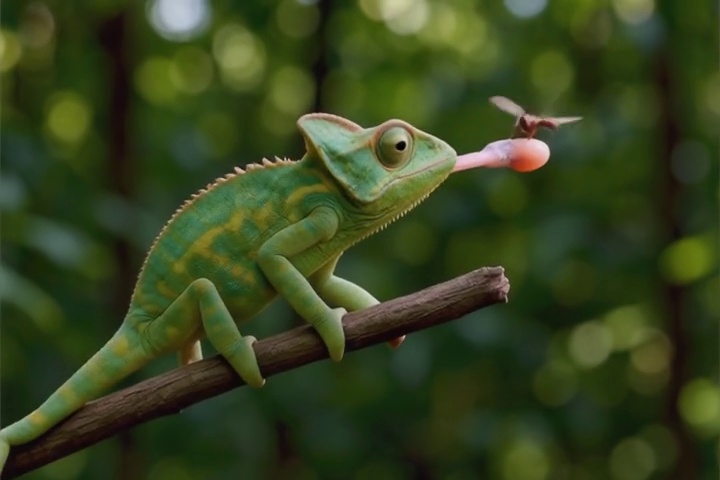} &
\includegraphics[width=0.18\textwidth]{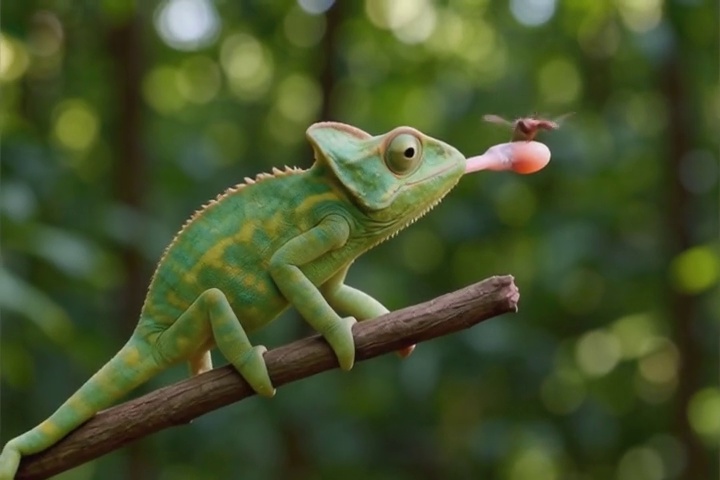} &
\includegraphics[width=0.18\textwidth]{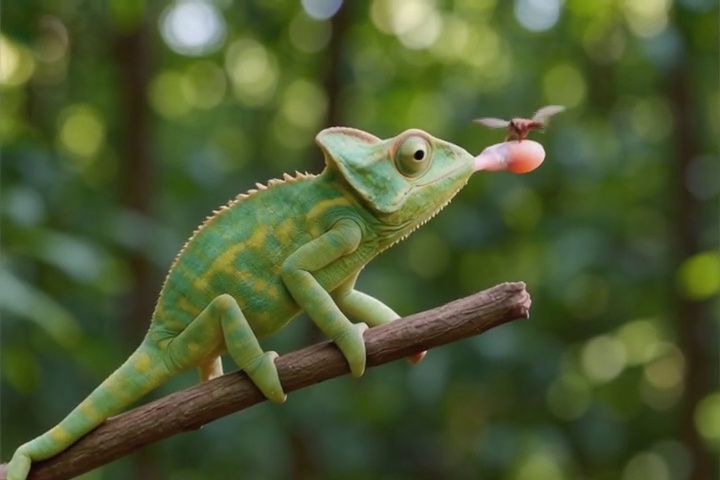} \\

\end{tabular}

\vspace{2mm}

An egg falling from the sky towards concrete ground. \\

\begin{tabular}{c c c c c}

\includegraphics[width=0.18\textwidth]{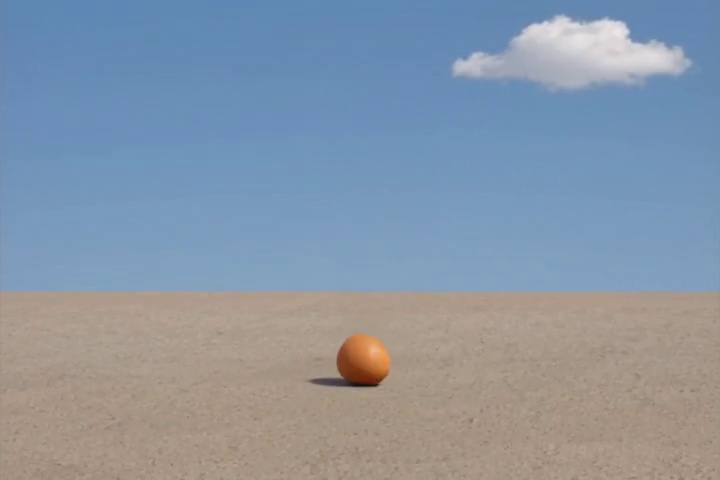} &
\includegraphics[width=0.18\textwidth]{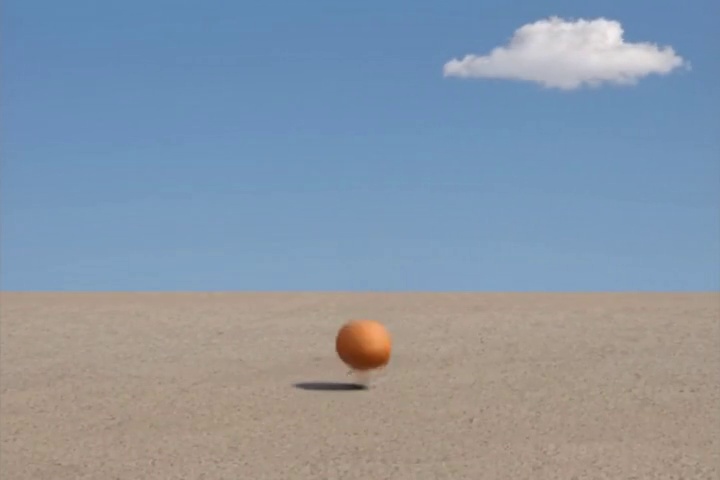} &
\includegraphics[width=0.18\textwidth]{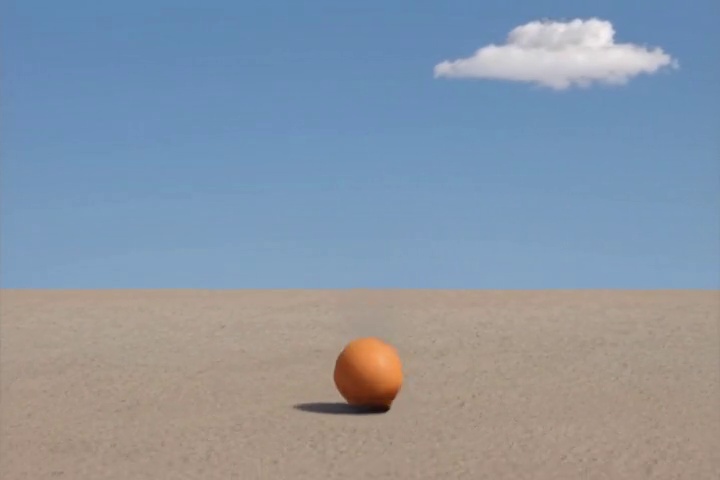} &
\includegraphics[width=0.18\textwidth]{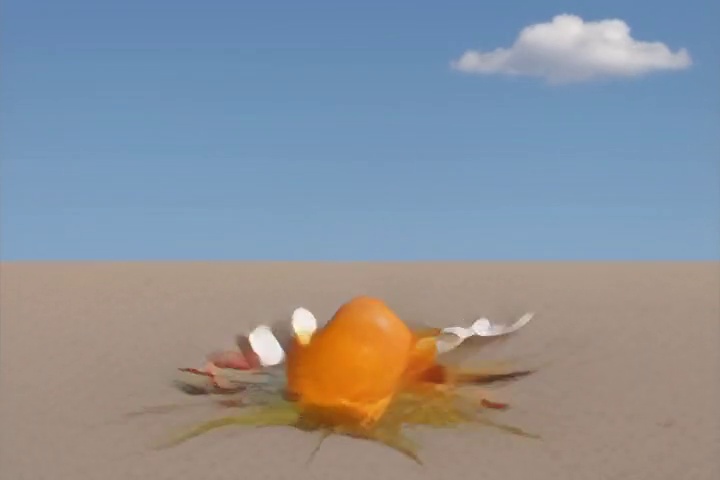} &
\includegraphics[width=0.18\textwidth]{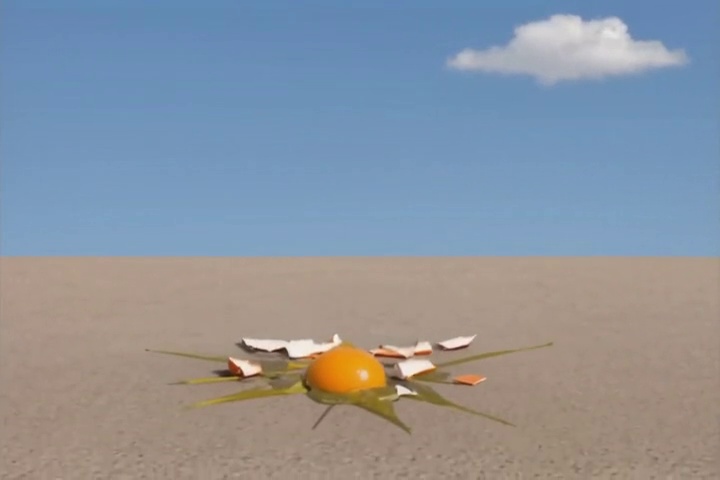} \\

\end{tabular}

\vspace{2mm}

Oil is poured into a glass of milk.  \\

\begin{tabular}{c c c c c}

\includegraphics[width=0.18\textwidth]{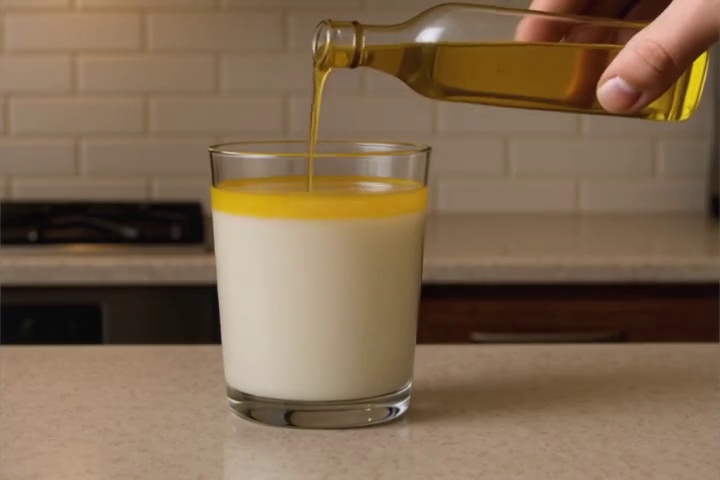} &
\includegraphics[width=0.18\textwidth]{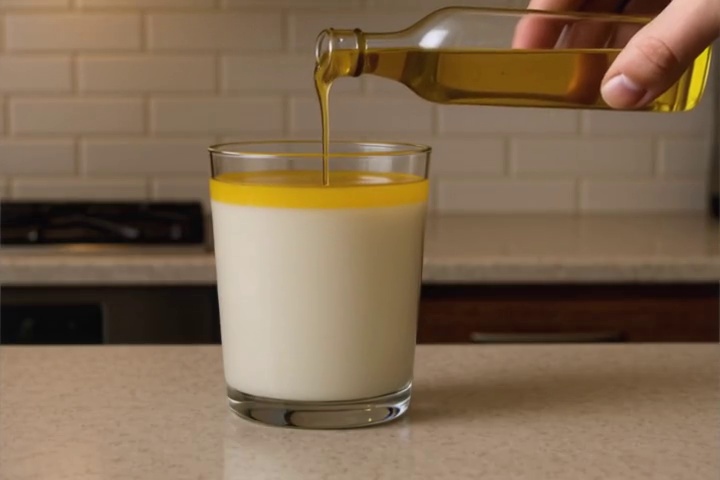} &
\includegraphics[width=0.18\textwidth]{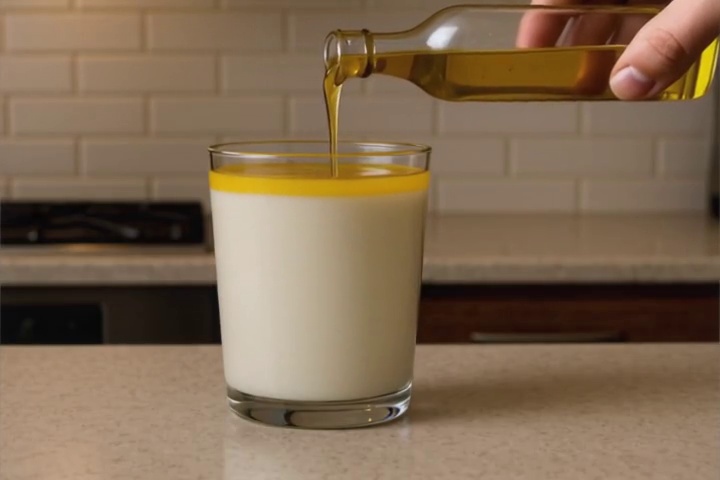} &
\includegraphics[width=0.18\textwidth]{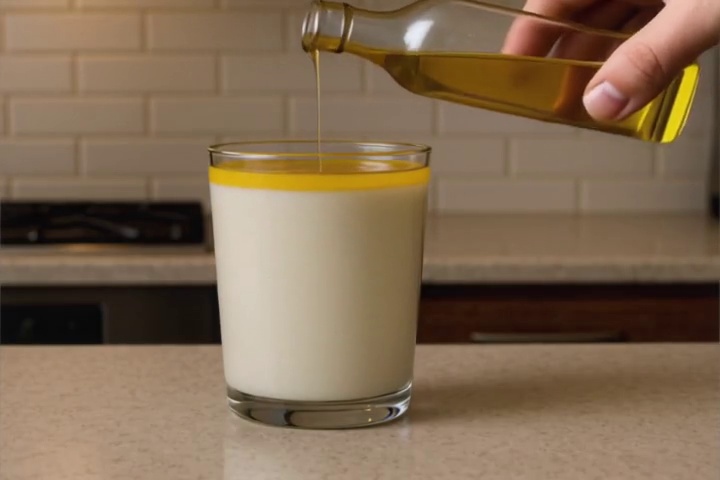} &
\includegraphics[width=0.18\textwidth]{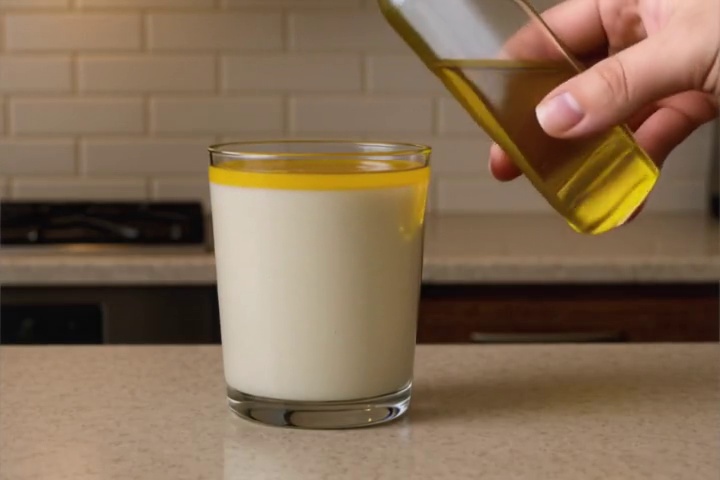} \\
\end{tabular}

\caption{Additional qualitative comparisons between VLIPP and our method on diverse physical phenomena.}
\label{fig:sup_qualitative_results_2}
\end{figure}

\begin{figure}[htbp]
\centering
\setlength{\tabcolsep}{0pt}
\renewcommand{\arraystretch}{0}

A small burning stick was thrown into a pile of hay.  \\

\begin{tabular}{c c c c c}

\includegraphics[width=0.18\textwidth]{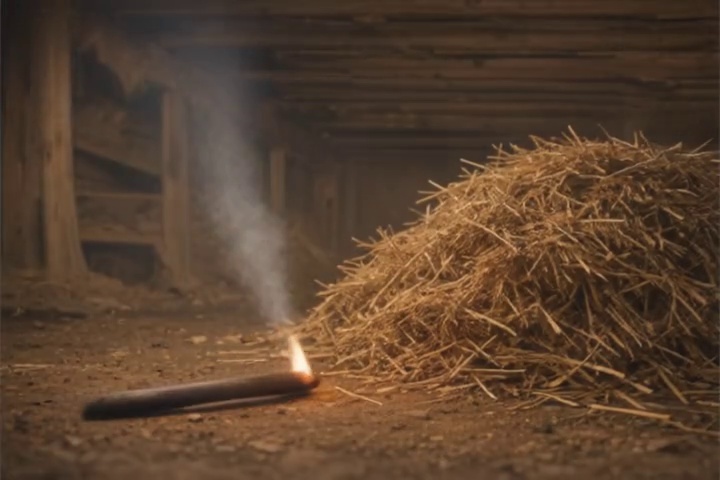} &
\includegraphics[width=0.18\textwidth]{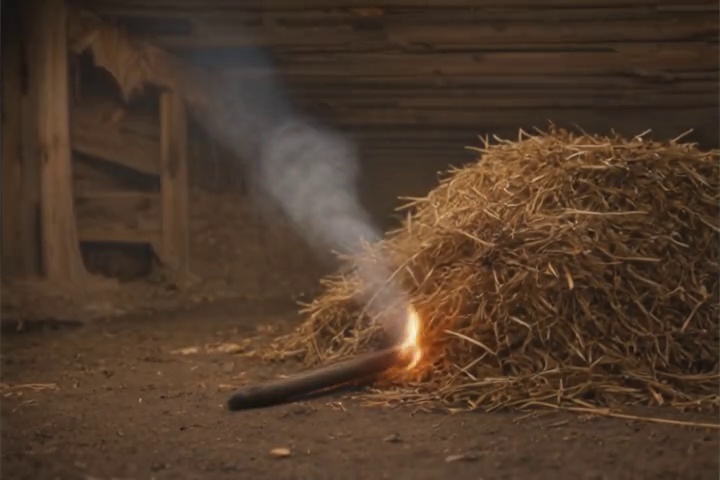} &
\includegraphics[width=0.18\textwidth]{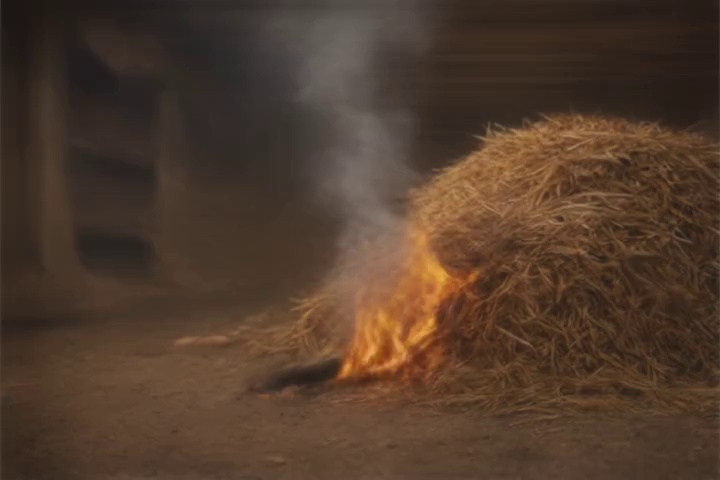} &
\includegraphics[width=0.18\textwidth]{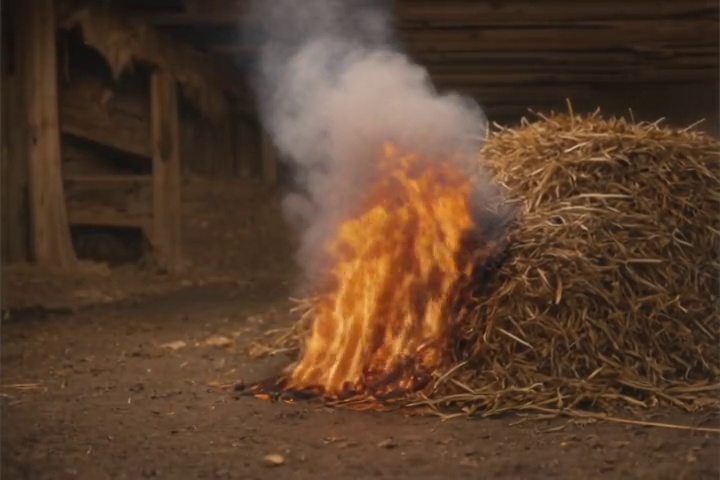} &
\includegraphics[width=0.18\textwidth]{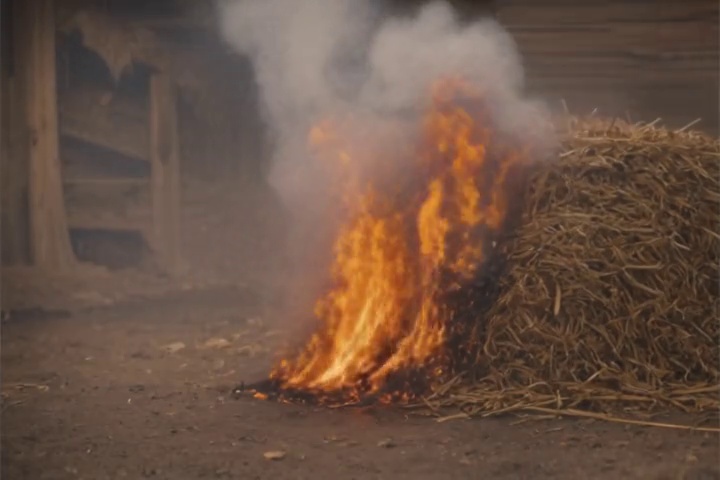} \\

\end{tabular}

\vspace{2mm}

A tennis ball is gently placed on the surface of a bucket filled with water.  \\

\begin{tabular}{c c c c c}

\includegraphics[width=0.18\textwidth]{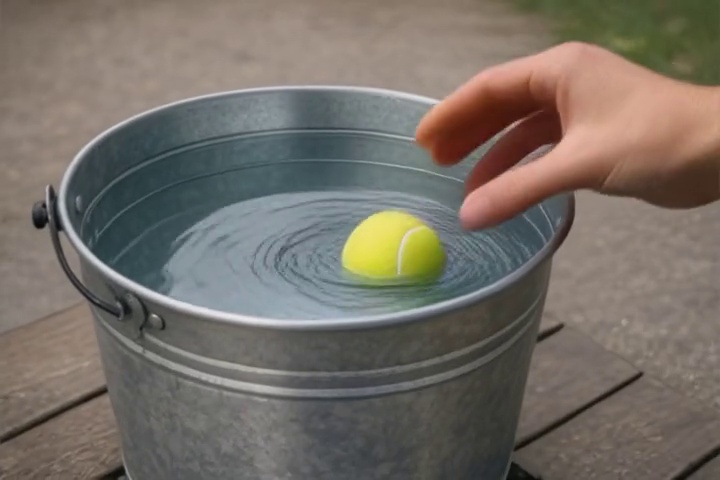} &
\includegraphics[width=0.18\textwidth]{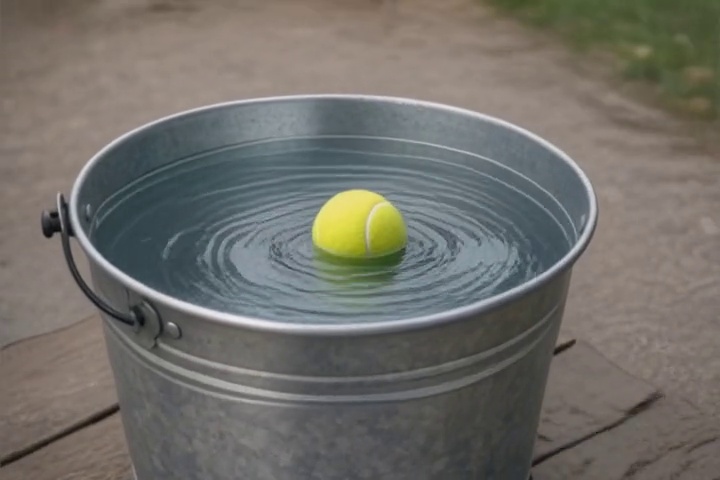} &
\includegraphics[width=0.18\textwidth]{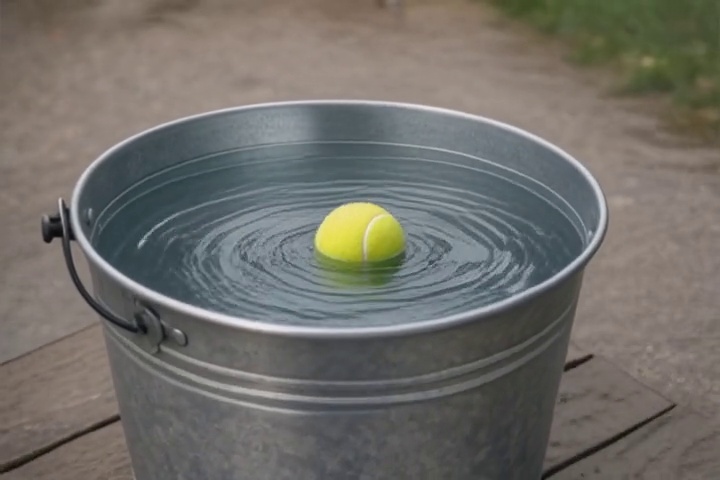} &
\includegraphics[width=0.18\textwidth]{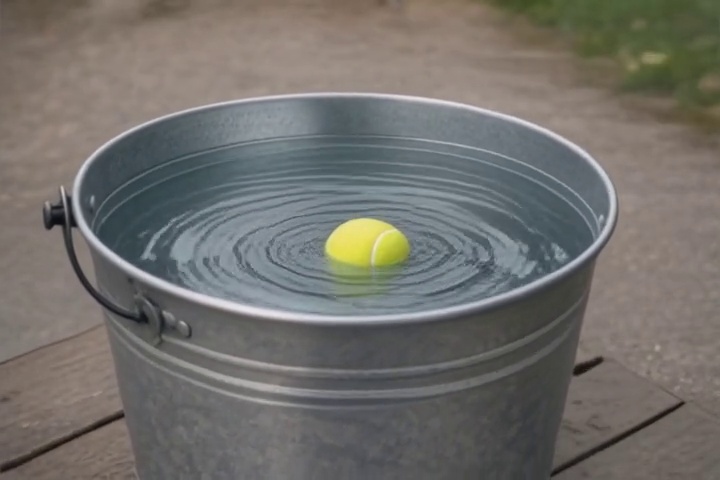} &
\includegraphics[width=0.18\textwidth]{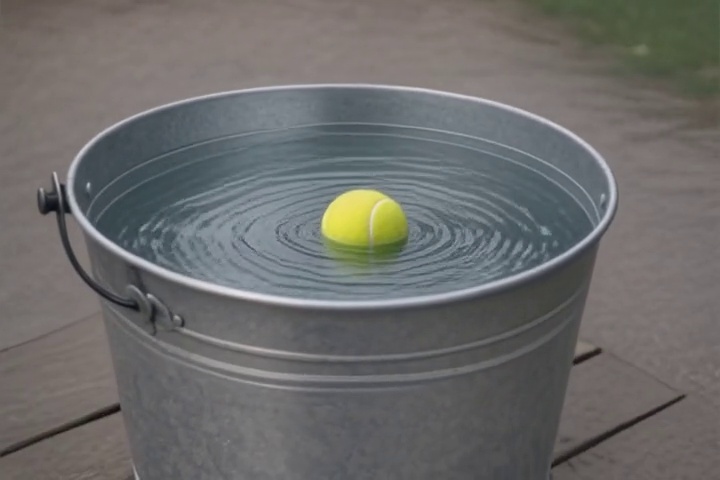} \\

\end{tabular}

\vspace{2mm}

A timelapse of a water-filled soft cloth being forcefully squeezed by hand, with the pressure intensifying rapidly over time.  \\

\begin{tabular}{c c c c c}

\includegraphics[width=0.18\textwidth]{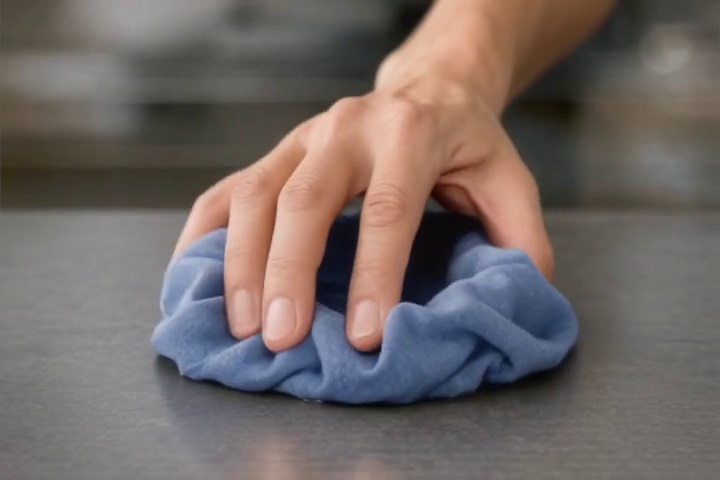} &
\includegraphics[width=0.18\textwidth]{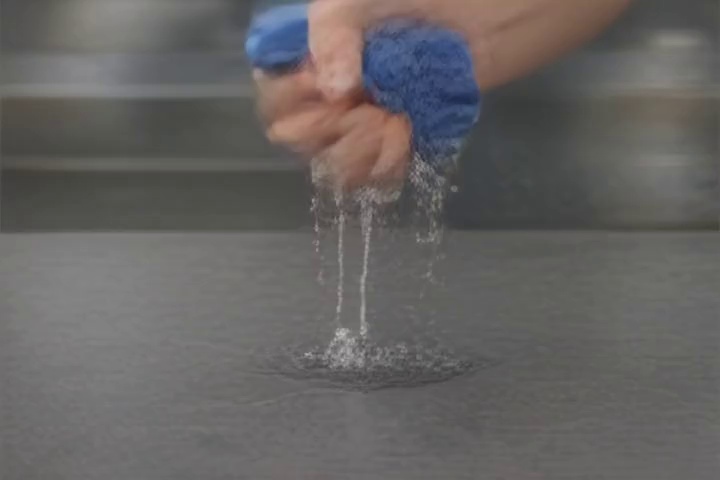} &
\includegraphics[width=0.18\textwidth]{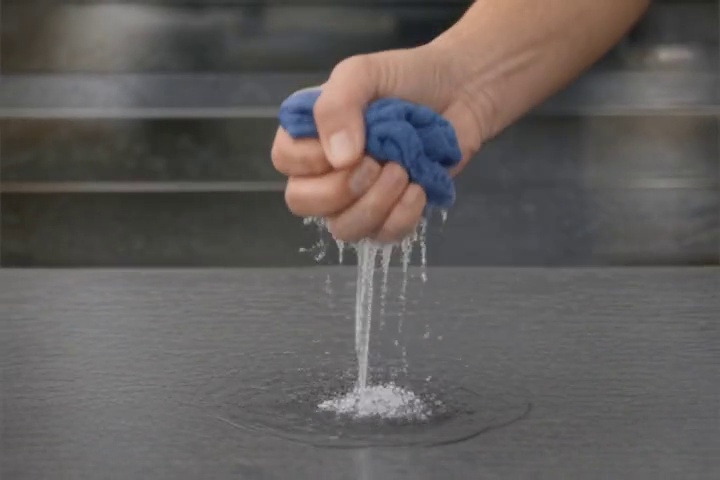} &
\includegraphics[width=0.18\textwidth]{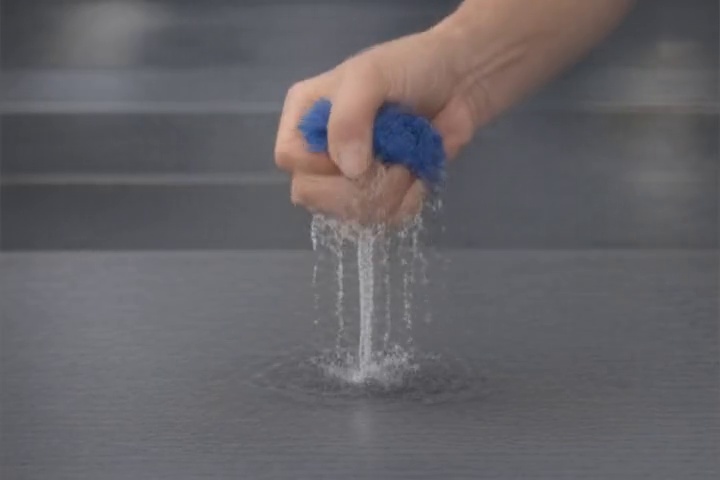} &
\includegraphics[width=0.18\textwidth]{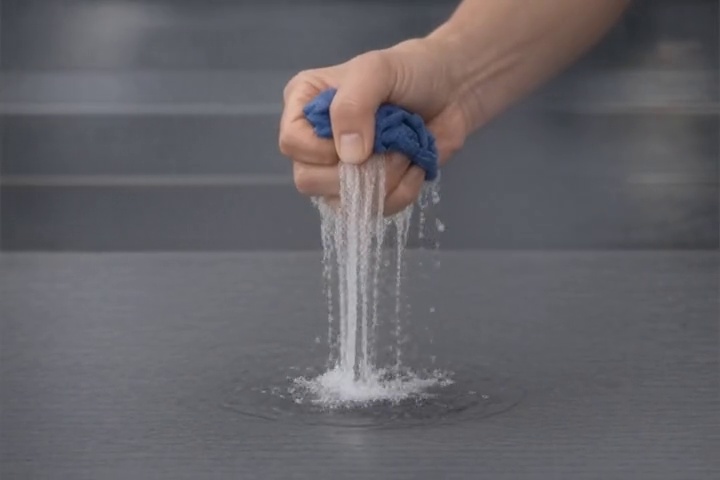} \\

\end{tabular}

\vspace{2mm}

A ray of light is shining diagonally on a glass bottle in the dark, with the shadow of the glass bottle appearing at the bottom.  \\

\begin{tabular}{c c c c c}

\includegraphics[width=0.18\textwidth]{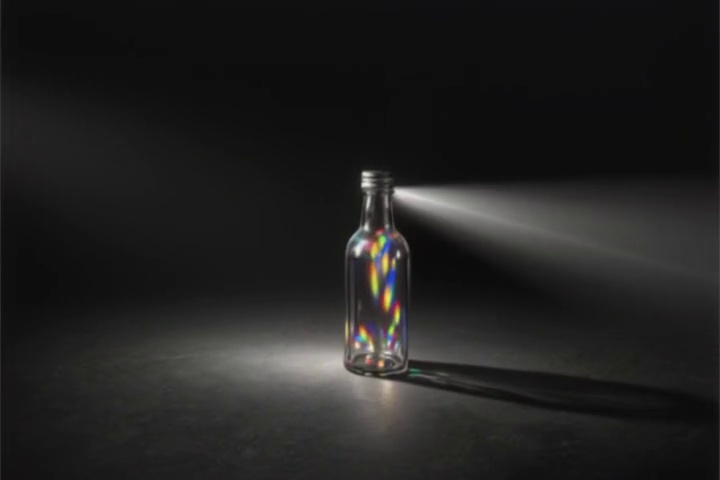} &
\includegraphics[width=0.18\textwidth]{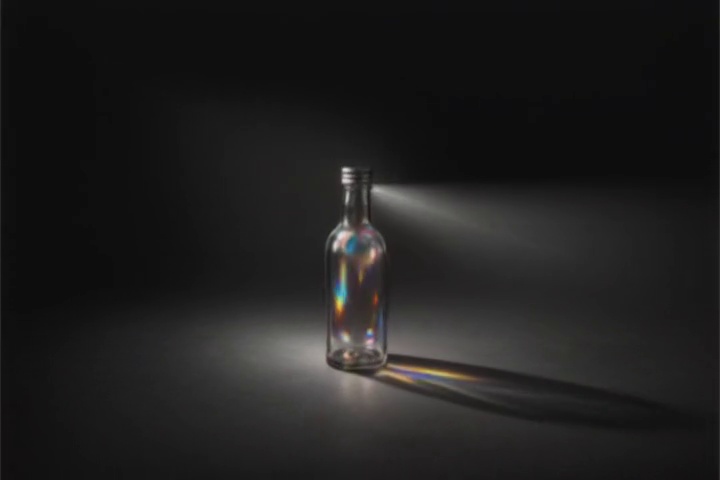} &
\includegraphics[width=0.18\textwidth]{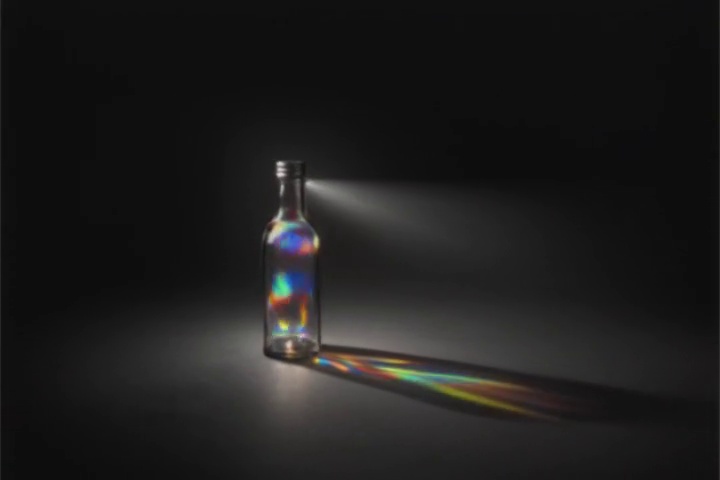} &
\includegraphics[width=0.18\textwidth]{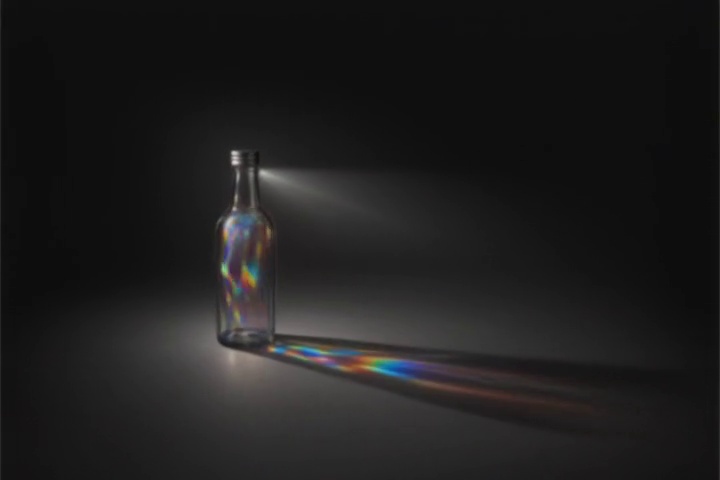} &
\includegraphics[width=0.18\textwidth]{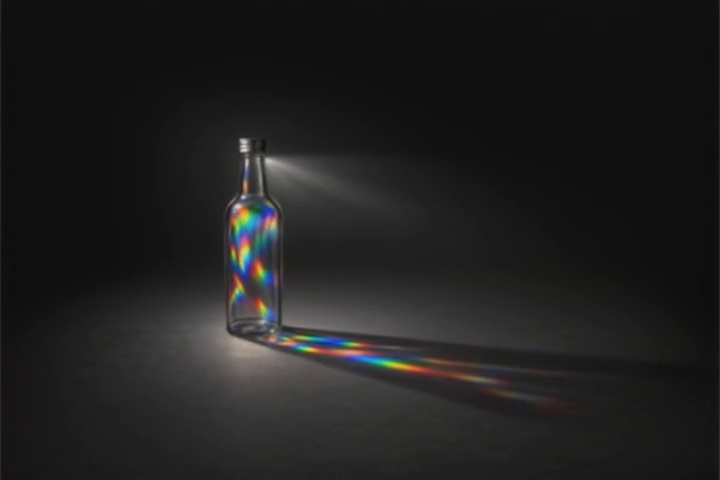} \\

\end{tabular}

\vspace{2mm}

A piece of white chalk is used to write on the rough, dark surface of a blackboard,showcasing the interaction between the chalk and the blackboard surface.  \\

\begin{tabular}{c c c c c}

\includegraphics[width=0.18\textwidth]{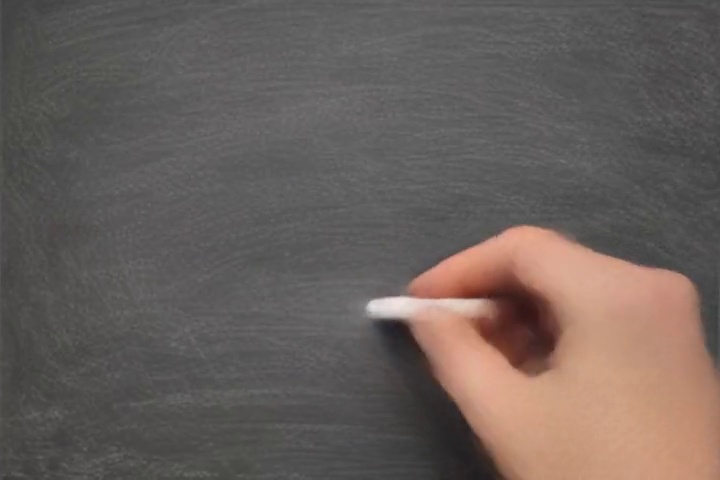} &
\includegraphics[width=0.18\textwidth]{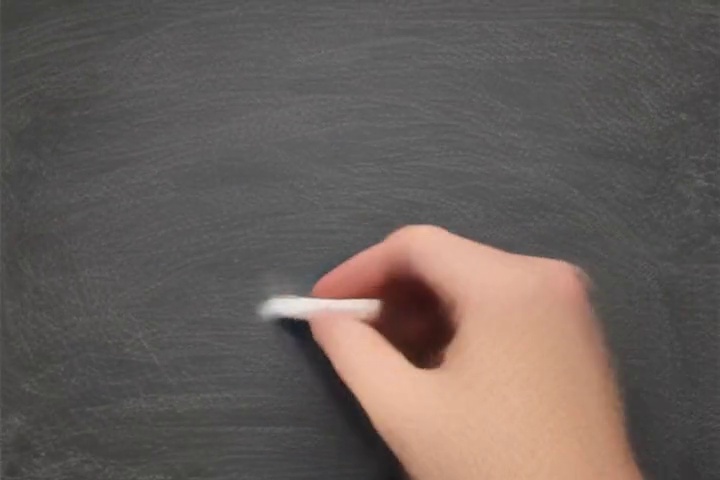} &
\includegraphics[width=0.18\textwidth]{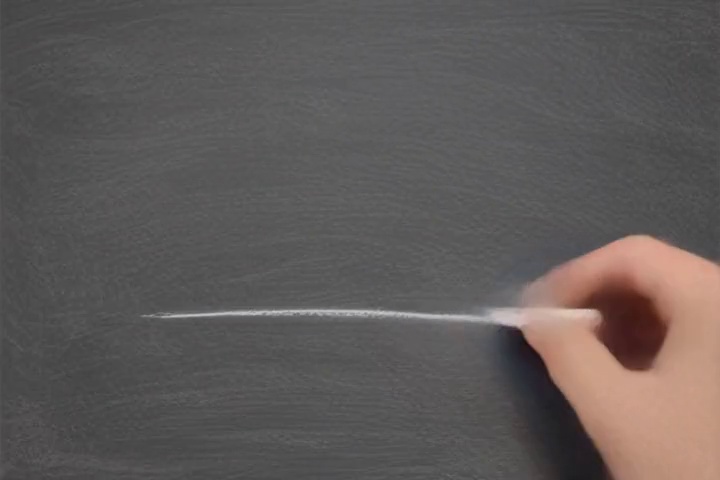} &
\includegraphics[width=0.18\textwidth]{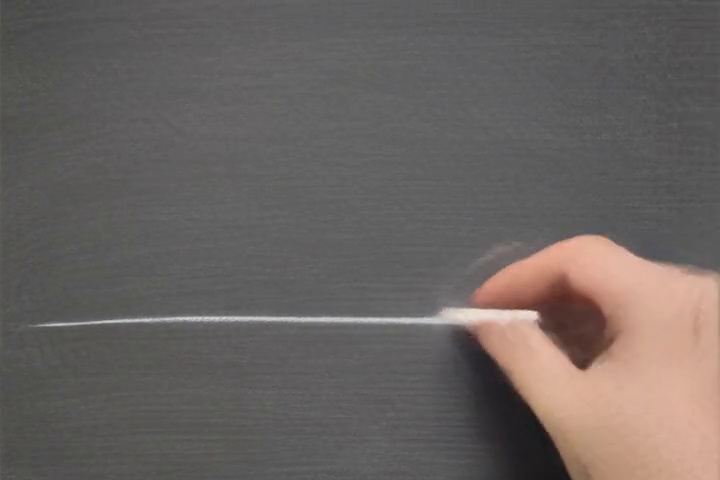} &
\includegraphics[width=0.18\textwidth]{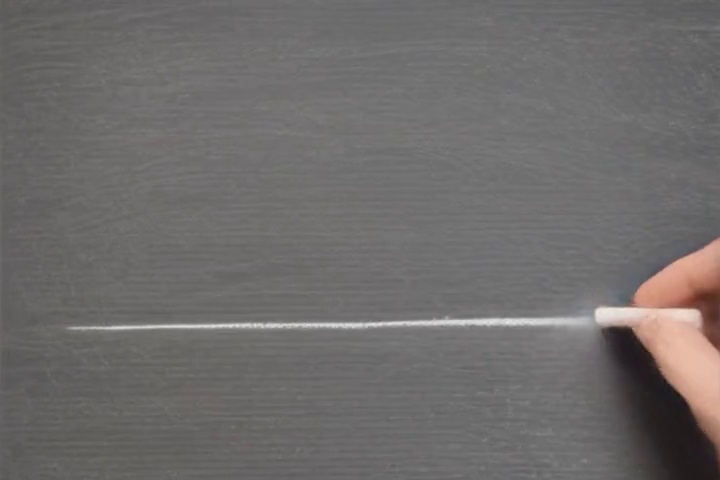} \\

\end{tabular}

\vspace{2mm}

A magnifying glass is gradually moving closer to a leaf, revealing the intricate details and textures of the veins and surface patterns as it approaches.  \\

\begin{tabular}{c c c c c}

\includegraphics[width=0.18\textwidth]{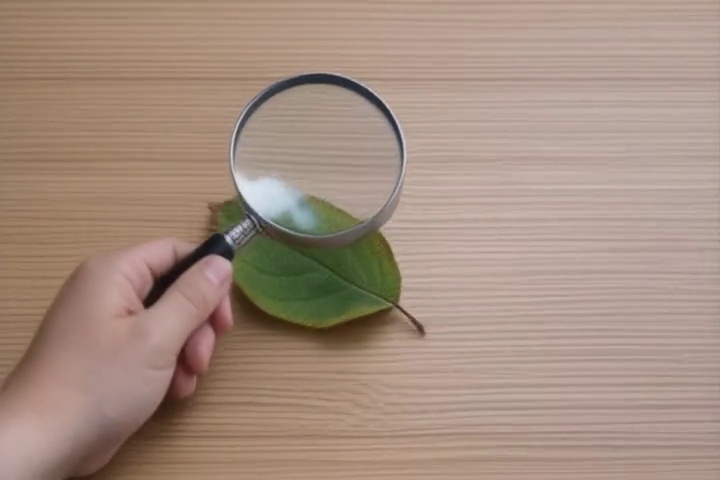} &
\includegraphics[width=0.18\textwidth]{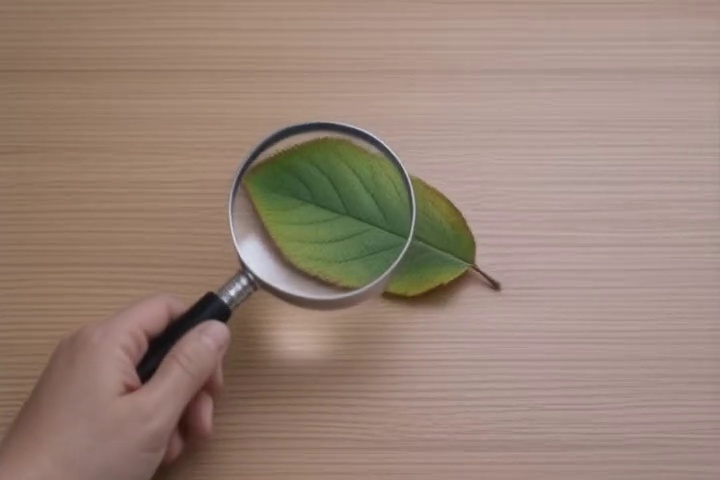} &
\includegraphics[width=0.18\textwidth]{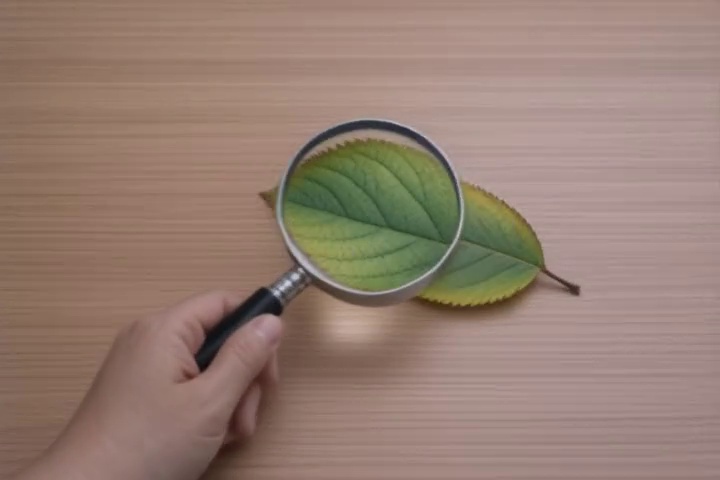} &
\includegraphics[width=0.18\textwidth]{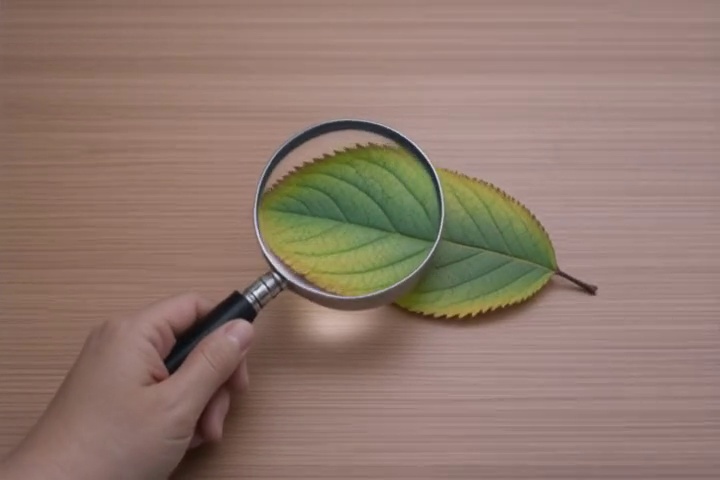} &
\includegraphics[width=0.18\textwidth]{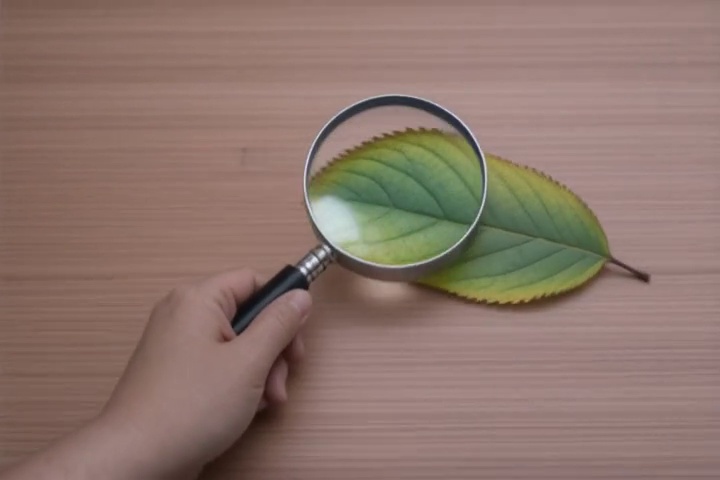} \\

\end{tabular}

\vspace{2mm}

A bunch of mist particles are suspended in the air under the sunlight.  \\

\begin{tabular}{c c c c c}

\includegraphics[width=0.18\textwidth]{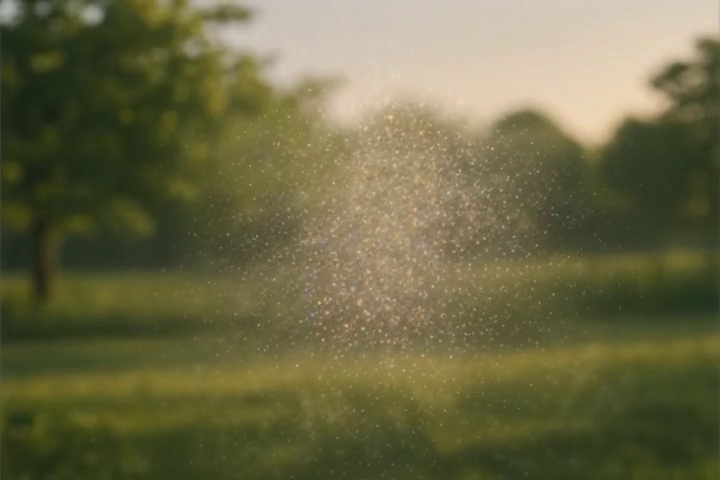} &
\includegraphics[width=0.18\textwidth]{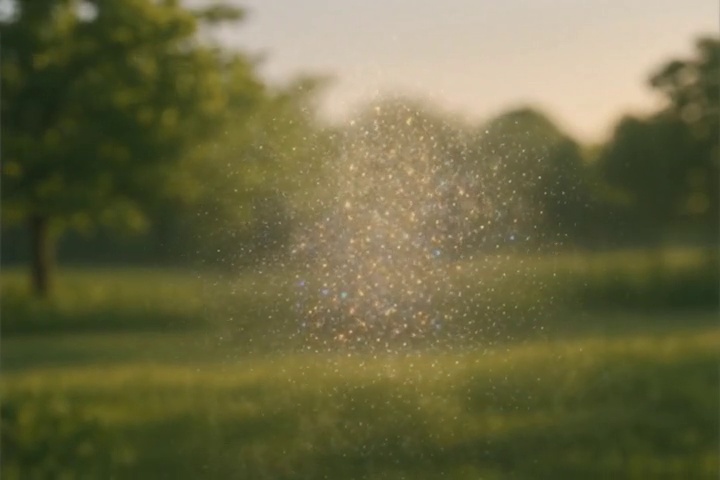} &
\includegraphics[width=0.18\textwidth]{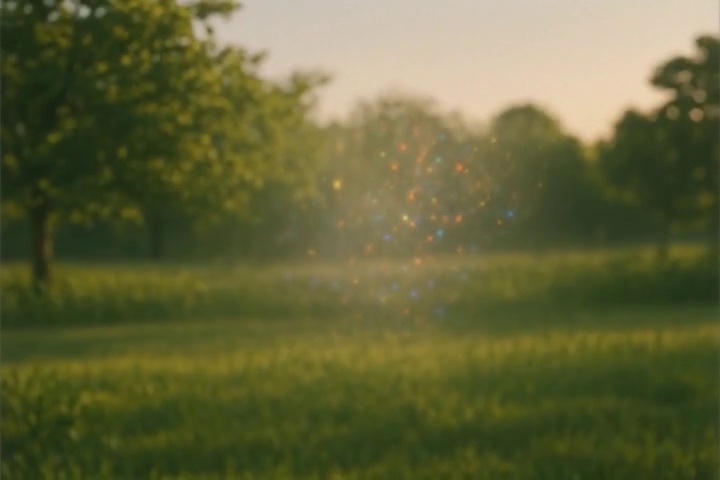} &
\includegraphics[width=0.18\textwidth]{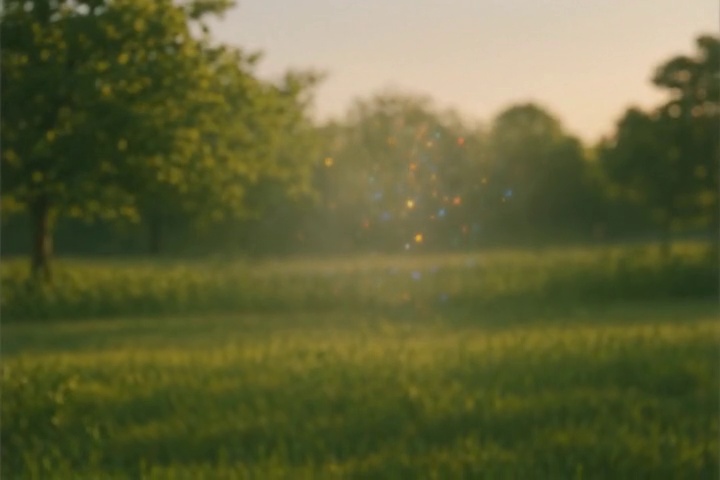} &
\includegraphics[width=0.18\textwidth]{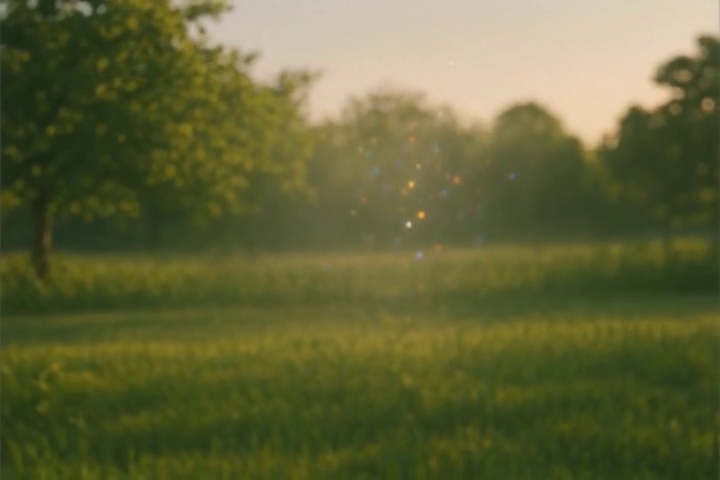} \\

\end{tabular}

\vspace{2mm}

A timelapse captures the gradual transformation of ice cream as the temperature rises significantly.  \\

\begin{tabular}{c c c c c}

\includegraphics[width=0.18\textwidth]{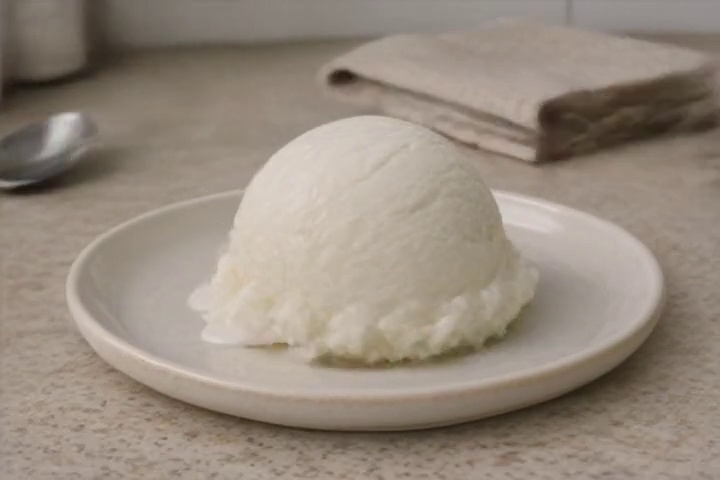} &
\includegraphics[width=0.18\textwidth]{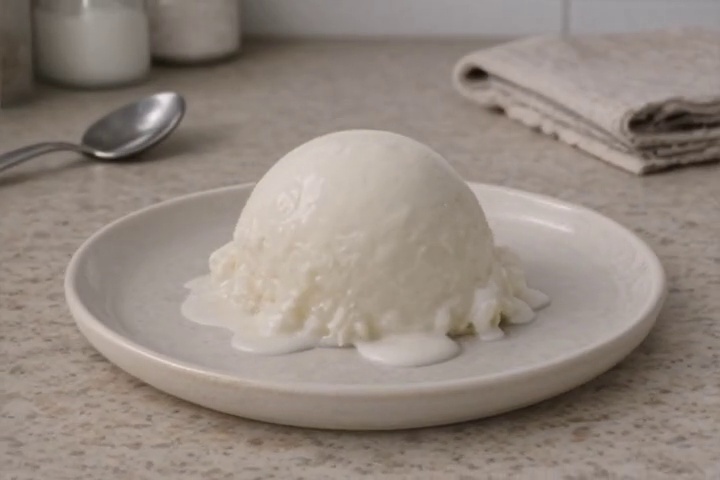} &
\includegraphics[width=0.18\textwidth]{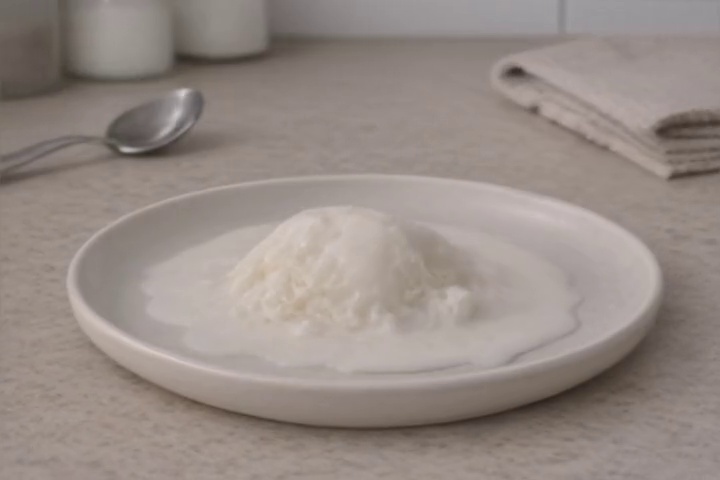} &
\includegraphics[width=0.18\textwidth]{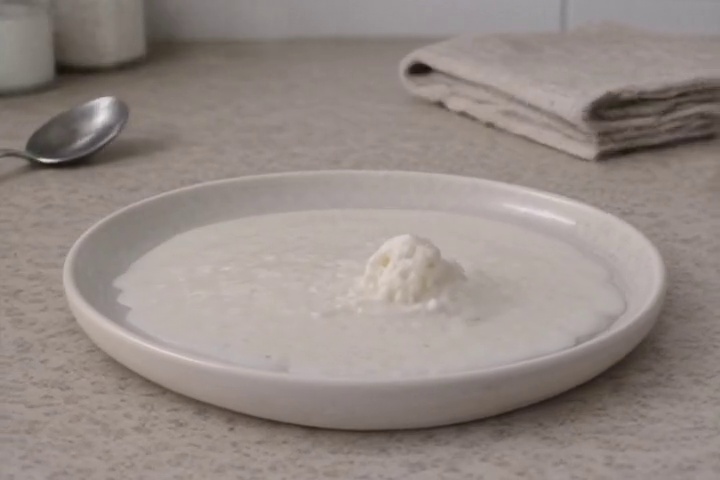} &
\includegraphics[width=0.18\textwidth]{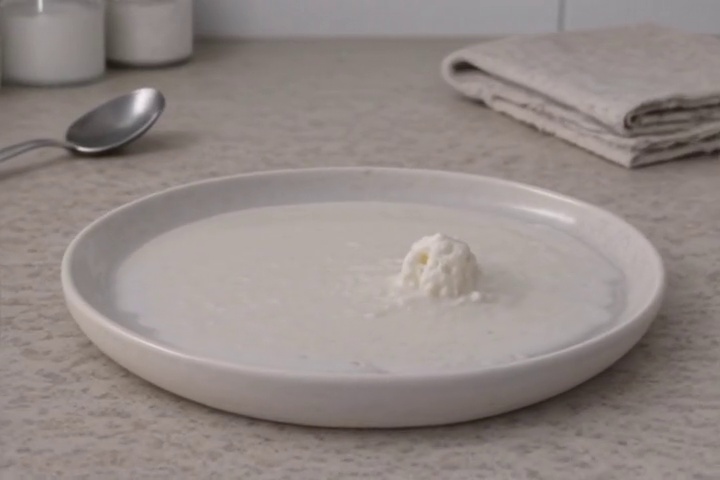} \\

\end{tabular}
\caption{More qualitative results generated by our method across diverse physical domains.}
\label{fig:sup_qualitative_results_4}
\end{figure}


\newpage
\clearpage
\bibliography{references}

\end{document}